%% file: main.tex
\documentclass[10pt,twocolumn,letterpaper]{article}

\usepackage[final]{cvpr}
\input{imports}

\crefname{section}{Sec.}{Secs.}
\Crefname{section}{Section}{Sections}
\Crefname{table}{Table}{Tables}
\crefname{table}{Tab.}{Tabs.}

\def\cvprPaperID{4844} % *** Enter the CVPR Paper ID here
\def\confName{CVPR}
\def\confYear{2022}

\begin{document}

%\title{Learning Embodied Visual Exploration from Humans}
% \title{Emergence of Efficient Object-Search Behavior in Embodied Agents \\
% from Large-Scale Imitation of Human Demonstrations}
%\title{Emergence of Strategic Object-Search Behavior in Embodied Agents \\
%from Large-Scale Human Imitation}
%\title{\datasetname: Learning Embodied Object-Search Strategies from Humans}
\title{Habitat-Web: Learning Embodied Object-Search Strategies \\from Human Demonstrations at Scale\vspace{-15pt}}

\author{
  Ram Ramrakhya$^1$ \,\,
  Eric Undersander$^2$ \,\,
  Dhruv Batra$^{1,2}$ \,\,
  Abhishek Das$^{2}$ \vspace{3pt}\\
  $^1$Georgia Institute of Technology \quad $^2$Meta AI Research\\
  {\tt\small $^1$\{ram.ramrakhya,dbatra\}@gatech.edu}
  \quad {\tt\small $^2$\{eundersander,abhshkdz\}@fb.com}
}
\maketitle

%%%%%%%%% ABSTRACT
\begin{abstract}
We present a large-scale study of imitating human demonstrations on tasks that
require a virtual robot to search for objects in new environments -- (1) \objnavfull
(\eg \myquote{find \& go to a chair}) and (2) \pickplace (\eg \myquote{find mug, pick mug,
find counter, place mug on counter}).
First, we develop a virtual teleoperation data-collection infrastructure --
connecting Habitat simulator running in a web browser to Amazon Mechanical
Turk, allowing remote users to teleoperate virtual robots, safely and at scale.
%
%Two version for the next sentence: \\
%1.
We collect $80k$ demonstrations for \objnav and $12k$ demonstrations for
\pickplace,
% \footnote{\label{note1}$80k$ (out of $100k$)
% demonstrations for \objnav and $12k$ demonstrations of \pickplace have been collected,
% which are used for all results in this paper.
% %
% We expect the complete data collection to be in place by the final version.},
which is an order of magnitude
% \todo{NOTE: I'm comparing to 3k from TEACH paper; check}
larger than existing human demonstration datasets
in simulation or on real robots.
%
%2. We collect 40k demonstrations for ObjectNav (in process of being scaled to 100k) and 12k demonstrations for \PicknPlace, which is 1 orders of magnitude [NOTE: I'm comparing to 3k from TEACH paper; check] larger than existing human demonstration datasets in simulation and 3 orders of magnitude larger than existing datasets on real robots.
%
%
Our virtual teleoperation data contains $29.3M$ actions, and is equivalent to
$22.6k$ hours of real-world teleoperation time, and illustrates
%This data is not only large-scale but also rich
%and
rich, diverse strategies for solving the tasks.
%
%This data enables us to answer the question --
Second, we use this data to answer the question -- how does large-scale
imitation learning (IL) (which has not been hitherto possible) compare to
reinforcement learning (RL) (which is the status quo)?
On \objnav, we find that IL (with no bells or whistles) using $70k$ human
%demonstrations is comparable to RL using $240k$ agent-gathered trajectories.
demonstrations outperforms RL using $240k$ agent-gathered trajectories.
This effectively establishes an `exchange rate' -- a single human demonstration
% appears to be worth $6$ agent-gathered ones.
appears to be worth ${\sim}4$ agent-gathered ones.
More importantly, we find the IL-trained agent learns efficient object-search behavior from humans --
it peeks into rooms, checks corners for small objects, turns in place to get a panoramic view --
none of these are exhibited as prominently by the RL agent, and to induce these behaviors
via contemporary RL techniques would require tedious reward engineering.
Finally, accuracy~\vs training data size plots show promising scaling behavior, suggesting
that simply collecting more demonstrations is likely to advance the state of art further.
On \pickplace, the comparison is starker -- IL agents achieve ${\sim}18\%$
success on episodes with new object-receptacle locations when trained
with $9.5k$ human demonstrations, while RL agents fail to get beyond $0\%$.
Overall, our work provides compelling evidence for investing in large-scale imitation learning.\\
Project page: \href{https://ram81.github.io/projects/habitat-web}{\tt{ram81.github.io/projects/habitat-web}}.
\end{abstract}

\input{sections/main/intro}

\input{sections/main/related}
\input{sections/main/dataset}
\input{sections/main/approach}
\input{sections/main/results}
\input{sections/main/analysis}

\input{sections/main/conclusion}

\input{sections/main/acknowledgement}

\bibliographystyle{ieeetr}
% \biboptions{sort&compress}
\bibliography{strings,main}

\input{sections/supplement/appendix}

\input{sections/supplement/analysis_figure}

% \input{sections/supplement/appendix}

%%%%%%%%% REFERENCES
% {\small
% \bibliographystyle{ieee_fullname}
% \bibliography{egbib}
% }

\end{document}

%% file: imports.tex
\usepackage{amsfonts}       % blackboard math symbols
\usepackage{nicefrac}       % compact symbols for 1/2, etc.

\usepackage{graphicx}
\usepackage{comment}
\usepackage{amsmath,amssymb,amstext} % define this before the line numbering.

\usepackage{gensymb}
\usepackage{relsize}
\usepackage[T1]{fontenc}
\usepackage[scaled=0.90]{inconsolata}
\usepackage[latin1]{inputenc}
\usepackage[english]{babel}

\usepackage{epsfig}
\usepackage{graphicx}
\usepackage{wrapfig}
\usepackage[belowskip=2pt,labelfont=bf,aboveskip=0pt,font=small,labelsep=period]{caption}
\usepackage{array}

\usepackage[belowskip=0pt,aboveskip=0pt,font=small]{subcaption}
\usepackage{longtable}
\usepackage{microtype}

\usepackage{textcomp}
\usepackage{stmaryrd}
\SetSymbolFont{stmry}{bold}{U}{stmry}{m}{n}
\usepackage{upgreek}
\usepackage{bm}
\usepackage{cases}
\usepackage{mathtools}
\usepackage{arydshln}
\usepackage{multirow}

\usepackage{color}
\usepackage[dvipsnames]{xcolor}
\definecolor{JalapenoRed}{RGB}{183,21,64}
\definecolor{Belize}{RGB}{41,128,185}
\definecolor{Amour}{RGB}{238,82,83}

\usepackage[pagebackref=true,breaklinks=true,colorlinks,bookmarks=false,allcolors=JalapenoRed]{hyperref}
\usepackage[capitalize]{cleveref}
\crefname{section}{Sec.}{Secs.}
\Crefname{section}{Section}{Sections}
\Crefname{table}{Table}{Tables}
\crefname{table}{Tab.}{Tabs.}

\usepackage{algorithm}
\usepackage{algpseudocode}
\usepackage{multirow}
\usepackage{rotating}
\usepackage{booktabs}
\usepackage{etoolbox}
\newcounter{magicrownumbers}
\preto\tabular{\setcounter{magicrownumbers}{0}}
\newcommand\rownumber{\stepcounter{magicrownumbers}\arabic{magicrownumbers})\,}

\usepackage{enumitem}
\usepackage[olditem,oldenum]{paralist}

\usepackage{alltt}
\usepackage{listings}

\usepackage{url}
\usepackage{xspace}
\usepackage{comment}
\usepackage{cuted}
\usepackage{caption}

\usepackage{quoting}
\quotingsetup{vskip=0pt}
\usepackage{epigraph}
\usepackage[normalem]{ulem}

\setdefaultleftmargin{.2cm}{.2cm}{}{}{}{}
\setlist{leftmargin=.2cm}

\newcommand{\ad}[1]{\textcolor{Belize}{#1}}

\usepackage{mysymbols}
\graphicspath{{images/}}

%% file: sections/main/intro.tex
\vspace{-15pt}
\section{Introduction}
\label{sec:intro}

\begin{figure*}[t]
  \centering
    \includegraphics[width=0.98\textwidth]{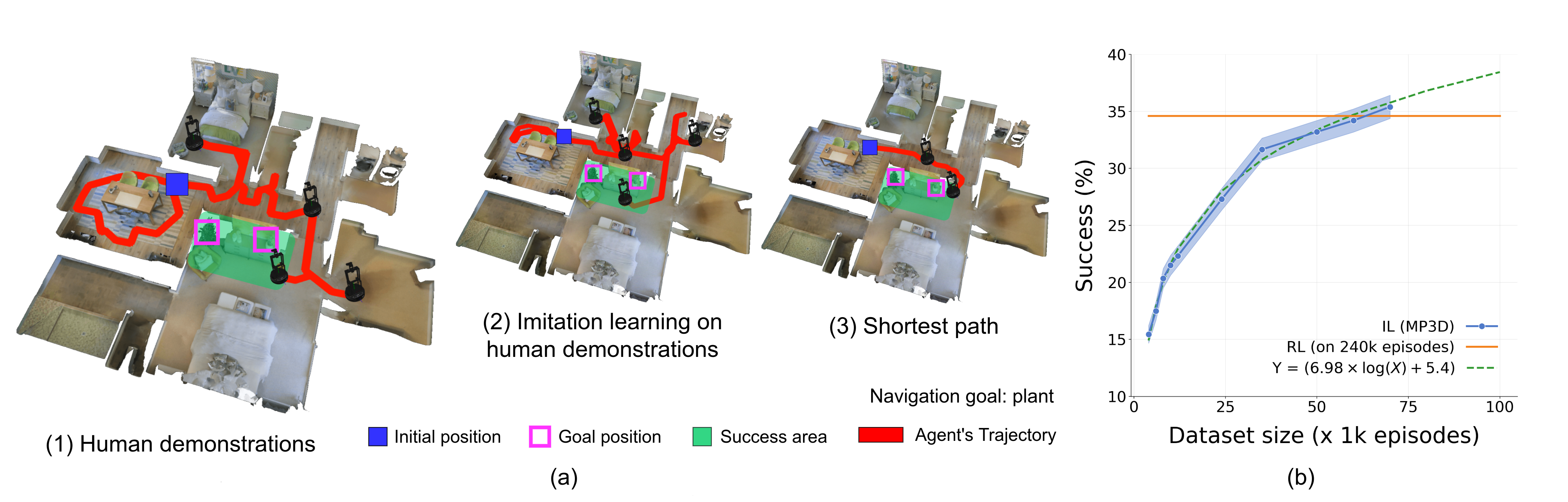}
    %\vspace{-4pt}
    \caption{a) Example \objnav 1) human demonstration,
      2) agent trained on human demonstrations, and
      3) shortest path.
      Notice how humans demonstrate sophisticated exploration behavior to succeed
      at this task in unseen environments, which is hard to engineer into the
      right reward for an RL agent and is unlikely to be captured in shortest path
      demonstrations.
      An agent trained on human demonstrations learns this exploration
      and object-search behavior.
      b) Success on the \objnav MP3D-\textsc{val} split~\vs no. of human demonstrations for training.}
    \vspace{-15pt}
    \label{fig:teaser}

%   \begin{subfigure}{0.75\linewidth}
%     \includegraphics[width=0.98\textwidth]{figures/teaser/teaser_isometric_1x3_v3.png}
%     \vspace{-5pt}
%     \caption{Example \objnav 1) human demonstration,
%       2) an agent trained on human demonstrations, and
%       3) a shortest path.
%       %
%       Notice how humans demonstrate sophisticated exploration behavior to succeed
%       right reward for an RL agent and is unlikely to be captured in shortest path
%       demonstrations.
%       %
%       An agent trained on human demonstrations learns some of this exploration
%       and object-search behavior.}
%     \label{fig:teaser}
%   \end{subfigure}
%   \,\,
%   \begin{subfigure}{0.22\linewidth}
%     \includegraphics[width=\linewidth]{figures/performance/dataset_size_comparison.png}
%     \vspace{25pt}
%     \caption{Success on the \objnav val split~\vs size of the human demonstrations training dataset.}
%     \label{fig:success_vs_dataset_size}
%   \end{subfigure}
%   \vspace{2pt}

%   \caption{}
%   \label{fig:teaser_scaling}
\end{figure*}

% General-purpose robots that can perform a diverse set of embodied tasks in a
% diverse set of environments \emph{have} must necessarily be good at visual exploration (among
% other things).
General-purpose robots that can perform a diverse set of embodied tasks in a
diverse set of environments \emph{have} to be good at visual exploration.
Consider the canonical example of asking a household robot, \myquote{Where are my keys?}.
To answer this (assuming the robot does not remember the answer from memory),
the robot would have to search the house, often guided by intelligent priors --
\eg peeking into the washroom or kitchen might be sufficient to be reasonably
sure the keys are not there, while exhaustively searching the living room might
be much more important since keys are more likely to be there.
While doing so, the robot has to internally keep track of where all it has been
to avoid redundant search, and it might also have to interact with objects,~\eg
check drawers and cabinets in the living room (but not those in the washroom or
kitchen!).

This example illustrates fairly sophisticated exploration, involving
a careful interplay of various implicit objectives (semantic priors, exhaustive
search, efficient navigation, interaction, \etc).
Many recent tasks of interest in the embodied AI community --
~\eg~\objnavfull~\cite{savva2017minos,anderson_arxiv18}, rearrangement~\cite{batra_arxiv20,weihs_arxiv21},
language-guided navigation~\cite{anderson_cvpr18,krantz_eccv20}
and interaction~\cite{alfred}, question answering~\cite{embodiedqa,eqa_matterport,das2020probing,das_phd_thesis_2020,eqa_multitarget}
-- involve some flavor of this visual exploration.
With careful reward engineering, reinforcement learning (RL) approaches to these tasks have achieved
commendable success~\cite{wijmans_iclr20,ye_corl20,ye_iccv21,thda_iccv21,chaplot_neurips20}.
However, engineering the `right' reward function so that the learned policy
exhibits desired behavior is unintuitive and frustrating (even for domain experts),
expensive (requiring multiple rounds of retraining under different rewards), and
not scalable to new tasks or behaviors.
For complex tasks (\eg object rearrangement or tasks specified in
open-ended natural language), RL from scratch may not even get off the ground.

In this work, we advance the alternative research agenda of imitation learning~\cite{schaal_neurips96} --
\ie collecting a large dataset of human demonstrations (that implicitly capture
intelligent behavior we wish to impart to our agents) and learning policies
directly from these human demonstrations.

First, we develop a safe scalable virtual teleoperation data-collection infrastructure --
connecting the Habitat simulator running in a browser to Amazon Mechanical Turk (AMT).
We develop this in way that enables collecting human demonstrations for a variety of
tasks being studied within the Habitat~\cite{savva_iccv19, andrew_neurips21}
ecosystem (\eg PointNav~\cite{anderson_arxiv18}, \objnav~\cite{savva2017minos,anderson_arxiv18},
ImageNav~\cite{zhu_icra17}, VLN-CE~\cite{krantz_eccv20}, MultiON~\cite{wani_neurips20},~\etc).

We use this infrastructure to collect human demonstration datasets for $2$ tasks
requiring visual search -- 1) \objnavfull (\eg \myquote{find \& go to a chair}) and
2) \pickplace (\eg \myquote{find mug, pick mug, find counter, place on counter}).
In total we collect $92k$ human demonstrations, $80k$ demonstrations for \objnav and $12k$ demonstrations for
%We are collecting $100k$ demonstrations for \objnav and $12k$ demonstrations for
\pickplace.
In contrast, the largest existing datasets have $3$-$10k$ human demonstrations
in simulation~\cite{padmakumar_arxiv21,thomason_corl20,hahn_arxiv20}
or on real robots~\cite{ajay_corl21,ebert2021bridge},
an order of magnitude smaller.
This virtual teleoperation data contains $29.3M$ actions, which
is equivalent to $22,600$ hours of real-world teleoperation time
assuming a LoCoBot motion model from~\cite{krantz_iccv21}
(details in appendix (Sec.~\ref{sec:motion_model})).
The first thing this data provides is a `human baseline' with sufficiently tight
error-bars to be taken seriously. On the \objnav validation split, humans
achieve $93.7${\scriptsize $\pm0.1$}$\%$ success and $42.5${\scriptsize $\pm0.5$}$\%$
Success Weighted by Path Length (SPL)~\cite{anderson_arxiv18}
(\vs $34.6\%$ success and $7.9\%$ SPL for the 2021 Habitat ObjectNav Challenge winner~\cite{ye_iccv21}).
The success rate ($93.7\%$) suggests that this task is largely doable for humans (but not $100\%$).
The SPL ($42.5\%$) suggests that even humans need to explore significantly.

Beyond scale, the data is also rich and diverse in the strategies that humans
use to solve the tasks. \figref{fig:teaser} shows an example trajectory of an
AMT user controlling a LoCoBot looking for a `plant' in a new house -- notice
the peeking into rooms, looping around the dining table -- all of which is
(understandably) absent from the shortest path to the goal.

We use this data to answer the question -- how does large-scale imitation learning
(IL) (which has not been hitherto possible) compare to large-scale reinforcement
learning (RL) (which is the status quo)?
On \objnav, we find that IL (with no bells or whistles) using only $70k$ human
demonstrations outperforms %comparable to
RL using $240k$ agent-gathered trajectories.
This effectively establishes an `exchange rate' -- a single human demonstration
appears to be worth ${\sim}4$ agent-gathered ones.
More importantly, we find the IL-trained agent learns \emph{efficient object-search
behavior} -- as shown in \figref{fig:teaser} and Sec.~\ref{sec:analysis}.
The IL agent learns to mimic human behavior of peeking into rooms, checking corners
for small objects, turning in place to get a panoramic view -- none of these
are exhibited as prominently by the RL agent.
Finally, the accuracy~\vs training-data-size plot (\figref{fig:teaser}b) %\figref{fig:success_vs_dataset_size})
shows promising scaling behavior, suggesting that simply collecting more demonstrations
is likely to advance the state of art further.
On \pickplace, the comparison is even starker -- IL-agents achieve ${\sim}18\%$
success on episodes with new object-receptacle locations when trained
with $9.5k$ human demonstrations, while RL agents fail to get beyond $0\%$.

On both tasks, we find that demonstrations from humans are essential; imitating
shortest paths from an oracle produces neither accuracy nor
the strategic search behavior. In hindsight, this is perfectly understandable --
shortest paths (\eg \figref{fig:teaser}(a3)) do not contain any exploration
but the task requires the agent to explore.
Essentially, a shortest path is inimitable, but imitation learning is invaluable.
%One key contribution of our experiments is to disentangle the inimitability of
%shortest-path oracles from the value of imitation learning.
Overall, our work provides compelling evidence
for investing in large-scale imitation learning of human demonstrations.

%% file: sections/main/related.tex
\section{Related work}

\textbf{Embodied Demonstrations from Humans}.
Prior expert demonstration datasets for embodied tasks combining vision and action
(and optionally language) can be broadly categorized into either consisting of
shortest-path trajectories from a planner with privileged
information~\cite{alfred,embodiedqa,gordon_cvpr18,anderson_cvpr18},
or consisting of human-provided trajectories~\cite{thomason_corl20,hahn_arxiv20,padmakumar_arxiv21}.
While some works in the former collect natural language data from
humans~\cite{alfred,anderson_cvpr18}, we contend that collecting navigation data
from humans is equally crucial.
%
% Commented by - Ram
% As we demonstrate in our experiments, an agent
% trained on shortest path trajectories generalizes poorly to unseen
% environments at evaluation time.
%
%
Datasets with human-provided navigation
trajectories are typically small. TEACh~\cite{padmakumar_arxiv21}, CVDN~\cite{thomason_corl20}
and WAY~\cite{hahn_arxiv20} have ${<}10k$ episodes, while the EmbodiedQA~\cite{embodiedqa}
dataset has ${\sim}700$ human-provided episodes -- all prohibitively
small for training proficient agents.
A key contribution of our work is a scalable web-based infrastructure
%, building on work by Newman~\etal~\cite{newman2020optimal},
for collecting human navigation
and interaction demonstrations, that is easily extensible to \emph{any} task
situated in the Habitat~\cite{savva_iccv19} simulator, including language-based
tasks. We have collected ${\sim}13$x more demonstrations (in total $92k$) compared to prior publicly available datasets.
In a similar vein, Abramson~\etal~\cite{abramson2020imitating}
study large-scale imitation learning on ${\sim}600k$
human demonstrations, but their dataset is not publicly available
and the environments used lack visual realism compared to Matterport3D~\cite{mp3d}.
%\update{($80k+$)} so far and
%are scaling up to $100k$.
% \TODO{cite bc-z \cite{jang2021bc}, bridge dataset \cite{ebert2021bridge}, check slack}.

\textbf{Exploration}.
Learning how to explore an environment to gather sufficient information for use
in downstream tasks has a rich history~\cite{schmidhuber_curiosity91}.
Curiosity-based approaches typically use reinforcement learning to maximize
intrinsic rewards that capture the surprise or state prediction error of the
agent~\cite{stadie_arxiv15,pathak_icml17,burda_arxiv18}.
State visitation count rewards are also popular for learning
exploration~\cite{bellemare_neurips16,tang_neurips17}.
We refer the reader to Ramakrishnan~\etal~\cite{ramakrishnan_arxiv20} for a
review of exploration objectives for embodied agents.
% Learning how to explore an environment is an open problem in embodied navigation
% for building agents that are flexible to deploy, as they can use knowledge from previously explored environments to quickly gather information in new ones.
% This allows us to apply these agents in various downstream tasks in new environments.
%
For improving exploration in \objnav specifically, SemExp~\cite{chaplot_neurips20}
made use of a modular policy for semantic mapping and path planning, Ye~\etal~\cite{ye_iccv21}
used time-decaying state visitation count reward, and Maksymets~\etal~\cite{thda_iccv21}
used area coverage reward.
%
% , building modular policies for semantic exploration \cite{chaplot_neurips20}, and SLAM based approaches \cite{chaplot_iclr20, chen_iclr19}.

Most relatedly, Chen~\etal~\cite{chen_iclr19} used ${\sim}700$ human navigation
trajectories from the EmbodiedQA dataset~\cite{embodiedqa} (ignoring the questions)
to learn task-independent exploration using imitation learning.
We likewise train agents via imitation learning on human demonstrations,
%we develop the infrastructure to scale this up to $100k+$ demonstrations,
%and
but rather than encouraging task-agnostic exploration, we consider human demonstrations
to be a rich \emph{task-specific} mix of exploration and efficient navigation,
that simple architectures without explicit mapping and planning modules
can be trained on.

%% file: sections/main/dataset.tex
\section{Habitat-WebGL Infrastructure}
\label{sec:web}

\begin{figure}[t]
    \centering
    \includegraphics[width=0.9\linewidth]{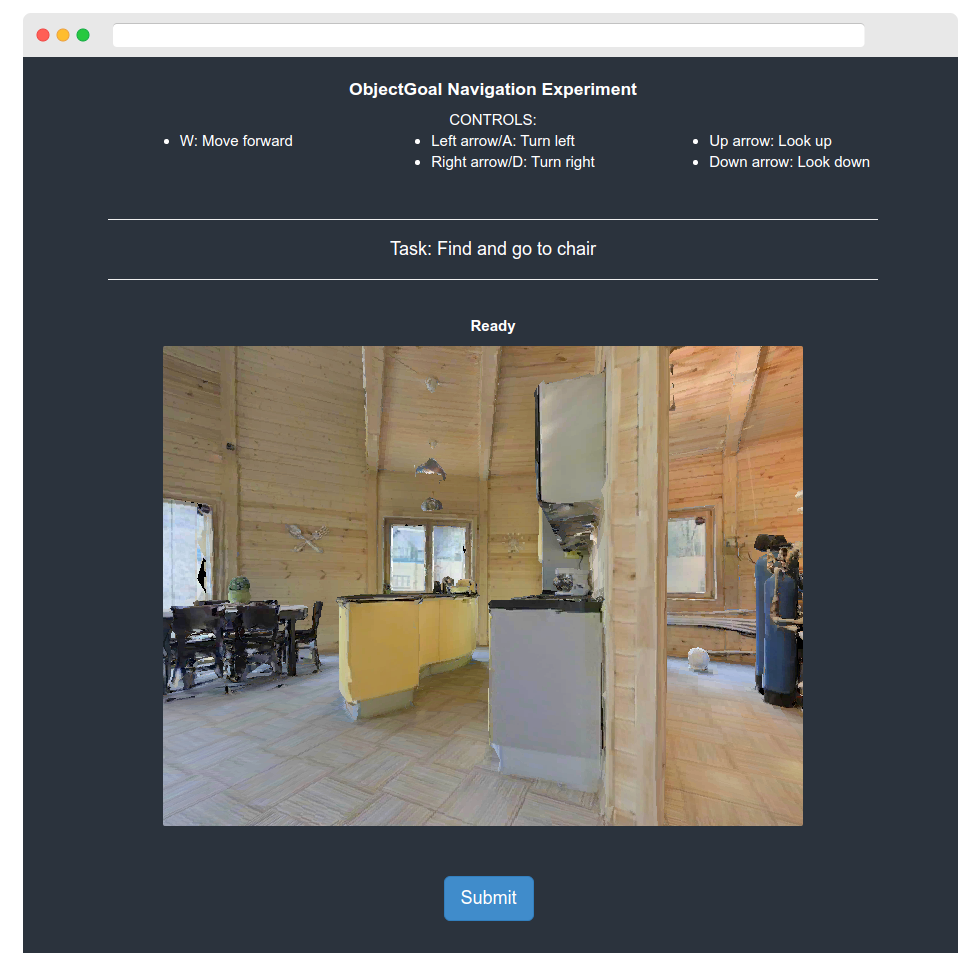}
    \vspace{5pt}
    \caption{Screenshot of our Amazon Mechanical Turk interface for collecting
        \objnav demonstrations. Users are provided the agent's first-person view
        of the environment and an instruction such as ``Find and go to chair''.
        They can make the agent look around and move in the environment via keyboard
        controls, and can submit the task upon successful navigation by clicking `Submit'.}
    \label{fig:amt_interface}
    \vspace{-15pt}
\end{figure}

To be able to train agents via imitation learning on human demonstrations,
we first need a reliable pipeline to collect human demonstrations at scale.
To this end,
we develop a web-based setup to connect the Habitat simulator\cite{savva_iccv19, andrew_neurips21}
to AMT users, building on the work of Newman~\etal~\cite{newman2020optimal}.

\textbf{Interface}.
\figref{fig:amt_interface} shows a screenshot of the interface an AMT user
interacts with to complete a data collection task.
This web application renders assets from Habitat-Sim
running on the user's browser via WebGL.
All data collection in this work was done in Matterport3D~\cite{mp3d} scans,
but any Habitat-compatible asset may be used in future.
Users can see the agent's first-person RGB view, and can move around and
grab / release objects using keyboard controls.
On the task page, users are provided an instruction and details about
keyboard controls to complete the task.
For \objnav, we provide an instruction of the form \myquote{Find and go to
the \goal}.
%
% When training, the agent only has access to goal object category ID and not the templated natural language instruction.
%
For tasks requiring interaction with objects (\eg \pickplace), we highlight the object
under the user's gaze by drawing a $3$D bounding box around it (pointed to by a crosshair
as in video games).
% Commented by - Ram
% we display a
% $3$D bounding box around the object at the center of the user's screen.
%
In our initial pilots, we found this to improve user experience when grabbing
objects instead of users having to guess when objects are available to be picked up.
When an object is successfully grabbed, it disappears from the first-person view
and immediately appears in the `inventory' area on the task interface.
When a grabbed object is released, it is dropped at the center of the user's
screen where the crosshair would be pointing to.
If the crosshair points to a distance, the object is dropped on the floor
from a height at a distance of $1m$ from the agent's location.
Upon completion, users submit the task by clicking `Submit'.
At this point, the sequence of keyboard actions, agent, and object states are
recorded in our backend server.

\textbf{Habitat simulator and PsiTurk}.
Our Habitat-WebGL application is developed in Javascript, and allows us to
access all C++ simulator APIs through Javascript bindings.
This lets us use the full set of simulation features available in Habitat.
To simulate physics, we use the physics APIs from Habitat $2.0$~\cite{andrew_neurips21},
including rigid body dynamics support (C++ APIs exposed as Javascript bindings).
%
% The web interface runs at $20$ frames per second, executing actions in the simulator
% at a frequency of $50$Hz.
Our interface executes actions entered by users every $50$ms (rendering $20$ frames per second)
and then steps physics for $50$ms in the simulator.
All of our tasks on AMT are served using PsiTurk and an NGINX reverse proxy, and
all data stored in a MySQL database.
%
% Commented by - Ram
% For each completed task, this includes the actions taken, and the agent and object
% states at every step.
%
We use PsiTurk to manage the tasks as it provides us with useful helper functions
to log task-related metadata, as well as launch and approve tasks.

See~\secref{sec:amt_interface_details} for details on how we validate human-submitted AMT tasks
and ensure data quality.

\section{Tasks and Datasets}
\label{sec:dataset}

\begin{figure*}[t]
    \vspace{-20pt}
    \centering
    \begin{minipage}[b]{0.36\textwidth}
        \resizebox{0.47\textwidth}{!}{
            \begin{minipage}[b]{0.9\textwidth}
                \begin{tabular}{@{}llrrcrr@{}}
                    \toprule
                    & & \multicolumn{2}{c}{\textsc{ObjectNav}} & & \multicolumn{2}{c}{\textsc{Pick-And-Place}} \\
                    \cline{3-4} \cline{6-7} \\[0.02in]
                    & & Human & Shortest Path &
                        & Human & Shortest Path \\
                    \midrule
                    \rownumber & Total Episodes & $80,217$ & $114,165$ & & $11,955$ & $25,747$ \\
                    \rownumber & Success & $88.9\%$ & $100.0\%$ & & $86.3\%$ & $100.0\%$ \\
                    \rownumber & SPL & $39.9\%$ & $94.9\%$ & & $21.2\%$ & $90.9\%$ \\
                    \rownumber & Occupancy coverage & $17.9\%$ & $4.6\%$ & & $26.5\%$ & $9.2\%$ \\
                    \rownumber & Sight coverage & $67.7\%$ & $33.2\%$ & & $70.3\%$ & $42.5\%$ \\
                    \bottomrule
                    \\[3pt]
		& & & \hspace{0pt} {\Large \textbf{(a)}}
                \end{tabular}
            \end{minipage}
        }
    \end{minipage}
    \hfill
    \begin{minipage}[b]{0.14\textwidth}
        \begin{subfigure}{\textwidth}
            \centering
            \includegraphics[width=\textwidth]{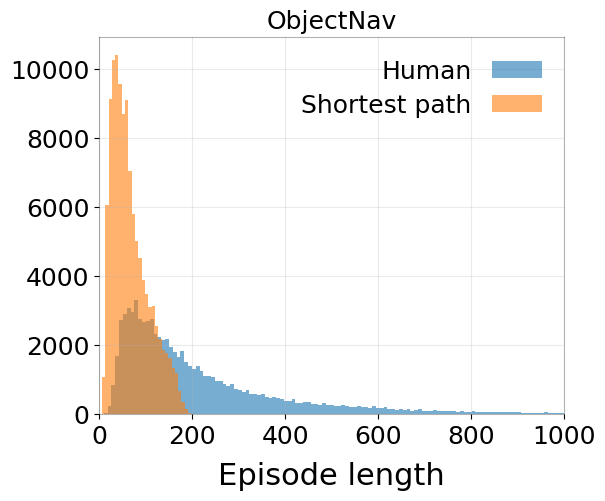}
            {\scriptsize \textbf{(b)}}
        \end{subfigure}
    \end{minipage}
    \hfill
    % \begin{minipage}[b]{0.12\textwidth}
    %     \begin{subfigure}{\textwidth}
    %         \centering
    %         \includegraphics[width=\textwidth]{figures/dataset_stats/pick_and_place_episode_length_dist.png}
    %     \end{subfigure}
    % \end{minipage}
    % \hfill
    % \begin{minipage}[b]{0.43\textwidth}
    %     \begin{subfigure}{\textwidth}
    %         \centering
    %         \includegraphics[width=\textwidth]{figures/dataset_stats/action_histogram.png}
    %     \end{subfigure}
    % \end{minipage}
    \begin{minipage}[b]{0.49\textwidth}
        \begin{subfigure}{\textwidth}
            \centering
            \includegraphics[width=\textwidth]{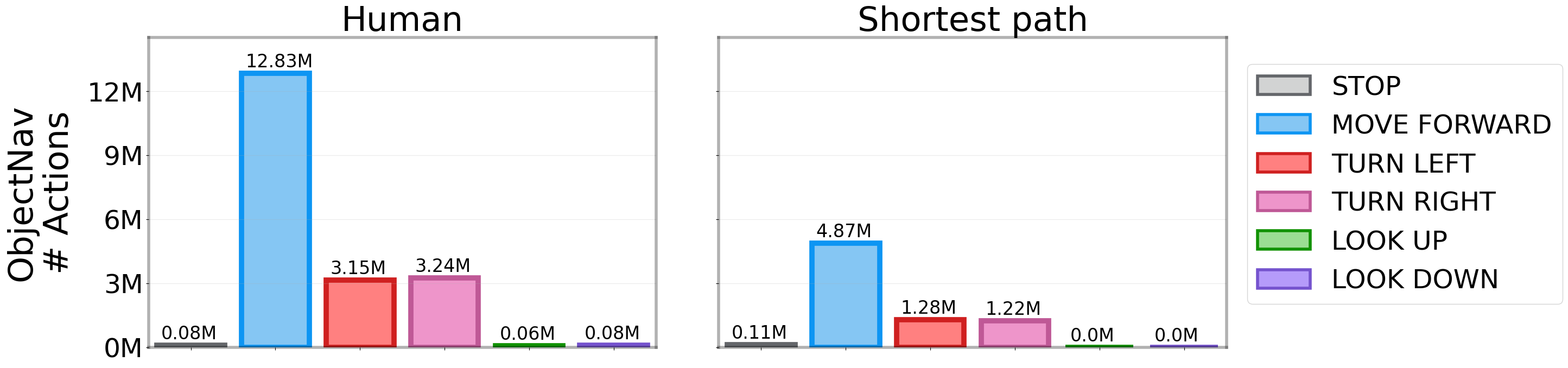}
            {\scriptsize \textbf{(c)}}
        \end{subfigure}
    \end{minipage}
    % \vspace{5pt}
    \caption{(a) Dataset statistics for human demonstrations~\vs shortest paths
        for \objnav and \pickplace. Coverage metrics are computed on subset of
        $1000$ episodes. (b) Comparison of episode lengths and action histograms
        for human demonstrations~\vs shortest paths.
        Human demonstrations are longer and have a more uniform action distribution
        than shortest paths.}
    \label{fig:dataset}
    \vspace{-20pt}
\end{figure*}

Using our web infrastructure, we collect demonstration datasets for two embodied
tasks -- \objnav~\cite{savva2017minos,anderson_arxiv18} and \pickplace, an
instantiation of object rearrangement~\cite{batra_arxiv20}.

\subsection{\objnavfull}
In the \objnavfull (\objnav) task, an agent is tasked with navigating to an
instance of a specified object category (\eg `chair') in an unseen environment.
The agent does not have access to a map of the environment and must navigate
using an RGBD camera and a GPS+Compass sensor which provides
location and orientation information relative to the start of the episode.
The agent also receives the goal object category ID as input.
The full action space is discrete and consists of \moveforward ($0.25m$),
\turnleft ($30^{\circ}$), \turnright ($30^{\circ}$), \lookup ($30^{\circ}$),
\lookdown ($30^{\circ}$), and \stopac actions. For the episode to be considered
successful, the agent must stop within $1m$ Euclidean distance of the
goal object within a maximum of $500$ steps and be able
to turn to view the object from that end position~\cite{objectnav_tech_report}.

\textbf{Human Demonstrations (\objnavhd)}. We collect $70k$ demonstrations on the $56$
training scenes from Matterport3D~\cite{mp3d} following the standard splits defined
in~\cite{mp3d,anderson_arxiv18}.
For each scene, we collect ${\sim}59$ demonstration episodes for each unique goal object
category with a randomly set start location of the human demonstrator for each episode.
This amounts to an average of ${\sim}1250$ demonstrations per scene.
Additionally, we collect $10k$ demonstrations on $25$ training scenes from Gibson~\cite{xia_cvpr18}.
For each Gibson scene, we collect ${\sim}66$ demonstration episodes for each unique goal object category.
This amounts to ${\sim}396$ demonstrations per scene.
Similar to when training artificial agents, humans can view
first-person RGB on the task interface, but unlike artificial agents, humans do
not get access to Depth and GPS+Compass.
We assume humans are sufficiently proficient at inferring depth and odometry from
vision, to the extent required to accomplish the goal.
%vision alone.
%
% In total, so far, we have collected \update{$50k$} \objnav demonstrations amounting to
In total, we collect $80k$ \objnav demonstrations amounting to
${\sim}19.5M$ steps of experience, each episode averaging
$243$ steps.
%
%We are in the process of scaling this up to $100k$ demonstration episodes ($= 26.2$
%million steps).

\textbf{Shortest Path Demonstrations}.
To compare against prior embodied datasets of shortest
paths~\cite{alfred,embodiedqa,gordon_cvpr18,anderson_cvpr18} and to demonstrate
the unique advantage of human demonstrations, we also generate a dataset of shortest paths.
The analysis in this section was performed on a subset of $35$k demonstrations of \objnavhd
(collected in first phase).
%
% Commented by - Ram
%shortest path demonstrations dataset.
%
These demonstrations are generated by greedily fitting actions to follow the
geodesic shortest path to the nearest navigable goal object viewpoint.
Since shortest paths are (by design) shorter than human demonstrations
(average $67$~\vs$243$ steps per demonstration), we compensate by
generating a larger number of shortest paths to roughly match the steps
with $35k$ human demonstrations ($7.6M$ steps from $114k$ shortest paths~\vs $8.4M$
steps from $35k$ human demonstrations).
%
% We generated $114k$ shortest path demonstrations to match steps in experience
% with human demonstrations for \objnav, each averaging $67$ steps, amounting to
% a total of ${\sim}7.6$ million steps of experience.

\textbf{Analysis}. Table~\ref{fig:dataset}a reports statistics of
our human and shortest path demonstration datasets.
%
% We report the two standard evaluation metrics for \objnav, Success and Success Weighted by Path Length (SPL) for both datasets.
%
Recall that an episode is considered a failure if the target object is not found
within $500$ navigation steps. Under this definition, humans fail on $11.1\%$
training set episodes; they fail on $0\%$ episodes if we relax the step-limit.
Surprisingly, SPL for humans is $39.9\%$ for training split episodes, significantly
lower than $94.9\%$ for shortest paths
% \NOTE{Ram: This could be due to some implementation detail related to shortest path follower, I haven't had a chance to debug this more closely. Not sure how to frame this here.},
underscoring the difficulty in searching for objects in in unseen environments.

%Commented by - Ram
% following close-to-shortest-path trajectories
% in unseen environments to solve this task.
%
%
% have perfect success and SPL as they don't have access to goal location and
% they don't necessarily follow the shortest path to goal object when solving the task.
%
We additionally report two metrics to demonstrate that the \objnav task requires
significant exploration. Occupancy Coverage (OC) measures percentage of total area
covered by the agent when navigating. To compute OC, we first divide the map into
voxel grids of $2.5m \times 2.5m \times 2.5m$ and increment a counter for each
visited voxel.
Sight Coverage (SC) measures the percentage of total navigable area visible to
the agent in its field of view (FOV) during an episode. To compute SC, we project
a mask on the top-down map of the environment using the agent's FOV, that is
iteratively updated at every step to update the area seen by the agent.
OC and SC metrics for human demonstrations show that humans traverse $3\text{-}4$x
and observe $2$x the area of the environment when performing this task
compared to shortest paths.

\figref{fig:dataset}b,c show episode length and action histograms for human and
shortest path demonstrations. Human demonstrations are longer (average ${\sim}243$~\vs
${\sim}67$ steps per demonstration) and have a slightly more uniform action distribution.

\subsection{Object Rearrangement -- \pickplace}

% To test generalization of behavior cloning for embodied tasks, we introduce the \pickplace task which require visual exploration and interaction in environment. We set up \pickplace task to test behavior cloned
% agent's ability to search for objects, interact with them, and ability to understand and executing a natural language instruction.
In the pick-and-place task (\pickplace), an agent must follow an instruction
of the form \myquote{Place the {\tt <object>} on the {\tt <receptacle>}},
without being told the location of the {\tt <object>} or {\tt <receptacle>} in a new environment.
The agent must explore and navigate to the object, pick it up, explore and
navigate to the receptacle, and place the previously picked-up object on it.
Similar to \objnav, agents are not equipped with a map of the environment, and
only have access to an RGBD camera and a GPS+compass sensor.
At a high level, \pickplace can be thought of as a natural extension of \objnav,
performing it twice in the same episode -- once to find the specified object and
again to find the specified receptacle -- delimited by grab and release actions.
For object interaction, we use the `magic pointer' abstraction defined in~\cite{batra_arxiv20}.
If the agent is not holding any object, the grab/release action will pick the object pointed
to by its crosshair (at the center of its viewpoint) if within $1.5m$ of the object.
If the agent is already holding an object, the grab/release action will drop the
object at the crosshair location.
If there is no drop-off point within $1.5m$ in the direction of the crosshair,
the object will be dropped on the floor $1m$ in front of the agent.
The full action space is discrete and consists of \moveforward ($0.15m$), \movebackward ($0.15m$),
\turnleft ($5^{\circ}$), \turnright ($5^{\circ}$), \lookup ($5^{\circ}$), \lookdown ($5^{\circ}$),
\grabrelease, \noop (step physics $50ms$), and \stopac.
For the episode to be considered successful, the agent must place the object on
top of the receptacle -- ~\ie the object center should be at a height greater than
the receptacle center, and within $0.7m$ of the receptacle object center -- within $1500$ steps.
We picked this $0.7m$ threshold distance between the object and receptacle based on pilots on AMT.
$0.7m$ was sufficiently strict for avoiding false positives in the collected demonstrations
where users are able to submit the task without necessarily placing the object on top
of the receptacle.
% for humans to place objects on top of the receptacle.
% %
% when collecting \pickplace demonstrations from humans

% make the task validation stricter
% when collecting data from humans, this check gets us trajectories where humans have to place objects on top of the receptacles to succeed in the task.}

\textbf{Human Demonstrations (\pickplacehd)}.
We collect human demonstrations for \pickplace on $9$ scenes from
Matterport3D~\cite{mp3d}.
In each episode, objects and receptacles are instantiated by randomly sampling
from $457$ possible object-receptacle pairs.  We initialize the object and
receptacle at randomly sampled locations in the environment, and collect
$3$ demonstrations for each object-receptacle pair.
The agent, object, and receptacle locations are randomized across all episodes
(including the $3$ we collect for each object-receptacle pair).
In total, we have $457 \times 3$ unique object-receptacle-agent position
initializations per scene, amounting to $457 \times 3 \times 9 = {\sim}12k$
demonstrations, which is ${\sim}11.5$M steps in experience, each episode
averaging $932$ steps.

\textbf{Shortest Path Demonstrations}. Similar to \objnav, we generate shortest path
demonstrations for \pickplace.
These demonstrations are generated by first using the geodesic shortest-path follower
% greedily fit actions to follow the geodesic shortest path
to the object, then using a heuristic action planner to face and pick up the object,
then following the geodesic shortest-path to the receptacle, and again using
a heuristic action planner to drop the object on the receptacle.
We generated $25.7$k shortest path demonstrations for \pickplace, each averaging
$342$ steps, amounting to a total of ${\sim}8.8$ million steps of experience.

\textbf{Analysis}. Table~\ref{fig:dataset}a reports statistics for human
and shortest path demonstrations.
Similar to \objnav, humans have significantly lower SPL, and $2$x higher occupancy
and sight coverage compared to shortest paths, suggesting the need for exploration.
%
% This demonstrates that humans explore more when solving a task allowing our agents get a higher visibility during training.
Comparing episode lengths and action histograms (see appendix (Sec.~\ref{sec:pick_place_stats}) for figure), % \figref{fig:dataset}b),
% shows a comparison of episode length and action histograms for both datasets.
human demonstrations are longer and make use of all $9$ actions.
% In contrast, shortest path demonstrations have shorter trajectories and uses only
% $5$ actions to complete the task.
Interestingly, humans often use the \movebackward action to backtrack,
which the shortest path agents do not use (by design),
instead of turning $180^{\circ}$ and moving forward.
This behavior does not appear in \objnav shortest path demonstrations because
there is just one target object, and so the geodesic shortest path would never
involve backtracking or making $180^{\circ}$ turns.
% we limit the action space on the interface to match \objnav action space.
% this behavior is unique to human trajectories and we would have to
% build heuristics in our shortest path agents to introduce this behavior.

% \begin{figure}[t]
%     \resizebox{1\linewidth}{!}{
%         \includegraphics[width=\linewidth]{figures/dataset_stats/episode_length_dist.png}
%         % \label{fig:dataset1}
%     }
%     \vspace{1pt}
%     \resizebox{1\linewidth}{!}{
%         \includegraphics[width=\linewidth]{figures/dataset_stats/action_histogram.png}
%         % \label{fig:dataset3}
%     }
%     \quad
%     \vspace{5pt}
%     \caption{Comparison of episode length and action histograms of human demonstrations with shortest path demonstrations for \objnav and \pickplace. Notice, the human demonstrations have longer trajectories on average and use all actions solving a task. The shortest path demonstrations have shorter trajectories on average and doesn't need to use all actions to solve the task.}
%     \label{fig:dataset}
% \end{figure}

    % \TODO{How do we simulate physics? Bullet?}
    % \TODO{At what framerate do we run Habitat on the client's browser?}
    % \TODO{Describe validation checks}
    % \TODO{Add more details of the infrastructure.}

% \end{compactitem}

%% file: sections/main/approach.tex
\section{Imitation Learning from Demonstrations}

\begin{figure*}[t]
    \centering
    \begin{minipage}[b]{0.3\textwidth}
        \begin{subfigure}{\textwidth}
            \includegraphics[width=\linewidth]{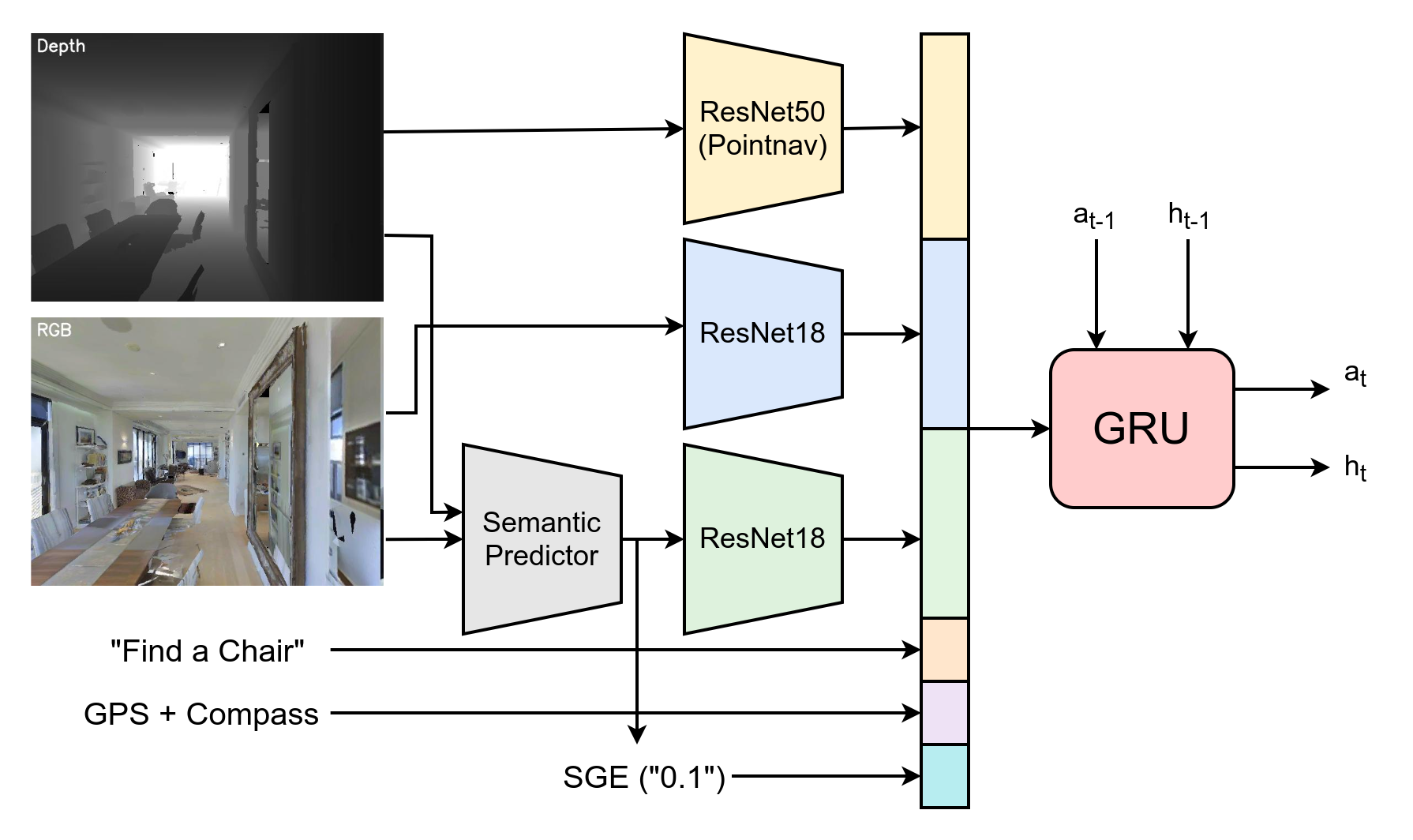}
            \caption{\objnav architecture}
            \label{fig:architecture_a}
        \end{subfigure}
    \end{minipage}
    %\hfill
    %\hspace{5pt}
    \begin{minipage}[b]{0.25\textwidth}
        \begin{subfigure}{\linewidth}
            \includegraphics[width=\linewidth]{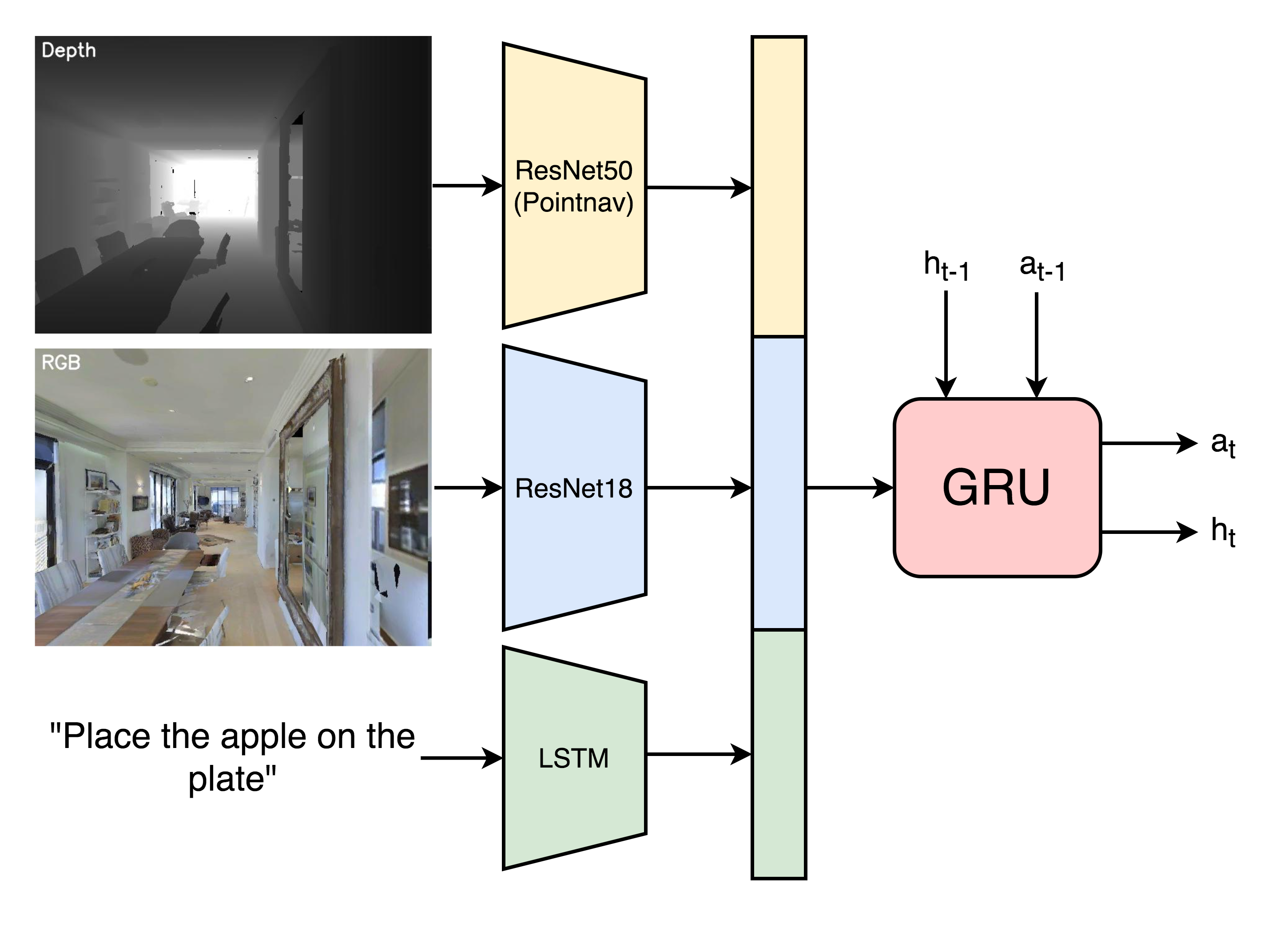}
            \caption{\pickplace architecture}
            \label{fig:architecture_b}
        \end{subfigure}
    \end{minipage}
    %\hfill
    %\hspace{-20pt}
    \begin{minipage}[b]{0.4\textwidth}
        \resizebox{0.55\textwidth}{!}{
            \begin{minipage}[b]{0.9\textwidth}
                \begin{tabular}{@{}llrr@{}}
                    \toprule
                     & Method & Success $(\mathbf{\uparrow})$ & SPL $(\mathbf{\uparrow})$ \\
                    \midrule
                    \rownumber & RL (ExploreTillSeen)~\cite{thda_iccv21} & $20.0\%$ & $6.5\%$ \\
                    \rownumber & RL (ExploreTillSeen + THDA)~\cite{thda_iccv21} & $28.4\%$ & $\mathbf{11.0}\%$ \\
                    \rownumber & RL (Red Rabbit)~\cite{ye_iccv21} & $34.6\%$ & $7.9\%$ \\
                    \rownumber & RL (EmbCLIP)~\cite{khandelwal2021simple} & $21.6\%$ & $8.7\%$ \\
                    \midrule
                    \rownumber & IL w/ Shortest Paths & $4.4\%$ & $2.2\%$ \\
                    \rownumber & IL w/ $35k$ Human Demos (similar $\#$steps as row 4) & $31.6\%$ & $8.5\%$ \\
                    \rownumber & IL w/ $50k$ Human Demos & $32.4\%$ & $9.1\%$ \\
                    \rownumber & IL w/ $50k$ Human Demos (includes $10k$ THDA~\cite{thda_iccv21}) & $33.2\%$ & $9.5\%$ \\
                    \rownumber & IL w/ $70k$ Human Demos & $\mathbf{35.4}\%$ & $10.2\%$ \\
                    \rownumber & IL w/ $70k$ Human Demos (includes $10k$ Gibson~\cite{xia_cvpr18}) & $33.9\%$ & $9.7\%$ \\
                    \rownumber & IL w/ $80k$ Human Demos & $33.8\%$ & $9.9\%$ \\
                    \midrule
                    \rownumber & Humans & $93.7\%$ & $42.5\%$ \\
                    \bottomrule
                    \\[5pt]
		& \hspace{0pt} {\large \textbf{(c)} \objnav results on MP3D-\textsc{val}}
                \end{tabular}
                % \caption{}
                % \label{tab:objnav_comparison}
            \end{minipage}
        }
    \end{minipage}
    \vspace{5pt}
    \caption{Our policy architectures for a) \objnav and b) \pickplace.
    Both are simple CNN+RNN networks that embed and concatenate all sensory inputs,
    which are then fed into a GRU to predict actions.
    c) \objnav results on the MP3D \textsc{val} split~\cite{anderson_arxiv18,mp3d}.}
    \label{fig:combined_architecture_val_results}
    \vspace{-15pt}
\end{figure*}

% \begin{figure*}[t]
%     \centering

%     % \quad

%     \vspace{5pt}
%     \caption{}
%     \label{fig:architecture}
%     \vspace{-15pt}
% \end{figure*}

% \begin{table}[h]
%     \centering
%     \resizebox{0.975\columnwidth}{!}{
%         \begin{tabular}{@{}llrr@{}}
%             \toprule
%              & Method & Success $(\mathbf{\uparrow})$ & SPL $(\mathbf{\uparrow})$ \\
%             \midrule
%             \rownumber & RL (ExploreTillSeen)~\cite{thda_iccv21} & $20.0\%$ & $6.5\%$ \\
%             \rownumber & RL (ExploreTillSeen + THDA)~\cite{thda_iccv21} & $28.4\%$ & $11.0\%$ \\
%             \rownumber & RL (Red Rabbit)~\cite{ye_iccv21} & $34.6\%$ & $7.9\%$ \\
%             \rownumber & IL w/ Shortest Paths & $4.4\%$ & $2.2\%$ \\
%             \rownumber & IL w/ $35k$ Human Demos (\objnav) & $31.6\%$ & $8.5\%$ \\
%             \rownumber & IL w/ $40k$ Human Demos (\objnav $+$ THDA~\cite{thda_iccv21}) & $30.4\%$ & $8.8\%$ \\
%             % \rownumber Ours w. GT segmentation & $00.0\%$ & $0.0\%$ \\
%             \rownumber & Humans & $93.7\%$ & $42.5\%$ \\
%             \bottomrule
%         \end{tabular}
%     }
%     \vspace{5pt}
%     \caption{ObjectNav results on the MP3D
%         \textsc{val} split~\cite{anderson_arxiv18,mp3d}.}
%     \label{tab:objnav_comparison}
%     \vspace{-10pt}
% \end{table}

We use behavior cloning to learn a policy from demonstrations.
% Commented by - Ram
% Behavior cloning learns an offline policy from state-action trajectories via
% supervised learning.
%
Let $\pi_\theta(a_t \,|\, o_t)$ denote a policy parametrized by $\theta$ that maps
observations $o_t$ to a distribution over actions $a_t$.
% Specifically, the policy \NOTE{$\pi_{\theta}$} with parameters \NOTE{$\theta$} maps an observation
% $o_{t-1}$ to a probability distribution over actions $a_{t}$.
%
% Let \tau denote a trajectory consisting of state, observation, human action tuples: $\tau = \big(s_0, o_0, a_0, \ldots, s_T, o_T, a_T\big)$ and $\Tau = \big\{ \tau^(i) \big \}_{i=1}^N$ denote a dataset of human demonstrations.
%
Let $\tau$ denote a trajectory consisting of state, observation, human action tuples:
$\tau = \big(s_0, o_0, a_0, \ldots, s_T, o_T, a_T\big)$ and $\Tau = \big\{ \tau^{(i)} \big \}_{i=1}^N$
denote a dataset of human demonstrations.
The learning problem can be summarized as:

% Commented by - Ram
% Given a dataset of $N$ trajectories $\{T_n\}_{n=1}^N$, each consisting of
% observation-action transition tuples $(o_{t1}, a_{t}) \in T_n$, the learning
% objective can be summarized as:

\vspace{-15pt}
\begin{equation}
    \theta^* =
        \text{arg\,min}_\theta
            \sum_{i=1}^N
                \sum_{(o_{t}, a_{t}) \in \tau^{(i)}}
                    -\log \Big( \pi_\theta(a_{t}|o_{t}) \Big)
\end{equation}
\vspace{-5pt}

% Commented by - Ram
% \begin{equation}
%     \theta_\pi^* =
%         \text{arg\,min}_\theta
%             \sum_{n=1}^N
%                 \sum_{(o_{t-1}, a_{t}) \in T_n}
%                     -\log \left( \pi_\theta(a_{t}|o_{t-1}) \right)
% \end{equation}

% Intuitively, in behavior cloning, the policy is trained to maximize the conditional likelihood of actions from the human demonstration.

%\subsection{Inflection weighting}
\xhdr{Inflection weighting}
introduced in Wijmans~\etal~\cite{eqa_matterport},
% We perform behavior cloning with inflection weighting~\cite{eqa_matterport}. As in \cite{eqa_matterport}, inflection weighting
adjusts the loss function to upweight timesteps where actions
change (\ie $a_{t-1} \neq a_t$).
Specifically, the inflection weighting loss coefficient is computed as
total no.~of actions in the dataset divided by the total no.~of inflection
points, and this coefficient is multiplied with the loss at each inflection
timestep where $a_{t-1} \neq a_t$.
%
% This adjustment is in proportion to how
% in proportion to the rarity of
%
This approach was found to be useful for tasks like navigation with long
sequences of the same actions, \eg several `forward' actions when navigating
corridors~\cite{eqa_matterport}.
We use inflection weighting in all our experiments and found it to help
over vanilla behavior cloning.
% NOTE: Remember to add this in full draft
% \item \TODO{Explain inflection weighting better}
% \item \TODO{How do we pick the inflection weighting coefficient?}

%\subsection{Base architecture}
Our \xhdr{base policy}
is a simple CNN+RNN architecture.
We first embed all sensory inputs using feed-forward modules.
%
% We separately encode RGB and Depth observations.
For RGB, we use a randomly initialized ResNet18~\cite{he_cvpr16}.
For depth, we use a ResNet50 that was pretrained on PointGoal navigation using DD-PPO~\cite{wijmans_iclr20}.
Then these RGB and depth features (and optionally other task-specific features)
are concatenated and fed into a GRU~\cite{cho_emnlp14} to predict a distribution
over actions $a_{t+1}$.
Task-specific architectural choices over this base policy are described in the
next sections.

% \vspace{-5pt}
\subsection{\textbf{\objnav}}
% \vspace{-5pt}

\figref{fig:architecture_a} shows our \objnav architecture.
%
% In addition to the input features described above, we use semantic segmentation
% features.
%
Similar to Anand~\etal~\cite{anand_arxiv18}, we feed in RGBD inputs of size
$640 \times 480$ passed through a \avgpool layer to reduce the resolution
(performing low-pass filtering + downsampling).
%
% In addition to RGBD observations the agent has access to GPS+Compass sensor which
% provides location and orientation relative to start of the episode.
%
The agent also has a GPS+Compass sensor,
which provides location and orientation relative to start of the episode.
GPS+Compass inputs are pass through fully-connected layers to embed them to $32$-d vectors.
In addition to RGBD and GPS+Compass,
following Ye~\etal~\cite{ye_iccv21}, we use two additional semantic features --
semantic segmentation (SemSeg) of the input RGB and a `Semantic Goal Exists' (SGE)
scalar which is the fraction of the visual input occupied by the goal category.
These semantic features are computed using a pretrained and frozen
RedNet~\cite{jiang2018rednet} that was pretrained on SUN RGB-D~\cite{song2015sun}
and finetuned on $100k$ randomly sampled front-facing views rendered in the Habitat
simulator.
Finally, we also feed in the object goal category embedded into a $32$-d vector.
All of these input features are concatenated to form an observation embedding,
and fed into a $2$-layer, $512$-d GRU at every timestep.
We train this policy for ${\sim}400$M steps ($= {\sim}21$ epochs on ${\sim}70k$ demonstration episodes).
% The two primary metrics we report are success and SPL
We evaluate checkpoints at every ${\sim}15$M steps for the last $50$M steps of
training, and report metrics for checkpoints with the highest success on
the validation split.

\vspace{-3pt}
\subsection{\textbf{\pickplace}}
\vspace{-3pt}

\figref{fig:architecture_b} shows our \pickplace architecture.
We feed in RGBD inputs of size $256 \times 256$.
In addition to RGBD observations, the policy gets as input language instructions
of the form \myquote{Place the {\tt <object>} on the {\tt <receptacle>}} encoded using
a single-layer LSTM~\cite{hochreiter_nc97}.
RGBD and instruction features are concatenated to form an observation embedding,
which is fed into a $2$-layer, $512$-d GRU at every timestep.
We train this policy for ${\sim}90$M steps ($= {\sim}10$ epochs on ${\sim}9.5k$
demonstration episodes).
%
% The two primary metrics we report are success and SPL.
We evaluate checkpoints at every ${\sim}10$M steps during training, and report
metrics for checkpoints with the highest success on the validation split.

%% file: sections/main/results.tex
\section{Experiments \& Results}
\label{sec:experiments}

\subsection{\textbf{\objnav}}

% Moved to supplementary
% % % % % % % % % % % % % % % % % % % % % % % % % % % % % % % % % % % % % % % %
% Table: ObjectNav ablations on MP3D-validation
% % % % % % % % % % % % % % % % % % % % % % % % % % % % % % % % % % % % % % % %
% \begin{table}[t]
%     \centering
%     \resizebox{0.975\columnwidth}{!}{
%         \begin{tabular}{@{}lrr@{}}
%             \toprule
%             Method & Success $(\mathbf{\uparrow})$ & SPL $(\mathbf{\uparrow})$ \\
%             \midrule
%             \rownumber Ours w.o. vision & $0.0\%$ & $0.0\%$ \\
%             % \rownumber Ours w.o. inflection weighting & $00.0\%$ & $0.0\%$ \\
%             \rownumber Ours w.o. semantic segmentation & $22.7\%$ & $6.1\%$ \\
%             \rownumber Ours w. shortest paths & $4.4\%$ & $2.2\%$ \\
%             \rownumber Ours & $31.6\%$ & $8.5\%$ \\
%             \rownumber \NOTE{Ours + THDA} & \NOTE{$29.9\%$} & - \\
%             % \rownumber Ours w. GT segmentation & $00.0\%$ & $0.0\%$ \\
%             \rownumber Humans & $93.7\%$ & $42.5\%$ \\
%             \bottomrule
%             \end{tabular}
%     }
%     \vspace{5pt}
%     \caption{ObjectNav ablation results on the MP3D
%         \textsc{val} split~\cite{anderson_arxiv18,mp3d}.}
%     \label{tab:objnav_ablations}
% \end{table}

Table~\ref{fig:combined_architecture_val_results}c reports results on the MP3D \textsc{val} split
for several baselines.
First, we compare our approach with two state-of-the-art RL approaches from prior work.
Maksymets~\etal~\cite{thda_iccv21} (row $1$) train their policy using a reward structure
that breaks \objnav into two subtasks -- exploration and direct navigation to goal object
once it is spotted.
This agent gets a positive reward for maximizing area coverage until it sees the goal object.
It then receives a navigation reward to minimize distance-to-object.
This policy achieves $20.0\%$ success and $6.5\%$ SPL (row $1$).
% which is $15.4\%$ worse on success and $3.7\%$ worse on SPL compared
% to behavior cloning on $70k$ human demonstrations (row $9$).
%
\cite{thda_iccv21} then combine this reward structure with Treasure Hunt Data Augmentation (THDA) --
inserting arbitrary $3$D target objects in the scene to augment the set of training episodes.
With THDA, this achieves $28.4\%$ success and $11.0\%$ SPL (row $2$), $7.0\%$ worse and $0.8\%$ better
respectively than behavior cloning on $70k$ human demonstrations (row $9$).
Ye~\etal~\cite{ye_iccv21} (row $3$) train their policy with a combination of exploration and
distance-based navigation rewards, and their representations with several auxiliary tasks (\eg inverse dynamics and predicting map coverage).
This achieves $34.6\%$ success and $7.9\%$ SPL (row $3$),
which is $0.8\%$ worse on success and $2.3\%$ worse on SPL than
our approach (row $9$).
Khandelwal~\etal~\cite{khandelwal2021simple} (row $4$) train a policy using
CLIP~\cite{radford2021learning} as a visual backbone with simple distance-based
navigation rewards.
This achieves $21.6\%$ success and $8.7\%$ SPL (row $4$), which is $13.8\%$ worse
on success and $1.5\%$ worse on SPL than our approach (row $9$).
%
%Next, as described in~\secref{sec:dataset}, we train
IL on a dataset of shortest paths %instead of human demonstrations
%
%As a reminder, since shortest paths are shorter than human demonstrations
%(average $67$~\vs$262$ steps per demonstration), we generated a larger $114k$ demonstrations dataset of shortest paths.
%
%The policy trained on shortest paths
%this
achieves $4.4\%$ success
and $2.2\%$ SPL (row $5$), significantly worse than training on $35k$
human demonstrations ($31.6\%$ success, $8.5\%$ SPL).
Recall that comparison of shortest path demonstrations was done with a subset of $35k$ \objnavhd demonstrations
that were collected in the first phase of the project.
%
% \update{Next, we collected $10k$ demonstrations on \objnav episodes generated in Gibson \cite{Xia_2018_CVPR},
% these scenes have a smaller layout compared to MP3D\cite{mp3d}. An IL agent trained on these $10k$ combined
% with $60k$ MP3D\cite{mp3d} human demonstrations achieves $33.4\%$ success and $9.2\%$ SPL.}
%
Next, we also collected $10k$ human demonstrations on \objnav episodes generated in THDA fashion -- \ie
asking humans to find randomly inserted objects. Notice that this involves pure exhaustive search,
since there are no semantic priors that humans can leverage in this setting.
%
%these episodes do not conform to any semantic priors
% \NOTE{THDA adds $3$D object models in simulated training environments at random locations,
% these objects may be goal themselves (increasing the goal space of \objnav during training) and serve to increase scene layout diversity.}
%
An IL agent trained on $10k$ THDA demonstrations combined with the original $40k$
%these \update{$10k$ combined with the original $35k$ human}
demonstrations achieves $33.2\%$ success and $9.5\%$ SPL (row $8$)
which is $0.8\%$ better on success and $0.4\%$ better on SPL than $50k$ non-THDA demonstrations (row $7$),
\ie adding these THDA demonstrations with exhaustive search behavior helps.
We also collected $10k$ demonstrations on Gibson\cite{xia_cvpr18} \objnav episodes to compare
effect of different scene datasets. An agent trained on $10k$ Gibson demonstrations combined with $60k$ MP3D
demonstrations achieves $33.9\%$ success and $9.7\%$ SPL (row $11$),
which is $1.5\%$ worse on success and $0.5\%$ worse on SPL
compared to when we use MP3D-only demonstrations (row $9$).

%demonstrations achieves $30.4\%$ success and $8.8\%$ SPL (row $6$), \ie adding these demos hurts.
%This suggests that adding the \emph{right} kind of human demonstrations is important, though $5k$ is
%admittedly a small sample to draw conclusions from.
%
Finally, we also benchmark human performance
on the MP3D \textsc{val} split
--  $93.7\%$ success, $42.5\%$ SPL (row $12$).
\begin{table}[t]
    \centering
    \resizebox{0.95\columnwidth}{!}{
        \begin{tabular}{@{}lrr@{}}
            \toprule
            Method & Success $(\mathbf{\uparrow})$ & SPL $(\mathbf{\uparrow})$ \\
            \midrule
            \rownumber IL wo/ Vision & $0.0\%$ & $0.0\%$ \\
            % \rownumber Ours w.o. inflection weighting & $00.0\%$ & $0.0\%$ \\
            \rownumber IL wo/ Semantic Input & $22.7\%$ & $6.1\%$ \\
            \rownumber IL w/ RGBD + Semantic Input & $31.6\%$ & $8.5\%$ \\
            % \rownumber \update{IL w/ $70k$ Human Demos ($+$ THDA~\cite{thda_iccv21})} & $35.4\%$ & $10.2\%$ \\
            % \rownumber Ours w. GT segmentation & $00.0\%$ & $0.0\%$ \\
            %\rownumber Humans & $93.7\%$ & $42.5\%$ \\
            \bottomrule
            \end{tabular}
    }
    \vspace{5pt}
    \caption{ObjectNav ablation results on the MP3D
        \textsc{val} split~\cite{anderson_arxiv18,mp3d}.}
    \label{tab:objnav_ablations}
    \vspace{-20pt}
\end{table}

\textbf{ObjectNav Sensor Ablations}. Table~\ref{tab:objnav_ablations} reports results on the MP3D \textsc{val} split
for various ablations of our approach trained on $35k$ human demonstrations.
% across all of the ablation experiments we use same set of hyperparameters.
First, without any visual input (row $1$),~\ie no RGBD and semantic inputs, the agent fails
to learn anything ($0\%$ success, $0\%$ SPL).
Second, without SemSeg and SGE features (and keeping only RGB and Depth
features) to the policy, performance drops by $8.9\%$ success and $2.4\%$ SPL (row $2$~\vs $3$).
%
% \update{Next, compared to EmbCLIP \cite{khandelwal2021simple} our policy performs $1.1\%$ better on
% success and $2.6\%$ worse on SPL (row 1 \vs 2). We hypothesize that EmbCLIP \cite{khandelwal2021simple} has higher SPL
% because the policy is learned using a distance to goal reward which leads to policy learning to follow shortest path.
% }
%
% Third, we trained a policy on a dataset of
% shortest paths instead of human demonstrations. Note that since shortest paths
% are shorter than human demonstrations \update{(average $67$~\vs$243$ steps per demo)}, we compensate by
% generating a larger number of shortest paths to roughly match the steps of
% experience ($7.6M$ steps from $114k$ shortest paths~\vs $10.6M$ steps from $35k$
% human demonstrations). The policy trained on shortest paths achieves $4.4\%$ success
% and $2.2\%$ SPL (row $3$), significantly worse than training on
% human demonstrations ($31.6\%$ success, $8.5\%$ SPL).}
% %
% Finally, to get a sense for human performance, we also collected demonstrations
% on the MP3D \textsc{val} split. Humans achieve $93.7\%$ success, $42.5\%$ SPL (row $5$).

\textbf{Habitat ObjectNav Challenge Results}.
Table~\ref{tab:objnav_test} compares our results with prior approaches
from the $2020$ and $2021$ Habitat Challenge leaderboards.
Our approach (IL w/ $70k$ demonstrations) achieves $27.8\%$ success and $9.9\%$ SPL (row $8$),
% and slightly outperforms all existing state of the art -- not really :)
% comparable to prior RL-trained counterparts -- $0.1\%$ worse success, $1.8\%$ better SPL than
outperforming prior RL-trained counterparts -- $3.3\%$ better success, $3.5\%$ better SPL than
Red Rabbit (6-Act Base)~\cite{ye_iccv21} (row 5),
%
% $3.1\%$ better success and $0.1\%$ better SPL than Red Rabbit (6-Act Tether)~\cite{ye_iccv21} (row 6),}
and $6.7\%$ better success, $1.1\%$ better SPL
than ExploreTillSeen + THDA~\cite{thda_iccv21} (row 7).

% % % % % % % % % % % % % % % % % % % % % % % % % % % % % % % % % % % % % % % %
% Table: ObjectNav results (success and SPL) on Habitat Challenge
% % % % % % % % % % % % % % % % % % % % % % % % % % % % % % % % % % % % % % % %
\begin{table}[h]
    \centering
    \resizebox{0.975\columnwidth}{!}{
        \begin{tabular}{@{}llrr@{}}
            \toprule
            & Team / Method & Success $(\mathbf{\uparrow})$ & SPL $(\mathbf{\uparrow})$ \\
            \midrule
            \rownumber & DD-PPO baseline~\cite{wijmans_iclr20,ye_iccv21} & $6.2\%$ & $2.1\%$ \\
            \rownumber & Active Exploration (Pre-explore) & $8.9\%$ & $4.1\%$ \\
            \rownumber & SRCB-robot-sudoer & $14.4\%$ & $7.5\%$ \\
            \rownumber & SemExp~\cite{chaplot2020object} & $17.9\%$ & $7.1\%$ \\
            \rownumber & Red Rabbit (6-Act Base)~\cite{ye_iccv21} & $24.5\%$ & $6.4\%$ \\
            \rownumber & Red Rabbit (6-Act Tether)~\cite{ye_iccv21} & $21.1\%$ & $8.1\%$ \\
            \rownumber & ExploreTillSeen + THDA~\cite{thda_iccv21} & $21.1\%$ & $8.8\%$ \\
            \midrule
            \rownumber & IL w/ $70k$ Human Demos & $\mathbf{27.8\%}$ & $\mathbf{9.9\%}$ \\
            \bottomrule
            \end{tabular}
    }
    \vspace{5pt}
    \caption{Results on Habitat ObjectNav Challenge
        \textsc{test-std}~\cite{habitat_challenge2020}.}
    \label{tab:objnav_test}
    \vspace{-5pt}
\end{table}

% Commented by - Ram
% is competitive with
%     RL-trained counterparts -- \NOTE{$0.1\%$} worse on success, \NOTE{$1.8\%$} better on SPL than
% Red Rabbit~\cite{ye_iccv21}, and \NOTE{$3.3\%$} better on success, \NOTE{$0.6\%$} worse on SPL
% than Treasure Hunt~\cite{thda_iccv21} -- with \emph{just} ${\sim}40k$ demonstrations.
%
% Using the RedNet segmentation and SGE feature the behavior cloning agent trained on human demonstrations (row 7) reaches $20.6\%$ success and $6.8\%$ SPL on the test-standard split of Habitat Challenge 2021 leaderboard.
%
% This is $3$x the performance of the DD-PPO baseline (row 1) that is a method
% which "solved" PointNav \cite{anderson_arxiv18}.

% % % % % % % % % % % % % % % % % % % % % % % % % % % % % % % % % % % % % % % %
% Figure: No. of steps of training~\vs performance on train, val
% Figure: Dataset size~\vs performance on val
% % % % % % % % % % % % % % % % % % % % % % % % % % % % % % % % % % % % % % % %

\textbf{Performance~\vs Dataset size}.
To investigate scaling behavior, we plot \textsc{val} success against the size
of the human demonstrations dataset in \figref{fig:teaser}b.
We created splits of the human demonstrations' dataset of increasing sizes,
from $4k$ to $70k$, and trained models with the same set of hyperparameters on each split.
All hyperparameters were picked early in the course of the data collection (on the $4k$ and $12k$ subsplits)
and fixed for later experiments.
So \textsc{val} performance in the small-data regime may be an optimistic estimate
and in the large data regime a pessimistic estimate. True scaling behavior may be even stronger.
%
% As we increase the dataset size the performance on val split also improves.
Increasing dataset size consistently improves performance and has not yet saturated,
suggesting that simply collecting more demonstrations is likely to
lead to further gains.
%
% We are collecting \update{$50k$ more demonstrations} to scale our dataset to $100k$ overall.

\begin{figure}[h!]
    \begin{subfigure}{0.45\linewidth}
        \includegraphics[width=\linewidth]{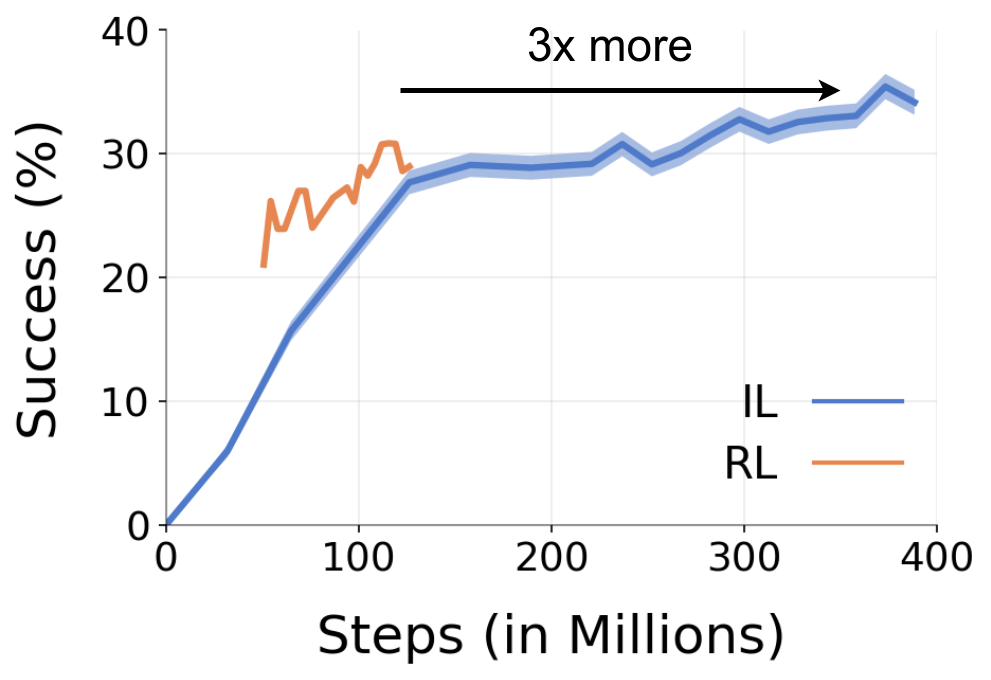}
        \caption{}
        \label{fig:steps_vs_perf}
    \end{subfigure}
    \begin{subfigure}{0.45\linewidth}
        \includegraphics[width=\linewidth]{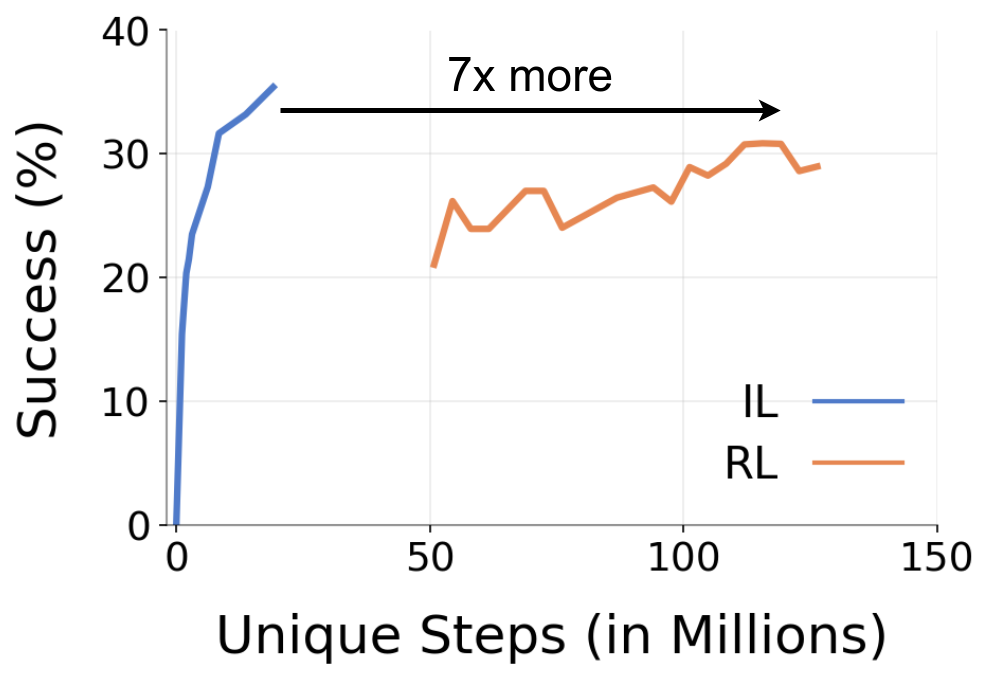}
        \caption{}
        \label{fig:unique_steps_vs_perf}
    \end{subfigure}
    \caption{Comparing RL and IL on (a) \textsc{val} success~\vs no. of training steps, and
        (b) \textsc{val} success~\vs no. of unique training steps.
        This distinguishes between an IL agent that learns from a static dataset~\vs an RL agent
        that gathers unique trajectories on-the-fly.}
    \label{fig:performance}
    \vspace{-5pt}
\end{figure}

\textbf{Sample Efficiency}.
\figref{fig:performance} plots \textsc{val} success against no.~of training steps of experience (in millions) in \figref{fig:steps_vs_perf}
and against \emph{unique} steps of experience in \figref{fig:unique_steps_vs_perf}.
Recall that IL involves ${\sim}21$ epochs on a static dataset of ${\sim}70k$ demos,
while RL (from~\cite{ye_iccv21}) gathers unique agent-driven trajectories on-the-fly.
\figref{fig:steps_vs_perf} shows that IL behaves like supervised learning (as expected) with improvements coming from long training schedules;
unfortunately, this means that wall-clock training times are not lower than RL.
\figref{fig:unique_steps_vs_perf} shows that
IL requires $7$x fewer unique steps of experience to outperform success and is thus much more sample-efficient.

%\figref{fig:performance} plots \textsc{val} success against no. of training steps of experience (in millions).
%
%To plot this, we created training subsplits ranging in size from $1.1$M to $10.1$M steps in experience,
%and trained models with the same set of hyperparameters on each split.
%
%Comparing unique steps of experience (\figref{fig:unique_steps_vs_perf}) taken by the IL agent trained on
%human demonstrations (which is a static dataset) against the Red Rabbit RL agent~\cite{ye_iccv21} (that gathers unique trajectories on-the-fly),
%our IL agent requires $12$x fewer steps to achieve comparable success,
%suggesting that in this data-performance regime, imitation learning is significantly more sample efficient.
%
%Comparing no. of steps of experience to get a sense for training time comparison (\figref{fig:steps_vs_perf}),
%our IL agent requires $3$x more policy updates to reach comparable success.

% \TODO{Figure: No. of steps of training~\vs performance on train, val -- compare to RL on sample complexity; ask Joel / Oleksandr for their plots.}

\textbf{Zero-shot results on Gibson}~\cite{xia_cvpr18} are in~\secref{sec:zero_shot_gibson}.

\subsection{\textbf{\pickplace}}

\begin{figure*}[t!]
    \vspace{-10pt}
    \centering
    \includegraphics[width=1.05\textwidth]{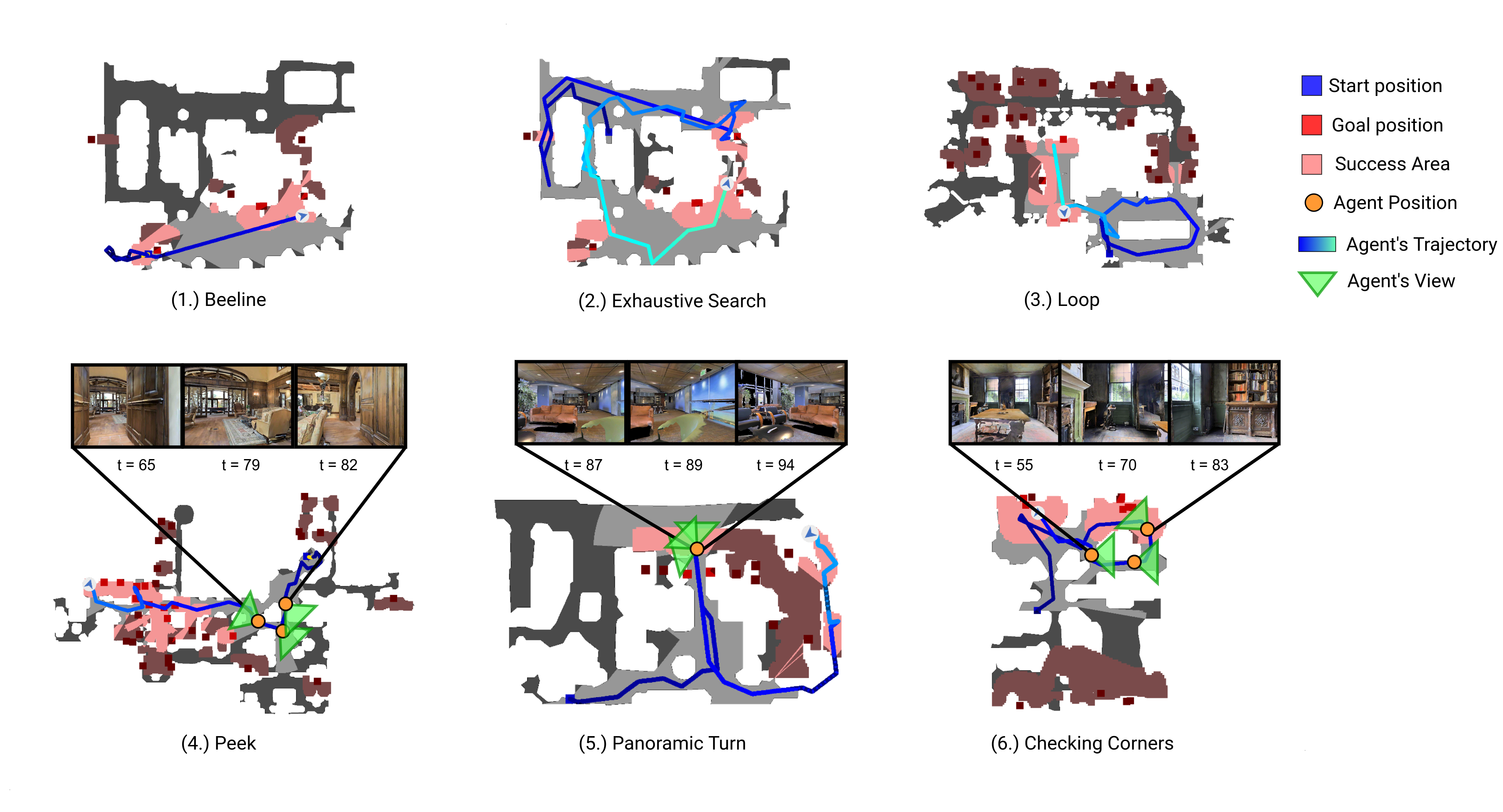}
    \vspace{-25pt}
    \caption{Visualizations of different learnt agent behaviors.
        Best viewed in the video at \href{https://ram81.github.io/projects/habitat-web}{\tt{ram81.github.io/projects/habitat-web}}.}
    \label{fig:qualitative}
    \vspace{-5pt}
\end{figure*}

% % % % % % % % % % % % % % % % % % % % % % % % % % % % % % % % % % % % % % % %
% Table: Pick-and-place results
% % % % % % % % % % % % % % % % % % % % % % % % % % % % % % % % % % % % % % % %
\begin{table} %[h!]
    \centering
    \resizebox{1\linewidth}{!}{
        \begin{tabular}{@{}cllrrcrr@{}}
            \toprule
            & & & \multicolumn{2}{c}{\textsc{val}} & & \multicolumn{2}{c}{\textsc{test}} \\
            \cmidrule{4-5} \cmidrule{7-8}
            & & Method & Success $\%$ $(\mathbf{\uparrow})$ & SPL $\%$ $(\mathbf{\uparrow})$
                & & Success $\%$ $(\mathbf{\uparrow})$ & SPL $\%$ $(\mathbf{\uparrow})$ \\
            \midrule
            \multirow{3}{*}{\rotatebox[origin=c]{90}{{\footnotesize \textsc{New Inits.}}} $\begin{dcases} \\ \\ \\ \end{dcases}$}
            \\[-10pt]
            & \rownumber & IL w/ Shortest Paths
                & $1.9$ & $1.8$
                & & $1.7$ & $1.6$ \\
            & \rownumber & IL w/ Human Demos
                & $17.6$ {\scriptsize $\pm0.8$} & $9.7$ {\scriptsize $\pm0.3$}
                & & $17.5$ & $9.8$ \\
            & \rownumber & Humans
                & $87.2$ & $21.8$
                & & $89.1$ & $21.9$ \\[5pt]

            \hdashline
            \multirow{3}{*}{\rotatebox[origin=c]{90}{{\footnotesize \textsc{New Instr.}}} $\begin{dcases} \\ \\ \\ \end{dcases}$}
            \\[-7pt]
            & \rownumber & IL w/ Shortest Paths
                & $1.3$ & $1.2$
                & & $1.1$ & $1.0$ \\
            & \rownumber & IL w/ Human Demos
                & $15.9$ {\scriptsize $\pm0.2$} & $8.4$ {\scriptsize $\pm0.4$}
                & & $15.1$ & $8.3$ \\
            & \rownumber & Humans
                & $85.0$ & $21.0$
                & & $86.1$ & $20.5$ \\[5pt]

            \hdashline
            \multirow{3}{*}{\rotatebox[origin=c]{90}{{\footnotesize \textsc{New Envs}}} $\begin{dcases} \\ \\ \\ \end{dcases}$}
            \\[-7pt]
            & \rownumber & IL w/ Shortest Paths & $-$ & $-$
                & & $0.2$ & $0.3$ \\
            & \rownumber & IL w/ Human Demos & $-$ & $-$
                & & $8.3$ & $4.1$ \\
            & \rownumber & Humans & $-$ & $-$
                & & $94.9$ & $20.5$ \\

            \bottomrule
            \end{tabular}
    }
    \vspace{5pt}
    \caption{Pick-and-place results on splits constructed with unseen initializations in seen environments (1-3), with
        unseen instructions (4-6), and with unseen environments (7-9).}
    \label{tab:pick_place_results}
    \vspace{-20pt}
\end{table}

\begin{table*}[h]
    \centering
    \resizebox{0.75\textwidth}{!}{
        \begin{tabular}{@{}llrrrrrrr@{}}
            \toprule
            & Method & OC (\%) & SC (\%) & GRTS (\%) & Peeks (\%) & PT (\%) & Beeline (\%) & ES (\%) \\
            \midrule
            \rownumber & IL w/ shortest paths & $4.2${\scriptsize $\pm1.1$} & $31.2${\scriptsize $\pm3.2$} & $20.5${\scriptsize $\pm4.3$} & $3.0${\scriptsize $\pm1.9$} & $0.0${\scriptsize $\pm0.0$} & $0.0${\scriptsize $\pm5.2$} & $10.1${\scriptsize $\pm3.5$} \\
            \rownumber & IL w/ human demos & $21.4${\scriptsize $\pm1.8$} & $72.1${\scriptsize $\pm3.5$} & $22.4${\scriptsize $\pm4.1$} & $19.6${\scriptsize $\pm4.4$} & $4.3${\scriptsize $\pm2.3$} & $10.3${\scriptsize $\pm5.1$} & $55.3${\scriptsize $\pm5.6$} \\
            \rownumber & RL \cite{ye_iccv21} & $14.6${\scriptsize $\pm1.6$} & $66.6${\scriptsize $\pm5.1$} & $27.7${\scriptsize $\pm8.5$} & $9.7${\scriptsize $\pm5.5$} & $0.0${\scriptsize $\pm0.0$} & $0.1${\scriptsize $\pm2.2$} & $49.0${\scriptsize $\pm7.0$} \\
            \rownumber & Humans & $15.4${\scriptsize $\pm1.6$} & $70.3${\scriptsize $\pm3.4$} & - & $13.8${\scriptsize $\pm3.9$} & $5.1${\scriptsize $\pm2.4$} & $23.6${\scriptsize $\pm4.8$} & $52.1${\scriptsize $\pm5.6$}  \\
            \bottomrule
            \end{tabular}
    }
    \vspace{5pt}
    \caption{Quantifying semantic exploration behaviors for IL agents trained
        on shortest paths (row $1$) and human demonstrations (row $2$), the Red Rabbit
    RL agent~\cite{ye_iccv21} (row $3$), and humans (row $4$).}
    \label{tab:objnav_sem_exp_metrics}
    \vspace{-20pt}
\end{table*}

% \noindent \textbf{Evaluation}.
% The agent is successful in solving an episode if it places the object specified in the instruction on top of the receptacle specified in the instruction. Specifically, an episode is considered successful if the euclidean distance between the object and receptacle center is within $0.7m$ and the object is at a height greater than receptacle center with max steps in environment limited to $1500$ steps.

\textbf{Results}.
We report results in Table~\ref{tab:pick_place_results} across three evaluation splits.
%
% We compare our approach with a baseline trained on shortest path demonstrations for all $3$ splits(row 1, 2, and 7).
% Our results are categorized into three evaluation splits of varying difficulty --
1) New Initializations: new locations of objects and receptacles. This tests
generalization to unseen locations in seen environments. 2) New Instructions:
compositionally novel object-receptacle combinations of objects and receptacles
individually seen during training.
% This tests compositional generalization to unseen combinations of known categories.
3) New Environments: generalization to $2$ scenes held out from training.
% 1) new object-receptacle locations for objects, receptacles, environments seen during training,
% 2) new object-receptacle combinations for objects and receptacles individually seen during training,
% and 3) new environments for objects and receptacles.
%
Similar to \objnav and as described in~\secref{sec:dataset}, we also report results
with shortest paths. Again, these paths are significantly shorter (average $342$~\vs $932$ steps per demonstration)
and hence, we generate a larger dataset of $25.7k$ episodes roughly matching the
cumulative steps of experience with human demonstrations ($8.8M$ shortest path
steps~\vs $11.5M$ human steps).
%
% In order to match the experience in steps with human demonstrations we generate
% $3$x shortest path demonstrations for training the shortest path baseline.
%
% With a dataset of same steps in experience and same set of hyperparameters the shortest
% path baselines achieves $0.2\%$ success and $0.1\%$ SPL for unseen environments split (row 7). Using human demonstrations provides absolute improvement of +$8.2\%$ in success and +$1.6\%$ in SPL over shortest path demonstrations. Baselines trained on human demonstrations outperform the baselines trained on shortest path demonstrations on all splits by large margins. This demonstrates the effectiveness of the human demonstrations for behavior cloning over shortest path demonstrations for the tasks that requires exploration and interactions in the environment.
Training on $9.5k$ human demonstrations achieves $17.5\%$ success, $9.8\%$
SPL on new object-receptacle initializations (row $2$).
Across splits, training on shortest paths hurts success by $8$-$16\%$.
Going to new object-receptacle pairs, success drops by $2.4\%$ (row $5$~\vs $2$),
and then going to new environments further hurts success by $6.8\%$ (row $8$~\vs $5$).
We also trained an RL policy with the exploration and distance-based rewards from \cite{thda_iccv21},
but it failed to get beyond $0\%$ success on new object-receptacle intializations.
See the appendix (Sec.~\ref{sec:pick_place_training}) for training details.

\textbf{Performance~\vs Dataset size}.
Similar to \objnav, we trained policies on $2.5k$ to $9.5k$ subsets of our \pickplace
data, and found that performance continues to improve with more data.
Figure in appendix (Sec.~\ref{sec:perf_vs_dataset_size_pick_place}).
%
%
% agent on val split (unseen initializations split of Table. \ref{tab:pick_place_results}) with respect to the dataset size, figure in Appendix.
% For this we create $3$ splits of the human demonstrations dataset of size $2.5$k, $5$k, and $10$k and train a baseline with same set of hyperparameters on these splits. As we increase the dataset size the performance on val split also improves. The plot in Fig.(X) \TODO{cite figure} shows that dataset size vs performance curve hasn't saturated at $10$k demonstrations for \pickplace indicating that performance on val split will increase with more data.

% \textbf{Generalization to unseen instructions and unseen environments}.
% Behavior cloning is notoriously sensitive to distribution shifts between training and evaluation. To probe this in our agents, we constructed three new evaluation splits.

% For all $3$ splits, agents trained on human demonstrations outperform agents trained on shortest path demonstrations by a large margin, refer to Table. \ref{tab:pick_place_results} for results. We also report human performance on val and test splits in Table. \ref{tab:pick_place_results}. Behavior cloned agents on these demonstrations are still quite far from human performance showing that the \pickplace task is far from being called "solved", our best agent is able to achieve $8.6\%$ success and $1.7\%$ SPL on unseen environments whereas human performance is at $94.9\%$ success and $20.5\%$ SPL.

%% file: sections/main/analysis.tex
% \vspace{-5pt}
\section{Characterizing Learned Behaviors}
\label{sec:analysis}

% Using our best agents trained on Our agents demonstrate promising performance (${\sim}27\%$ val success on \objnav and ${\sim}8\%$ test success on \pickplace) in tasks that require navigation in large and complex environments and interaction with objects environments. In this section we analyze \objnav agent behavior qualitatively and quantitatively compare the semantic exploration metrics of these behavior cloned agents with state-of-the-art RL baselines. We restrict the analysis of semantic exploration metrics to \objnav agents because the \pickplace dataset doesn't have objects and receptacles at semantically meaningful locations. We scope our analysis to the baseline trained on human demonstrations with RedNet-predicted semantic input, unless otherwise noted.

% \textbf{Qualitative Analysis}.
To characterize the behaviors learnt by our best IL
agents, we first sample $300$ validation \objnav episodes for each method and manually
categorize the behavior observed.
% ~\eg beeline to goal once seen, peeking into rooms, looping, exhaustive search,~\etc.
A subset of observed behaviors are visualized in~\figref{fig:qualitative}.
Our agents demonstrate sophisticated object-search behaviors~\eg peeking into
rooms to maximize sight coverage (SC), instead of occupancy coverage (OC),
checking corners of rooms for small objects, beelining to goal object once seen,
exhaustive search (ES), turning in place to get a panoramic view (PT), and looping back to recheck some areas.
Amusingly, unlike shortest path / RL agents, these IL agents also stand idle and
`look around' \ie turn in place, like humans.
% this is not the case with agent trained on shortest path demonstrations as the
% shortest path follower doesn't need to look around when generating the trajectories.
% \subsection{Quantitative Analysis}.
Table~\ref{tab:objnav_sem_exp_metrics} quantifies these behaviors.
See appendix (Sec.~\ref{sec:sem_exp_metrics}) for details on how these were computed.
Agents trained with IL on human demonstrations have higher coverage
(both occupancy and sight), peeking behavior,
panoramic turns, beelines, and exhaustive search than RL.
RL-trained agents achieve higher average Goal Room Time Spent (GRTS) --~\ie time
spent in the room containing the target object -- but also have significantly higher
variance in GRTS across scenes compared to IL agents.
See appendix (Sec.~\ref{sec:sem_exp_metrics}) for a per-scene breakdown of GRTS as well as histograms of time spent
in each room (instead of just target room) when searching for a target object.
We also discuss limitations of our approach in the appendix Sec.~\ref{sec:limitations}.

%% file: sections/main/conclusion.tex
\vspace{-5pt}
\section{Conclusion}
\label{sec:conclusion}
\vspace{-5pt}

We developed the infrastructure to collect
human demonstrations at scale
and using this, trained imitation learning (IL) agents
on $92k+$ human demonstrations for \objnav and \pickplace.
On \objnav, we found that IL using $70k$ human
demonstrations outperforms RL using $240k$ agent-gathered trajectories, and on
\pickplace, IL agents get to ${\sim}18\%$ success while RL fails to get beyond $0\%$.
Qualitatively, we found that IL agents pick up on sophisticated object-search
behavior implicitly captured in human demonstrations, much more prominently than RL agents.
Overall, we believe our work makes a compelling case for investing in large-scale imitation learning
of human demonstrations.

%% file: sections/main/acknowledgement.tex
\xhdr{Acknowledgements}.
We thank Devi Parikh for help with brainstorming and direction,
Joel Ye for answering questions about his Red Rabbit~\cite{ye_iccv21},
Oleksandr Maksymets for the THDA~\cite{thda_iccv21} episode generation pipeline,
and Erik Wijmans for help with debugging DDP.
The Georgia Tech effort was supported in part by NSF, ONR YIP, and ARO PECASE. The views and conclusions contained herein are those of the authors and should not be interpreted as necessarily representing the official policies or endorsements, either expressed or implied, of the U.S. Government, or any sponsor.

%% file: sections/supplement/appendix.tex
\clearpage
\newpage
\newpage
\appendix
\section{Appendix}
\subsection{Pick\&Place}
\label{sec:pick_and_place}

Recall that in the pick-and-place task (\pickplace), an agent must follow an instruction
of the form \myquote{Place the {\tt <object>} on the {\tt <receptacle>}},
without being told the location of the {\tt <object>} or {\tt <receptacle>} in a new environment.
The agent must explore and navigate to the object, pick it up, explore and
navigate to the receptacle, and place the previously picked-up object on it.
In this section, we go over statistics of the human demonstrations dataset,
how our \pickplace imitation learning (IL) agents scale as a function of training
dataset size, and details of our reinforcement learning baseline for \pickplace.

\vspace{-5pt}
\subsubsection{Dataset Stats}
\label{sec:pick_place_stats}
\vspace{-5pt}

\begin{figure*}[t!]
    \centering
    \begin{minipage}[a]{0.24\textwidth}
        \begin{subfigure}{\textwidth}
            \centering
            \includegraphics[width=\textwidth]{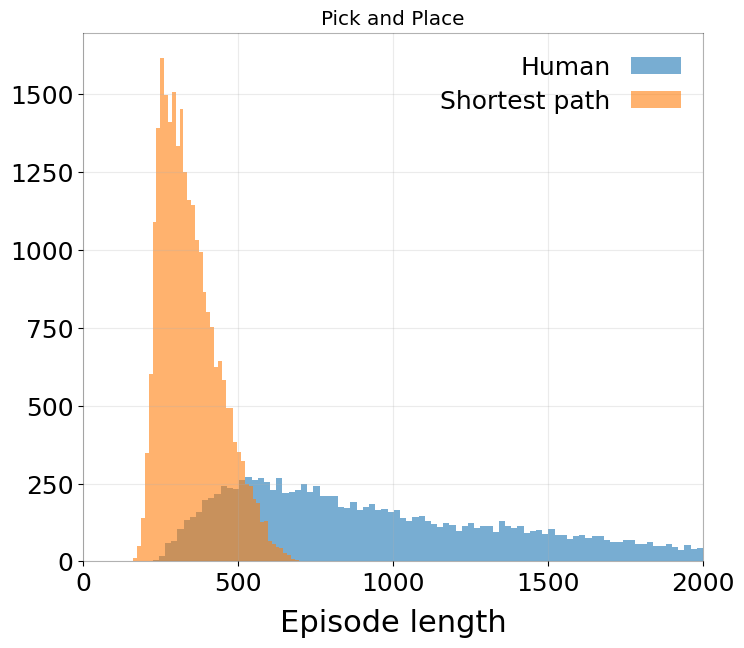}
        \end{subfigure}
    \end{minipage}
    \hfill
    \begin{minipage}[a]{0.73\textwidth}
        \begin{subfigure}{\textwidth}
            \centering
            \includegraphics[width=\textwidth]{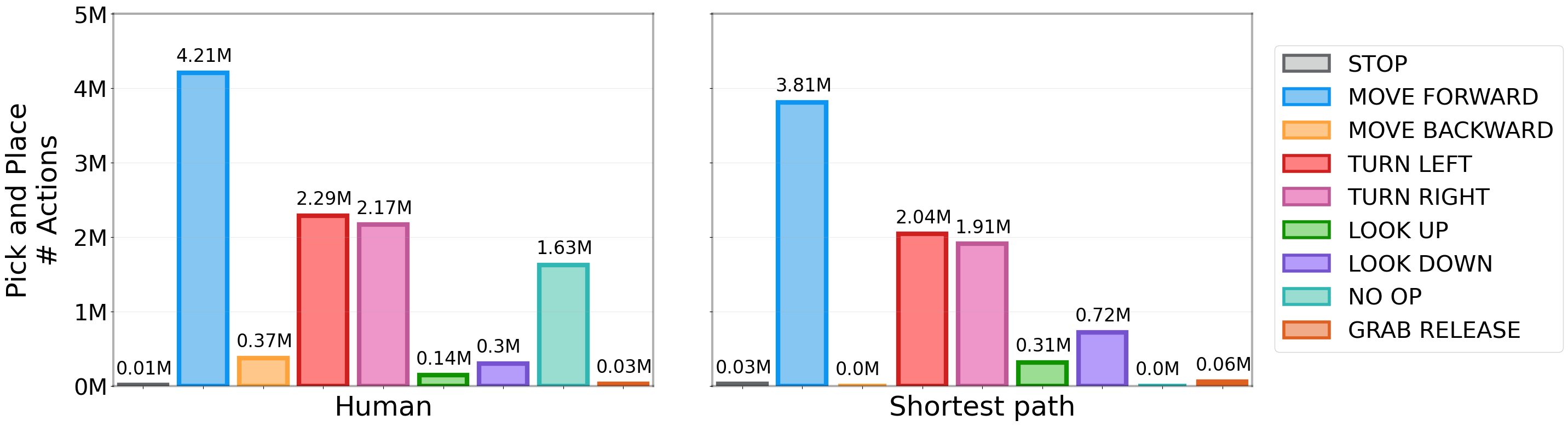}
        \end{subfigure}
    \end{minipage}
    \vspace{5pt}
    \caption{Comparison of episode lengths and action histograms
        for human demonstrations~\vs shortest paths for \pickplace.
        Human demonstrations are longer and have a more uniform action distribution
        than shortest paths.}
    \label{fig:pick_place_dataset_stats}
    \vspace{-15pt}
\end{figure*}

\figref{fig:pick_place_dataset_stats} compares the episode length and action histograms for human
and shortest path demonstrations for \pickplace. Human demonstrations are longer
(average $932$ vs $342$ steps per demonstration) and have a more uniform action
distribution compared to shortest paths. Human demonstrations also make use of all
$9$ actions whereas shortest path demonstrations use only $6$ actions. Notice,
humans also tend to stand idle and do nothing ($50$ms of idle time is translated to a NO\_OP action).
They likely use this time to strategize their next set of actions to explore the environment,
which is not the case in shortest path demonstrations (by design).

\begin{figure}[t!]
    \centering
    \includegraphics[width=1.05\columnwidth]{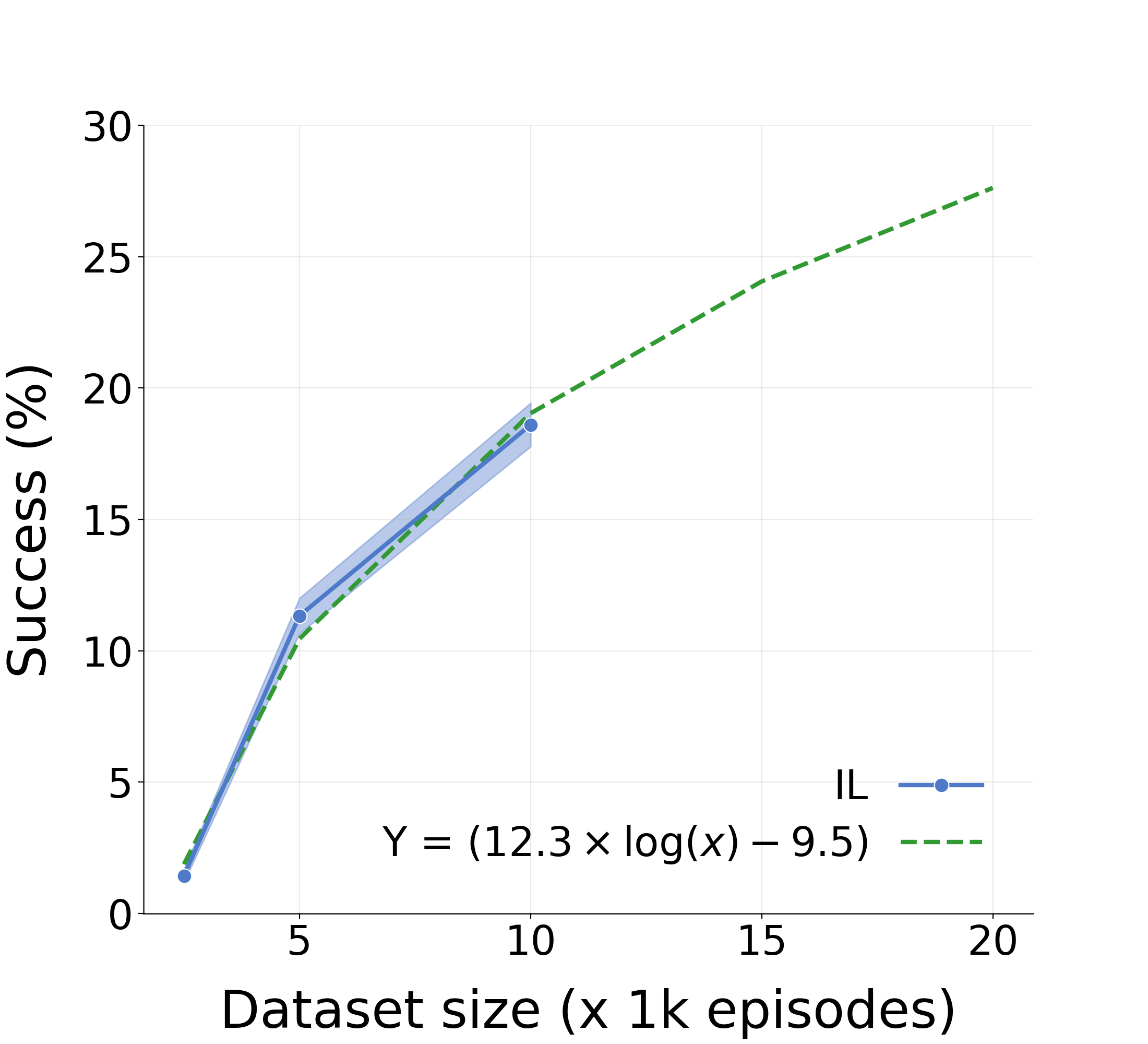}
    \caption{Dataset size vs performance plot for \pickplace.}
    \label{fig:dataset_size_vs_perf_pp}
\end{figure}

\subsubsection{RL Baseline}
\label{sec:pick_place_training}
% In this section we describe the implementation details of the reinforcement learning baseline.
Similar to the imitation learning baseline, our base policy is a simple CNN+RNN architecture.
We first embed all sensory inputs using feed-forward modules.
%
% We separately encode RGB and Depth observations.
For RGB, we use a randomly initialized ResNet18~\cite{he_cvpr16}.
For depth, we use a ResNet50 that was pretrained on PointGoal navigation using DDPPO~\cite{wijmans_iclr20}.
In addition to RGBD observations, the policy gets as input language instructions
of the form \myquote{Place the {\tt <object>} on the {\tt <receptacle>}} encoded using
a single-layer LSTM~\cite{hochreiter_nc97}.
RGBD and instruction features are concatenated to form an observation embedding,
which is fed into a $2$-layer, $512$-d GRU at every timestep.
We train this policy for ${\sim}100$M steps on ${\sim}9.5k$ episodes.

\textbf{Rewards}.
The agent receives a sparse success reward $r_{success}$,
a slack reward $r_{slack}$ to motivate faster goal-seeking,
an exploration reward $r_{explore}$, an object seen reward
$r_{seen}$, a grab/release success reward $r_{grab\_release}$,
and a drop penalty reward $r_{drop\_penalty}$ to penalize dropping the object far from the receptacle.
For incentivizing exploration, we use a visitation-based coverage reward from Ye~\etal\cite{ye_iccv21}.
We first divide the map into a voxel grid of $2.5m \times 2.5m \times 2.5m$ voxels and reward the
agent for visiting each voxel. Similar to \cite{ye_iccv21}, we smooth $r_{explore}$ by decaying it
by number of steps the agent has spent in the voxel (visit count $v$).
To ensure that the agent prioritizes \pickplace (and not just exploration), we decay $r_{explore}$ based
on episode timestep $t$ with a decay constant of $d = 0.995$.
The agent is provided a reward for exploration until it sees the object.
Once it sees the object, it receives a significant positive reward $r_{seen}$, and then the reward switches to a
path-efficiency based navigation reward.
In addition, the agent also receives a significant positive reward when it successfully grabs
a object or releases the object close to the receptacle.

\begin{subequations}
\begin{align}
    r_{\text{total}} &= r_{\text{success}} + r_{\text{slack}} + r_{\text{explore}} + r_{\text{grab\_release}} \\
    & + r_{\text{drop\_penalty}} + r_{\text{seen}} \\
    r_{\text{success}} &= 5.5 \quad\quad\,\,\, \text{{\scriptsize on success}} \\
    r_{\text{slack}} &= -10^{-4} \quad \text{{\scriptsize per step}} \\
    r_{\text{seen}} &= 1.5 \quad \text{{\scriptsize First time object seen}} \\
    r_{\text{drop\_penalty}} &= -3.5 \quad \text{{\scriptsize Object dropped > $2m$ away from receptacle}} \\
    r_{\text{grab\_release}} &= 2.0 \quad \text{{\scriptsize Grab / release success}} \\
    % r_{\text{explore}} &= 0.25\frac{d^{t}}{v}
    r_{\text{explore}} &= 0.25 \times \frac{d^{t}}{v} \quad \text{{\scriptsize Until object seen}}
\end{align}
\end{subequations}

\textbf{Results}.
A policy trained with this reward for $100$M steps fails to get beyond $0\%$ success on the \pickplace task.
The agent learns to pick up the object at the start of training if it sees the object while navigating but it fails to search for the receptacle and place the object on top of receptacle.
Overall, throughout training, the agent doesn't solve the task successfully even once demonstrating the difficulty of the task and inadequacy of the above reward structure.

\subsubsection{Performance~\vs Dataset Size}
\label{sec:perf_vs_dataset_size_pick_place}
\figref{fig:dataset_size_vs_perf_pp} plots \textsc{Val} success of our IL agent~\vs
the size of the \pickplace human demonstrations dataset.
We trained policies on $2.5k$ to $9.5k$ subsets of the data.
Performance continues to improve with more data and has not saturated.

\subsection{Zero-shot \objnav results on Gibson}
\label{sec:zero_shot_gibson}

To test generalization of the IL agents trained on human demonstrations, we report zero-shot results
by transferring our policy trained on $40k$ human demonstrations to the Gibson dataset~\cite{xia_cvpr18} \textsc{val} split
in Table \ref{tab:objnav_gibson_val}.
To enable zero-shot transfer of semantic features, we remap the common goal categories (chair, couch, potted plant, bed,
toilet, TV, dining-table) from Matterport3D~\cite{mp3d} to Gibson goal category IDs.
Our IL agent achieves $49.4\%$ success and $16.4\%$ SPL (row 7) with no finetuning on Gibson dataset.
% -- with just $40k$ human demonstrations from Matterport3D dataset.
Comparing our zero-shot results to approaches trained on Gibson, our IL agent
is $33.6\%$ better on success and $11.5\%$ better on SPL than an RL baseline
that takes RGBD + Semantics as input (row 3~\vs row 7).
Next, we compare our approach with SemExp~\cite{chaplot2020object} which builds explicit semantic maps and 
learns a goal-oriented semantic exploration policy which learns semantic priors for efficient navigation.
Our approach is $4.9\%$ worse on success and $3.5\%$ worse on SPL compared to SemExp~\cite{chaplot2020object} (row 6~\vs row 7).
\cite{ramakrishnan2022poni} uses a modular framework for \objnav by using a potential function conditioned on a
semantic map which is used to decide where to look for unseen object in an environment. Our approach is $24.1\%$ worse on success
and  $24.6\%$ worse on SPL.

\begin{table}[t]
    \centering
    \resizebox{0.975\columnwidth}{!}{
        \begin{tabular}{@{}lrr@{}}
            \toprule
            Method & Success $(\mathbf{\uparrow})$ & SPL $(\mathbf{\uparrow})$ \\
            \midrule
            \rownumber Random & $0.4\%$ & $0.4\%$ \\
            \rownumber RGBD+RL \cite{savva_iccv19} & $8.2\%$ & $2.7\%$ \\
            \rownumber RGBD+Semantics+RL \cite{mousavian2018sem_midlevel} & $15.9\%$ & $4.9\%$ \\
            \rownumber Classical Map + FBE & $40.3\%$ & $12.4\%$ \\
            \rownumber Active Neural SLAM \cite{chaplot_iclr20} & $44.6\%$ & $14.5\%$ \\
            \rownumber SemExp~\cite{chaplot2020object} & $54.4\%$ & $19.9\%$ \\
            \rownumber PONI~\cite{ramakrishnan2022poni} & $73.6\%$ & $41.0\%$ \\
            \midrule
            \rownumber \textbf{IL w/ $70k$ Human Demos (Zero-Shot)} & $49.5\%$ & $16.4\%$ \\
            \bottomrule
            \end{tabular}
    }
    \vspace{5pt}
    \caption{ObjectNav results on the Gibson \textsc{val} split.}
    \label{tab:objnav_gibson_val}
\end{table}

\subsection{Estimating time using a LoCoBot motion model}
\label{sec:motion_model}

To estimate the time a robot would take to execute the collected
human trajectories in the real world, we use the LoCoBot motion model from Krantz~\etal~\cite{krantz_iccv21}.
This model consists of a rotation function that maps turn angle to time and a translation function
that maps straight-line distance to time.
For estimating time required for grab/release actions, we replace them with
$0.15m$ forward steps and use the straight-line distance
translation function.
We use the \movebase controller from \cite{krantz_iccv21} for all our time estimates,
with the following rotation and translation equations:

\begin{equation}
    y^{rotate} =
        0.000358 \phi^2 + 0.108 \phi + 2.23
\end{equation}

\begin{equation}
    y^{translate} =
        4.2 x + 0.362
\end{equation}

\subsection{Characterizing Learnt Behaviors}
\label{sec:sem_exp_metrics}

\begin{figure}[t!]
    \centering
    \includegraphics[width=1.05\columnwidth]{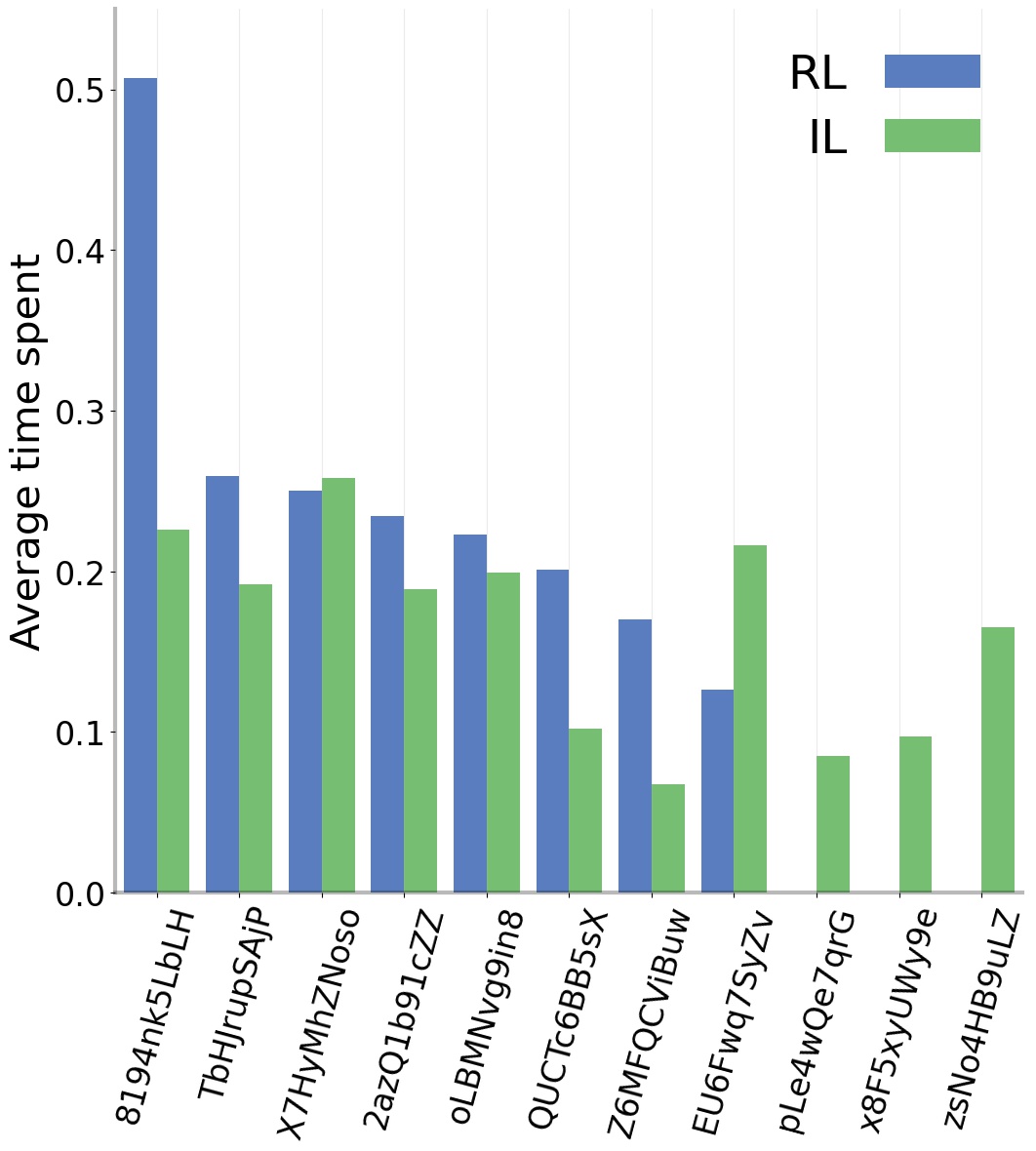}
    \caption{Per scene breakdown of GRTS for IL and RL agent on MP3D \textsc{val} split.}
    \label{fig:grts_rl_vs_il}
\end{figure}

In this section, we describe the metrics used to characterize the exploration behavior
exhibited by these agents in Sec. 7 in the main paper.
These include 1) Occupancy Coverage (OC) and 2) Sight Coverage (SC) introduced in Sec. 4 in the main paper,
as well as 3) Goal Room Time Spent (GRTS) -- the number of steps as a fraction of total episode length
an agent takes within the room bounding box containing the target object,
4) Peeks -- check if the agent steps back into the last visited room after taking
just ${\sim}10$ steps in another room, 5) Panoramic Turn (PT) -- whether the agent
stands at one place and turns left and right to get sweeping views, 6) Beeline -- if
the agent takes $10$ continuous forward actions before reaching the goal in the
last $15$ steps, 7) Exhaustive Search (ES) -- $\geq 75\%$ sight coverage.
%
% We use the same set of annotations to compute both the metrics.
To compute these metrics, we use the semantic annotations in Matterport3D.
These annotations provide $3$D bounding box coordinates for each room category in an environment.
We use these bounding box coordinates to track the rooms an agent visits during an episode.
%
% We use these two metrics as measures to quantitatively evaluate how well does agent
% explore the environment semantically when it failed to reach the goal.
GRTS gives us a measure of how often the agent ends up reaching
goal object room but doesn't successfully locate the object.
A higher GTRS suggests that the agent is at least good at reaching semantically
meaningful locations in search of the goal object.
We find that RL agents have higher average GRTS but also significantly higher variance in GRTS across scenes
while our IL agents have lower average GRTS but more consistently spend time in the target room (see~\figref{fig:grts_rl_vs_il}).
To evaluate not just the final room the agent ends up at, but all
the rooms it visits through the course of an episode, we also plot distributions of the
time spent per room category for each goal object (see~\figref{fig:pRTS_per_object})
for human demonstrations~\vs IL agents trained on human demonstrations~\vs RL agents.

\subsection{Inter-human Variance in \objnav}

\begin{figure}[t!]
    \centering
    \includegraphics[width=0.45\textwidth]{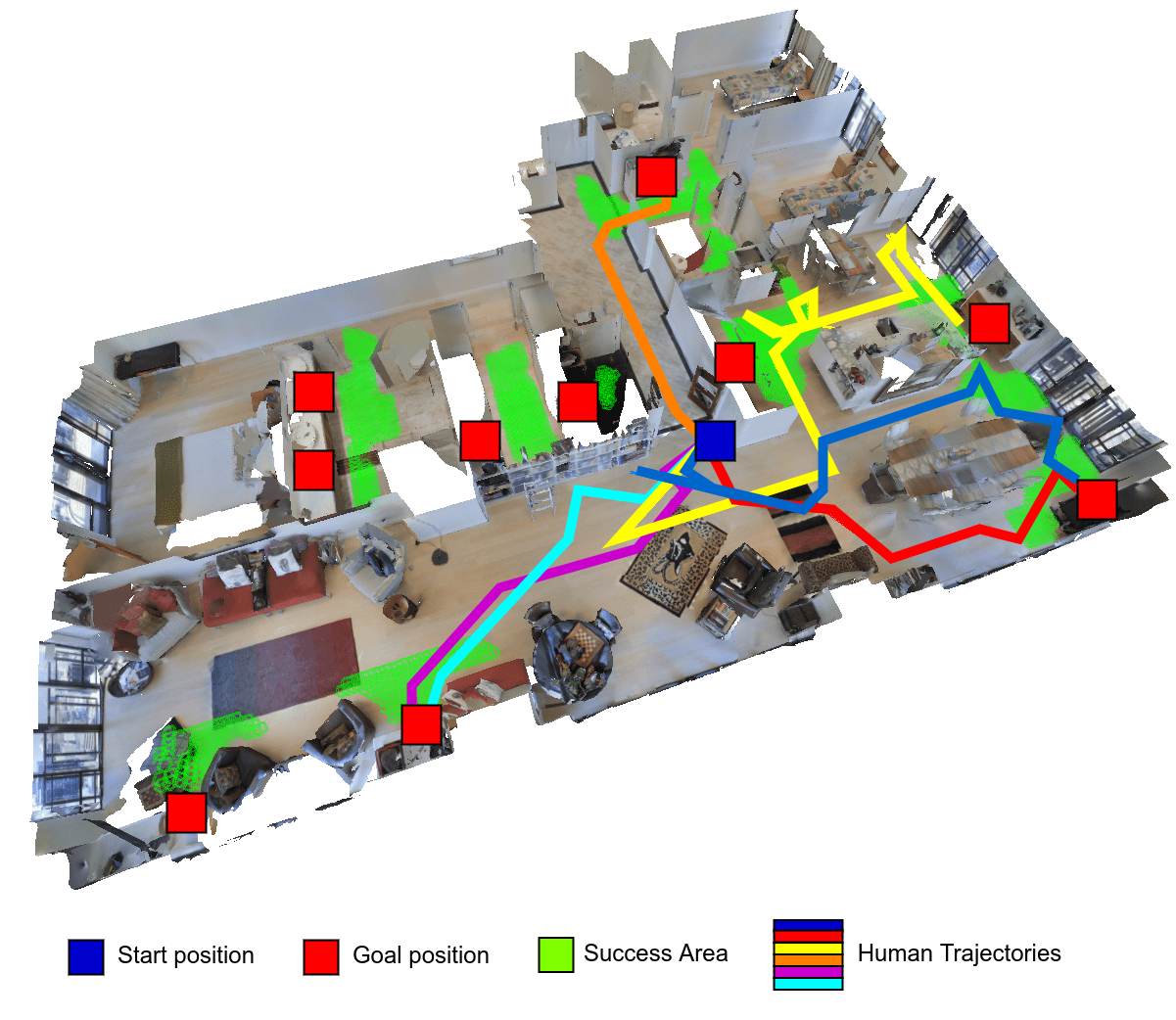}
    \caption{Visualizing multiple human demonstrations for \objnav all starting from the same start position and searching for `cabinet'.}
    \label{fig:human_trajectories}
\end{figure}

To get a sense for the variance in \objnav human demonstrations,
we collected $20$ unique human-provided trajectories
for the same initial location and target object (`cabinet').
This is visualized in~\figref{fig:human_trajectories}.
%
% To be able to visualize this, we specifically
% for the task of finding a cabinet to capture the variance in the trajectories different people take to solve the task.
%
% For this, we create a \objnav task (a AMT HIT) with same initial position for $20$ AMT workers
% and use this collected data to visualize a sample of $6$ trajectories in \figref{fig:human_trajectories}.
%
We see that there is quite a bit of diversity in navigation trajectories across humans.
They often navigate to different instances of the goal object category `cabinet', and
% We observe that each AMT worker takes a different route to the goal object and they don't necessarily
% end up reaching to the same instance of the goal object, which is not the case for shortest path trajectories (by design).
%
even when multiple humans go to the same object instance, the routes taken are different (red~\vs blue trajectory).
% These trajectories capture more area (sight and occupancy coverage) in an environment.

\begin{figure}[t!]
    \centering
    \begin{minipage}[a]{0.48\columnwidth}
        \begin{subfigure}{\columnwidth}
            \centering
            \includegraphics[width=\columnwidth]{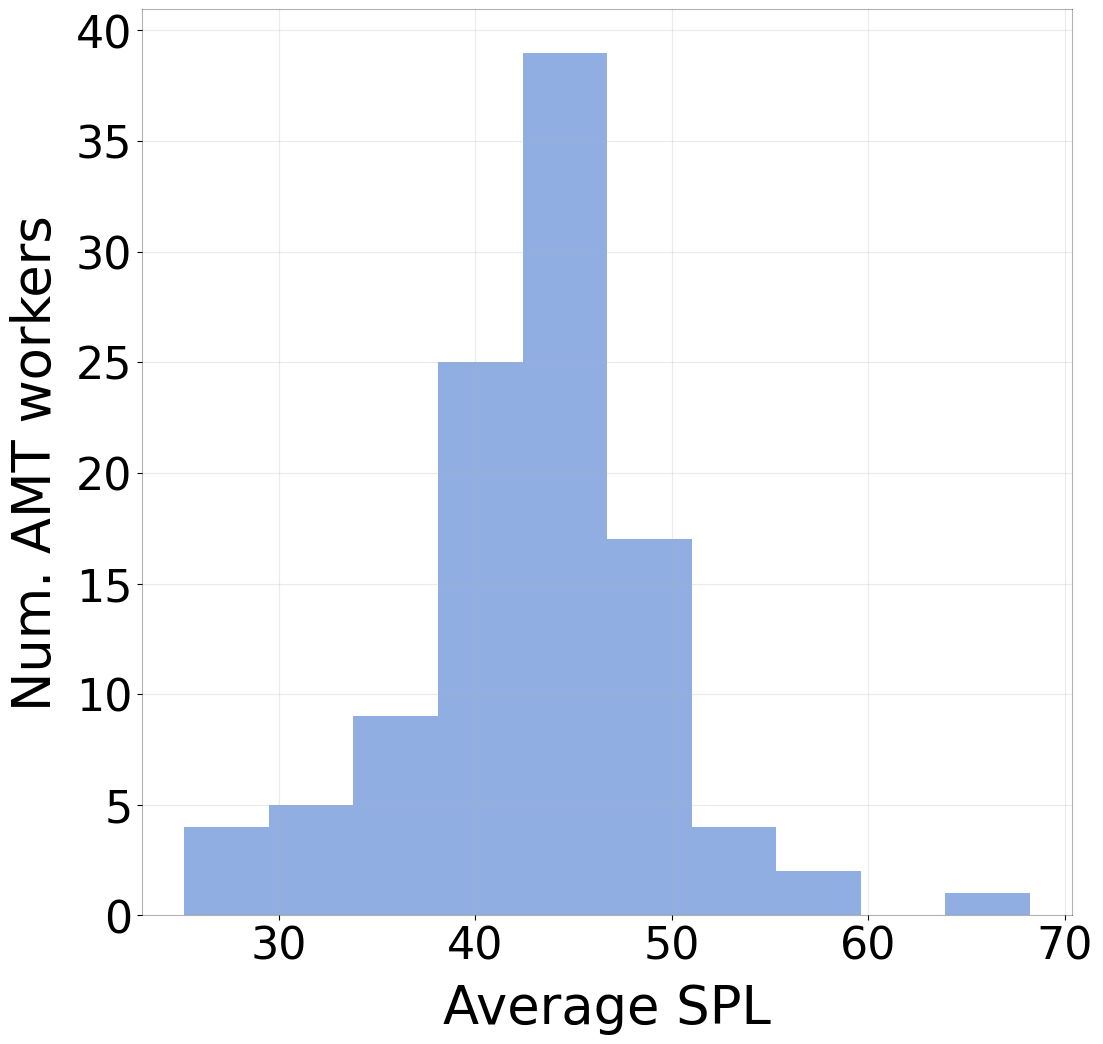}
            \caption{}
            \label{fig:human_spl_a}
        \end{subfigure}
    \end{minipage}
    \hfill
    \begin{minipage}[a]{0.48\columnwidth}
        \begin{subfigure}{\columnwidth}
            \centering
            \includegraphics[width=\columnwidth]{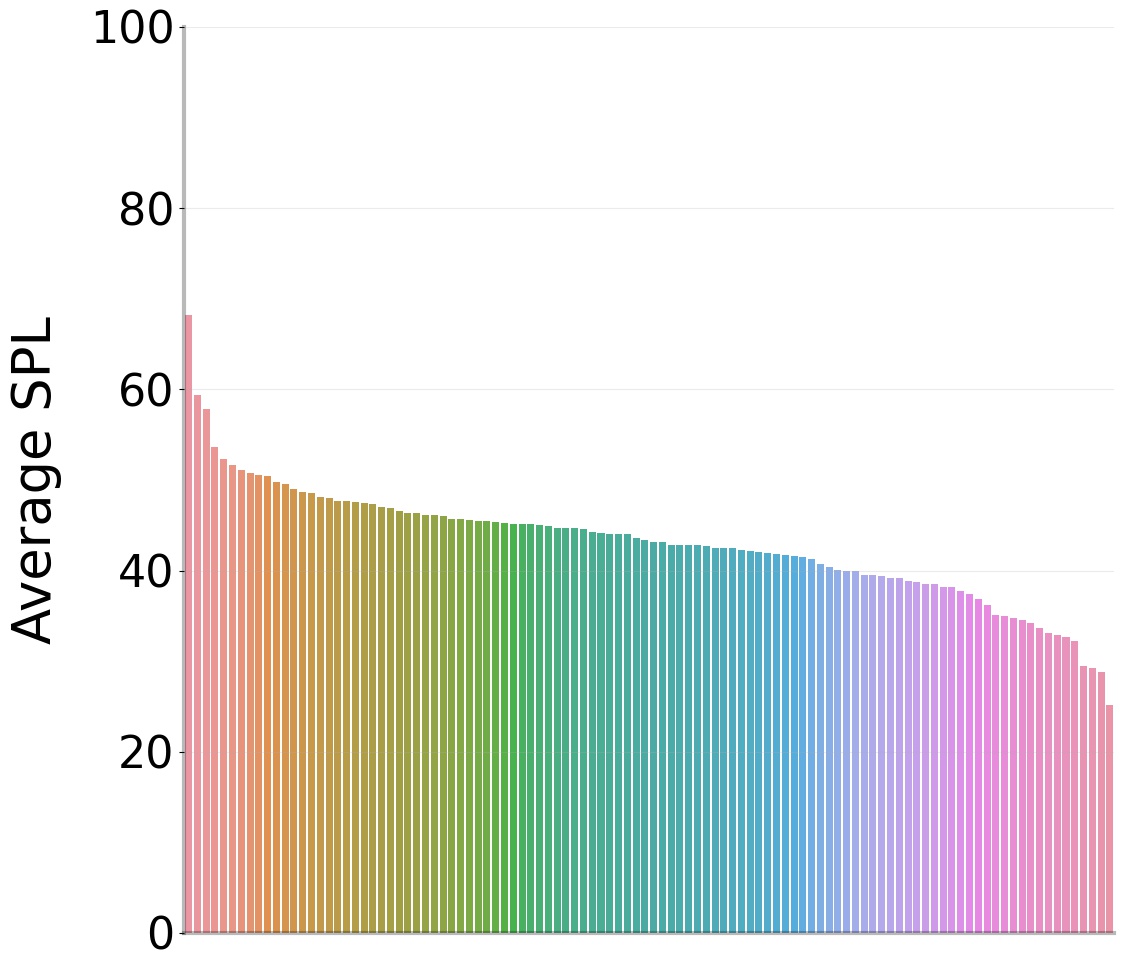}
            \caption{}
            \label{fig:human_spl_b}
        \end{subfigure}
    \end{minipage}
    \vspace{5pt}
    \caption{a) Histogram of average SPL for each AMT user for \objnav.
        b) Plot showing average SPL for AMT users for \objnav.
        These plots clearly demonstrate some humans are better at solving the \objnav task than others.}
    \label{fig:human_spl}
\end{figure}

We also plot the average SPL per AMT user in our dataset in~\figref{fig:human_spl}.
% This lets us analyze if some humans are much more efficient at \objnav.
%
We find that human performance has a lot of variability, ranging from
$25.2\%$ to $68.2\%$ (\figref{fig:human_spl_a}).
The SPL range that has the most humans is ${\sim}50\%$.
The best-performing human annotator achieves an SPL of $68.2\%$
averaged over $6$ episodes (\figref{fig:human_spl_b}),
which is particularly close to shortest paths and arguably \emph{super-human}.
% In \figref{fig:human_spl_a}, we plot a histogram of average SPL per user,
% we compute the average SPL for demonstrations collected from each user, and
% the plot in \figref{fig:human_spl_b} shows average SPL for each AMT worker.

\begin{figure}[t]
    \centering
    % \begin{minipage}[a]{0.48\textwidth}
    %     \begin{subfigure}{\textwidth}
    %         \centering
    %         \includegraphics[width=\textwidth]{figures/interface/amt_infra.png}
    %     \end{subfigure}
    % \end{minipage}
    % \hfill
    \begin{minipage}[a]{0.95\columnwidth}
        \begin{subfigure}{\columnwidth}
            \centering
            \includegraphics[width=\textwidth]{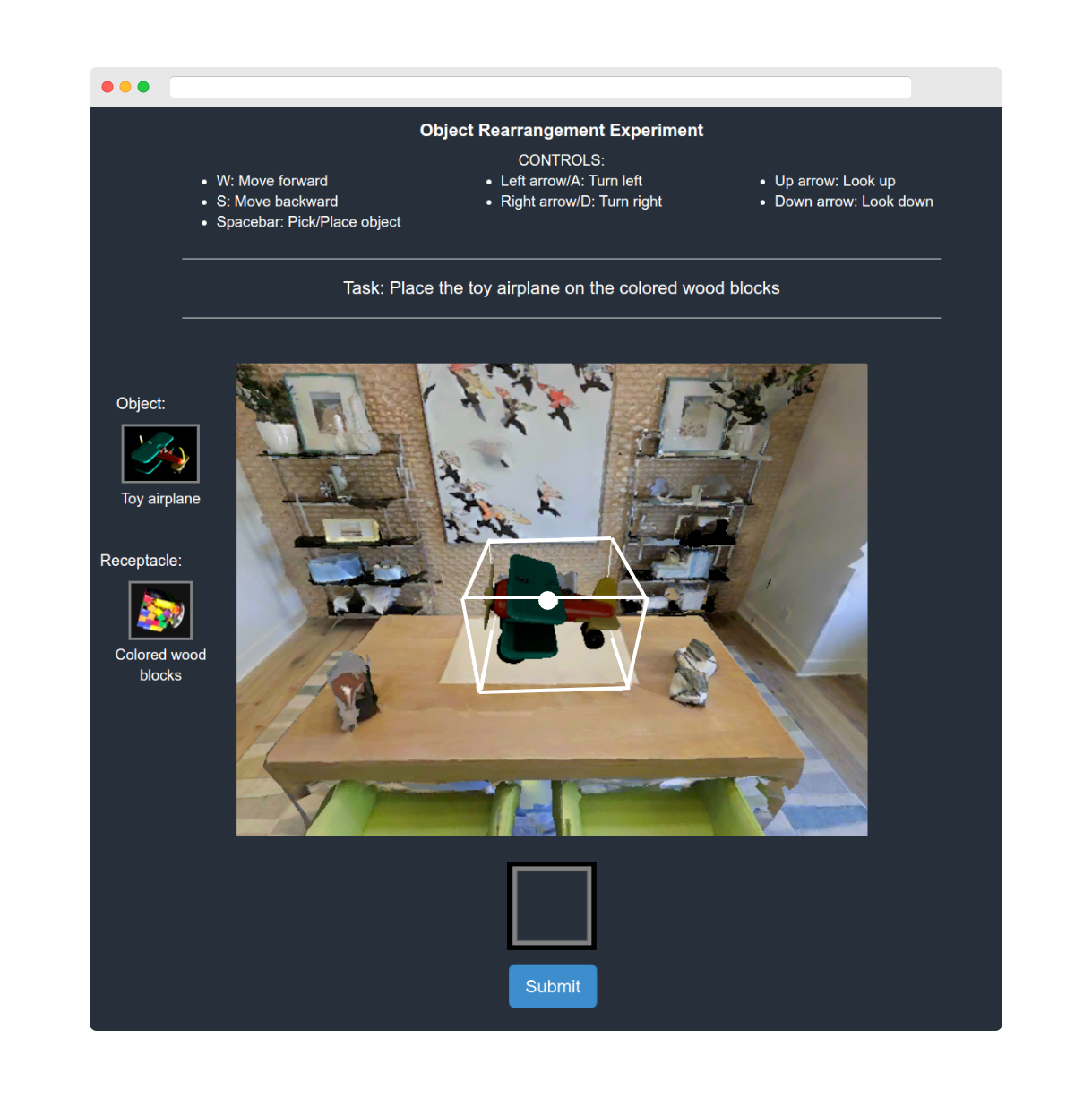}
        \end{subfigure}
    \end{minipage}
    \vspace{5pt}
    \caption{Screenshot of our Amazon Mechanical Turk interface for collecting
        \pickplace demonstrations. Users are provided the agent's first-person view
        of the environment and an instruction such as "Pick the toy airplane and place it on the the colored wood blocks".
        They can make the agent look around and move in the environment via keyboard
        controls, and can submit the task upon successful completion by clicking the `Submit' button.}
    \label{fig:amt_interface_pp}
\end{figure}

\begin{figure*}[t!]
    \centering
    \includegraphics[width=1.05\textwidth]{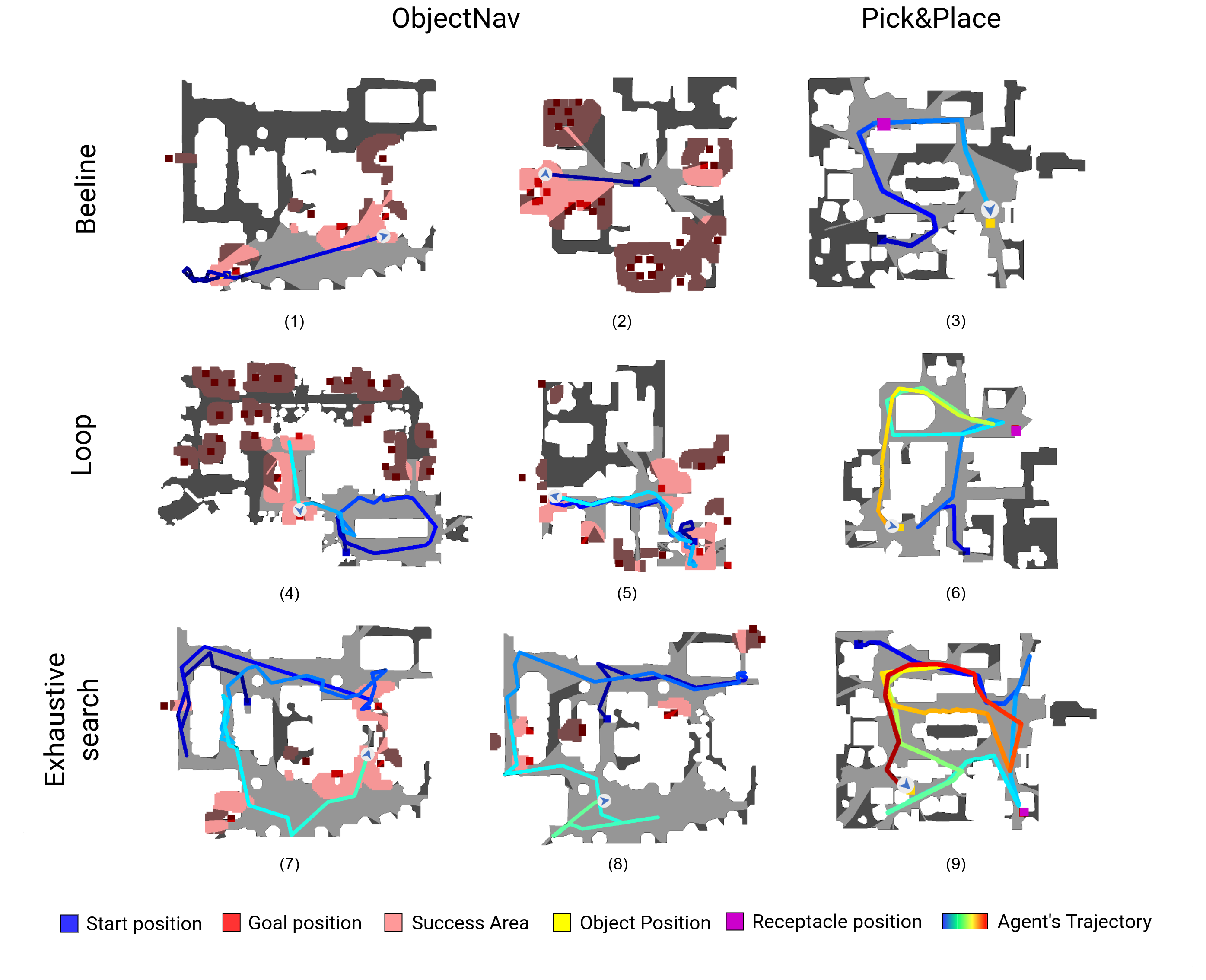}
    \caption{Visualizations of learnt agent behaviors for \objnav and \pickplace.
        Best viewed in videos at \href{https://sites.google.com/view/object-search-supp}{\tt{sites.google.com/view/object-search-supp}}.}
    \label{fig:supp_qualitative}
\end{figure*}

\begin{figure*}[t!]\ContinuedFloat
    \centering
    \includegraphics[width=1.0\textwidth]{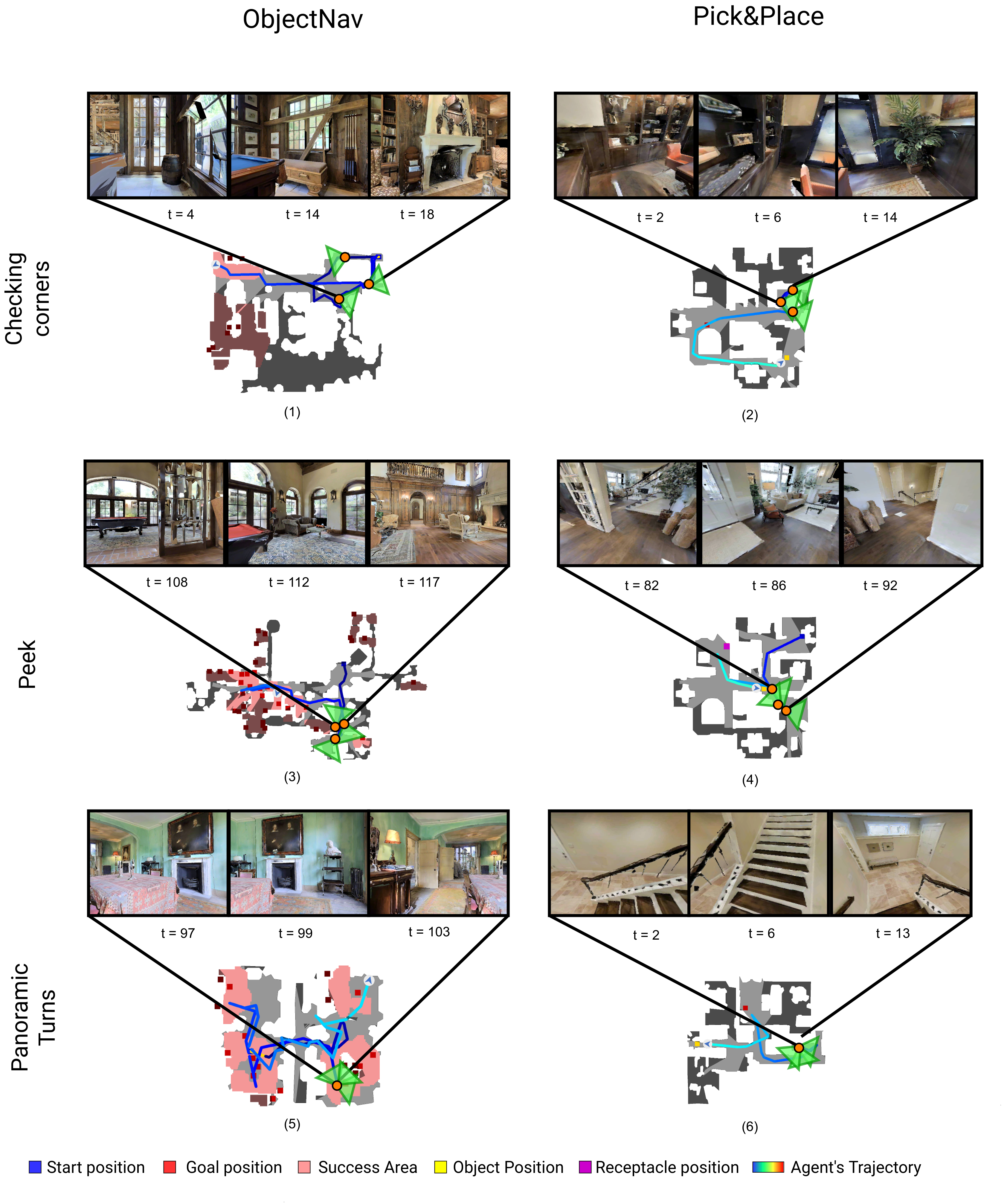}
    \caption{Visualizations of learnt agent behaviors for \objnav and \pickplace.
        Best viewed in videos at \href{https://ram81.github.io/projects/habitat-web}{\tt{ram81.github.io/projects/habitat-web}}.}
    \label{fig:supp_qualitative_2}
\end{figure*}

% \begin{figure}[t]
%     % \resizebox{0.95\linewidth}{!}{
%     \includegraphics[width=\linewidth]{figures/interface/amt_interface.png}
%     % }
%     \vspace{5pt}
%     \caption{Screenshot of AMT interface.\NOTE{Should we put this screenshot in supplement?}}
%     \label{fig:amt_interface}
% \end{figure}

% \subsection{Pick&Place RL Baseline}

\subsection{AMT Interface}
\label{sec:amt_interface_details}

\figref{fig:amt_interface_pp} shows a screenshot of our AMT interface for collecting \pickplace demonstrations.
For the \pickplace task, we provide humans with an
instruction of the form \myquote{Place the {\tt <object>} on the {\tt <receptacle>}},
without being told the location of the {\tt <object>} or {\tt <receptacle>} in a new environment,
and they can see agent's first-person view of the environment.
They can make the agent move, look around, and interact with the environment using keyboard controls.
Once the AMT user completes the task they can submit the task by clicking the `Submit' button.
We then run task-specific validation checks to ensure only successful tasks get submitted.

\textbf{Validation}.
To ensure data quality, every submitted AMT task goes through a set of validation
checks.
%
% Commented by - Ram
% These validation checks are task-specific, and the application is modularized
% to make it easy to plug in arbitrary validation checks specific to each task.
%
For \objnav, we use the same set of validation checks as the Habitat challenge
evaluation setup,~\ie a task is considered successful only when the user has
moved the agent to within $1m$ of the goal object.
We do not limit the maximum number of steps to allow users on AMT to explore
the environment.
This captures key human exploration behavior necessary to succeed at these tasks.
Similarly, for \pickplace, a task is considered successful when the target object
is placed on a receptacle object.
Specifically, we check if the Euclidean distance between the centers of the target
and receptable objects is less than $0.7m$, and that the target object is at a
height greater than the receptacle center.
%
% We describe these tasks in detail in the following sections.

\subsection{Limitations}
\label{sec:limitations}
Our approach is fundamentally limited by the limitations of imitation learning as our approach uses vanilla behavior cloning with inflection weighting. Additionally, these agents trained on human demonstrations exhibit some common failure cases. Some examples of common failure cases are -- reaching close to the goal object but not within goal radius and ending episode early, trying to move straight when agent is colliding and getting stuck, looping around multiple instances of the goal object and as a result, exceeding maximum episode steps, and exploring the environment and not finding the goal object.
Our approach is also limited by the amount of human demonstrations we can gather and the agent architecture being trained on this dataset.
Currently, we use a vanilla CNN+RNN architecture to learn imitation learning policies but we can build better architecture which make full use of the rich semantic information these human demonstrations have.

%% file: sections/supplement/analysis_figure.tex
\begin{figure*}[t]
    \centering
    \begin{minipage}[a]{0.32\textwidth}
        \begin{subfigure}{\textwidth}
            \centering
            \includegraphics[width=\textwidth]{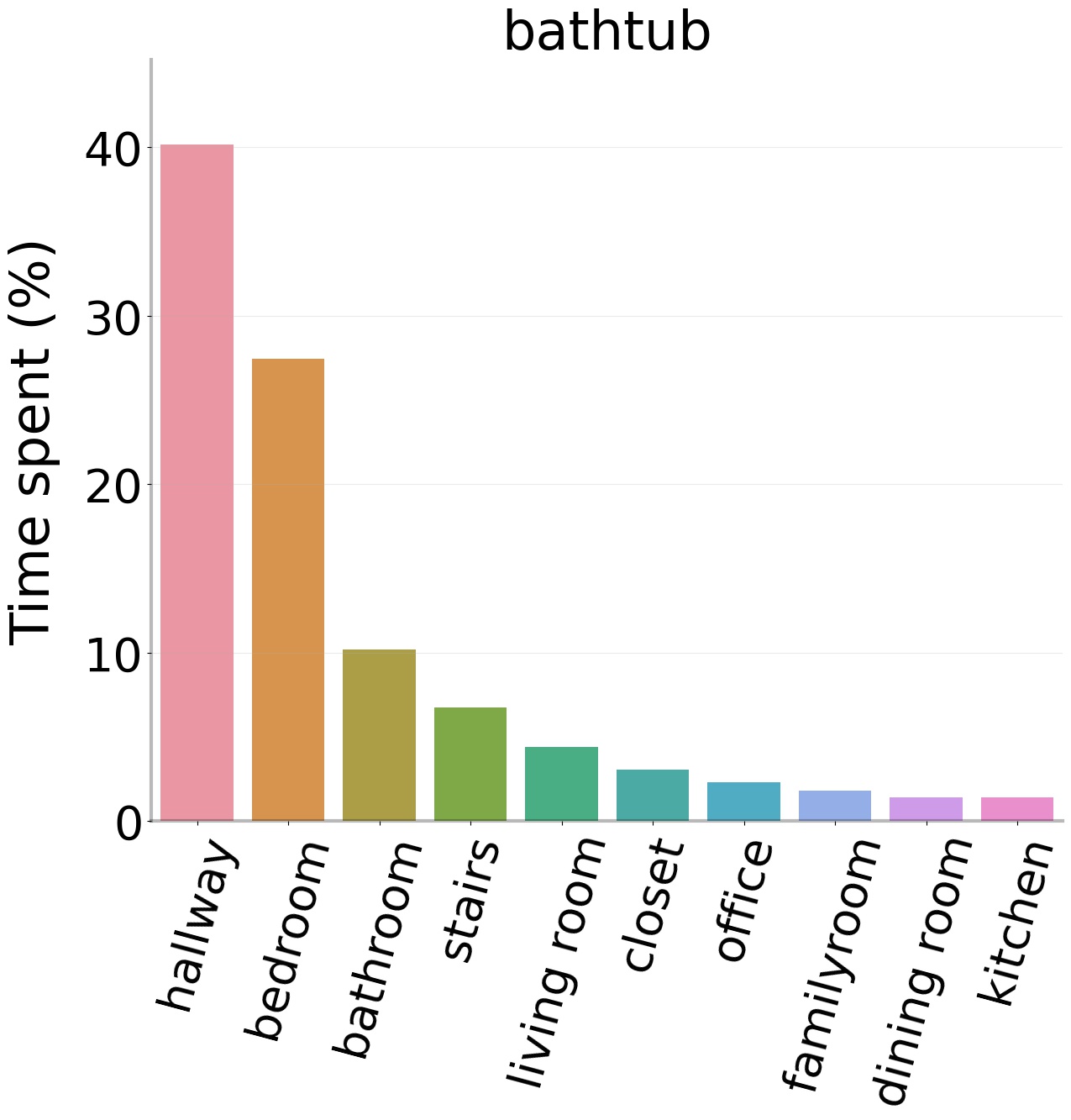}
            %\caption{Humans}
        \end{subfigure}
    \end{minipage}
    \hfill
    \begin{minipage}[a]{0.32\textwidth}
        \begin{subfigure}{\textwidth}
            \centering
            \includegraphics[width=\textwidth]{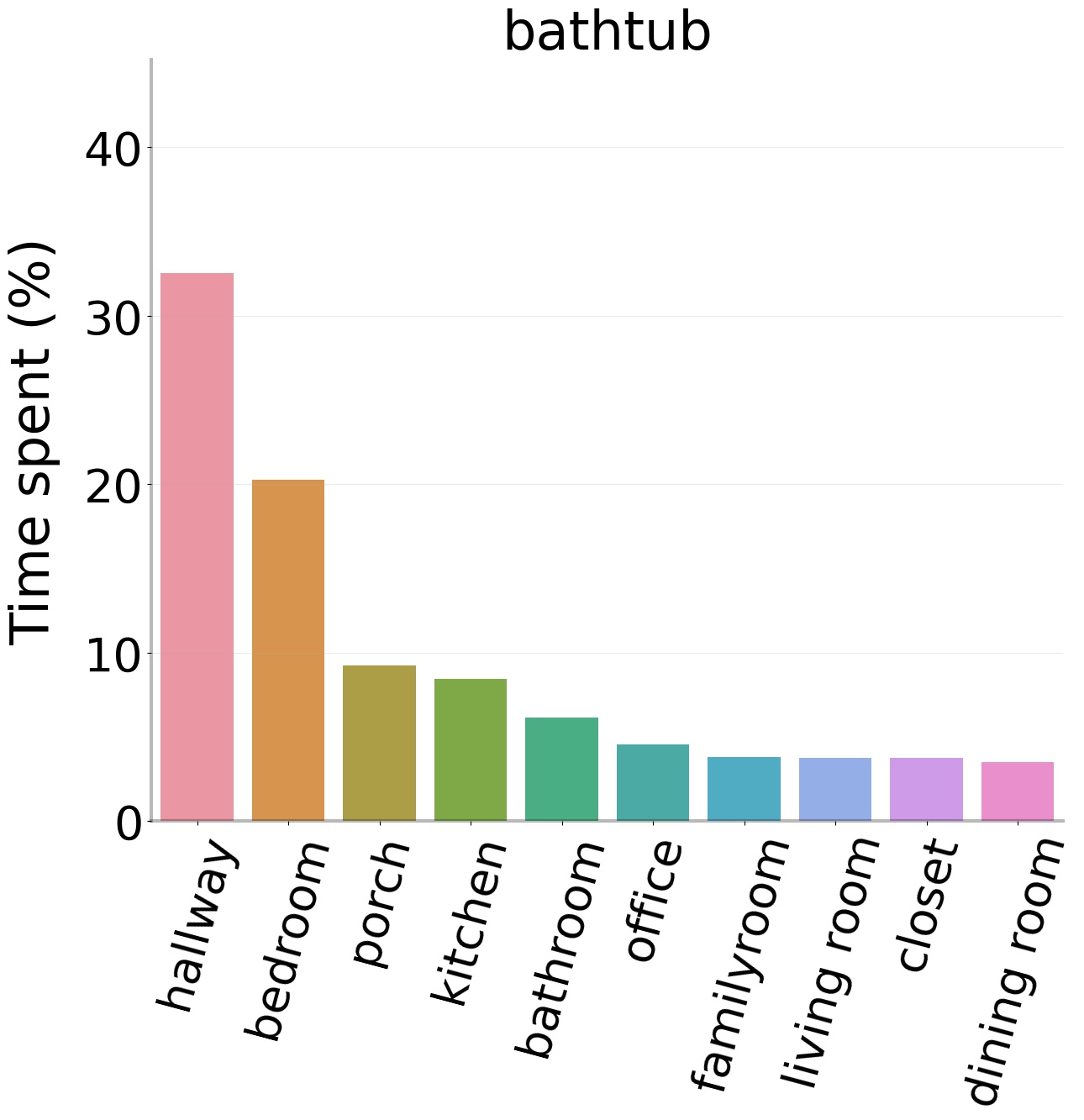}
            %\caption{IL on $40k$ Human demos}
        \end{subfigure}
    \end{minipage}
    \hfill
    \begin{minipage}[a]{0.32\textwidth}
        \begin{subfigure}{\textwidth}
            \centering
            \includegraphics[width=\textwidth]{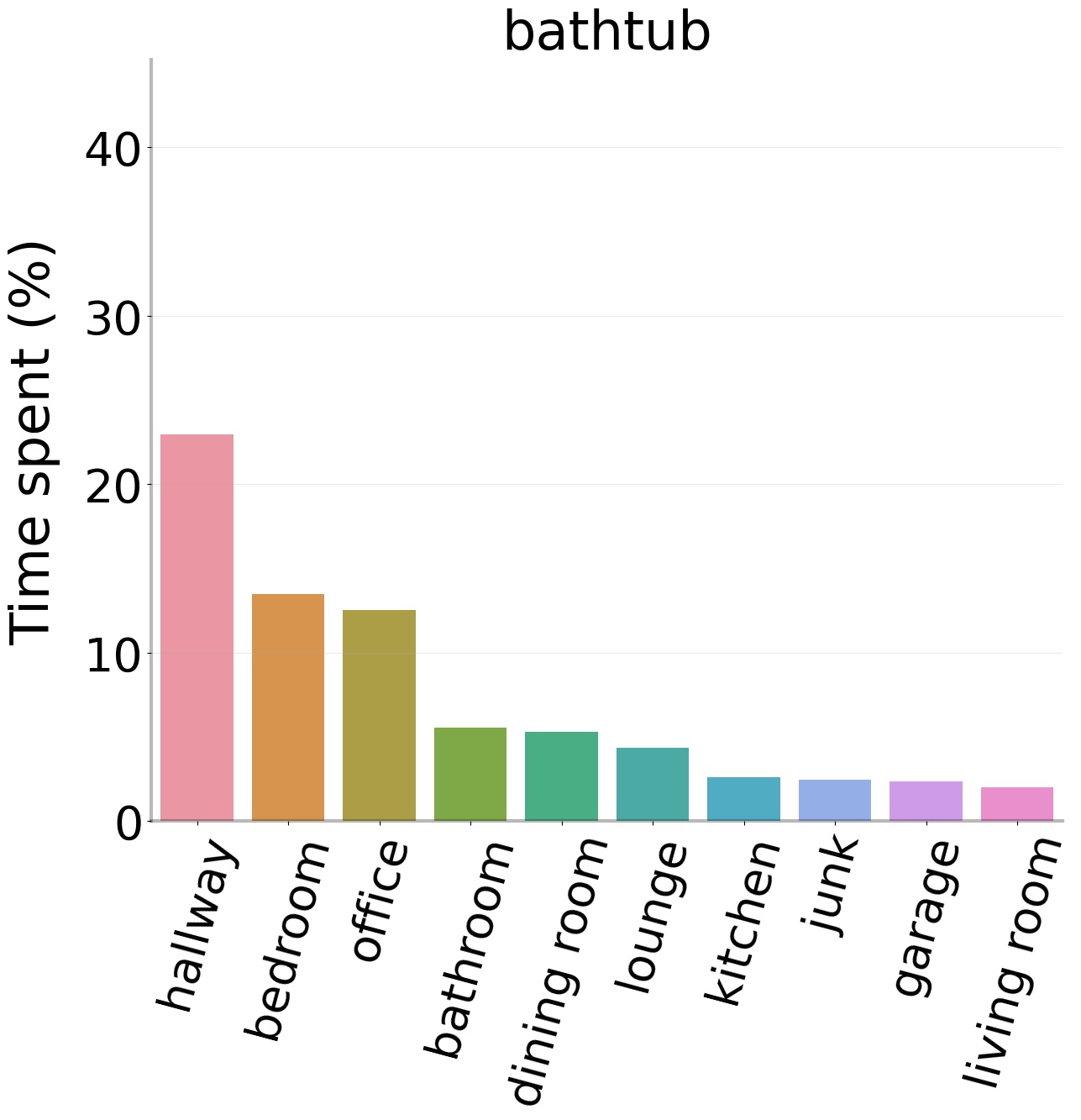}
            %\caption{RL}
        \end{subfigure}
    \end{minipage}

    \begin{minipage}[a]{0.32\textwidth}
        \begin{subfigure}{\textwidth}
            \centering
            \includegraphics[width=\textwidth]{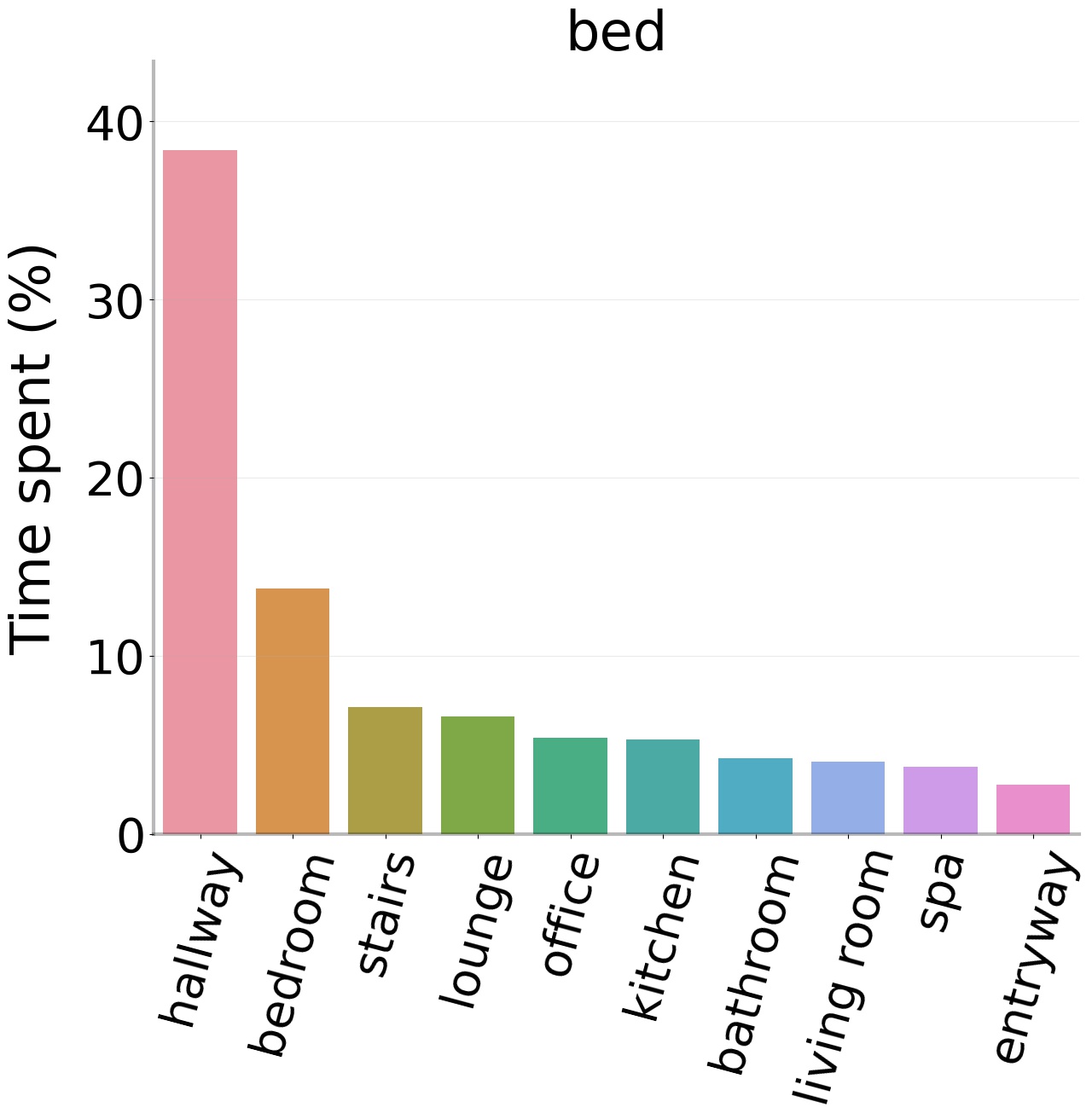}
            %\caption{Humans}
        \end{subfigure}
    \end{minipage}
    \hfill
    \begin{minipage}[a]{0.32\textwidth}
        \begin{subfigure}{\textwidth}
            \centering
            \includegraphics[width=\textwidth]{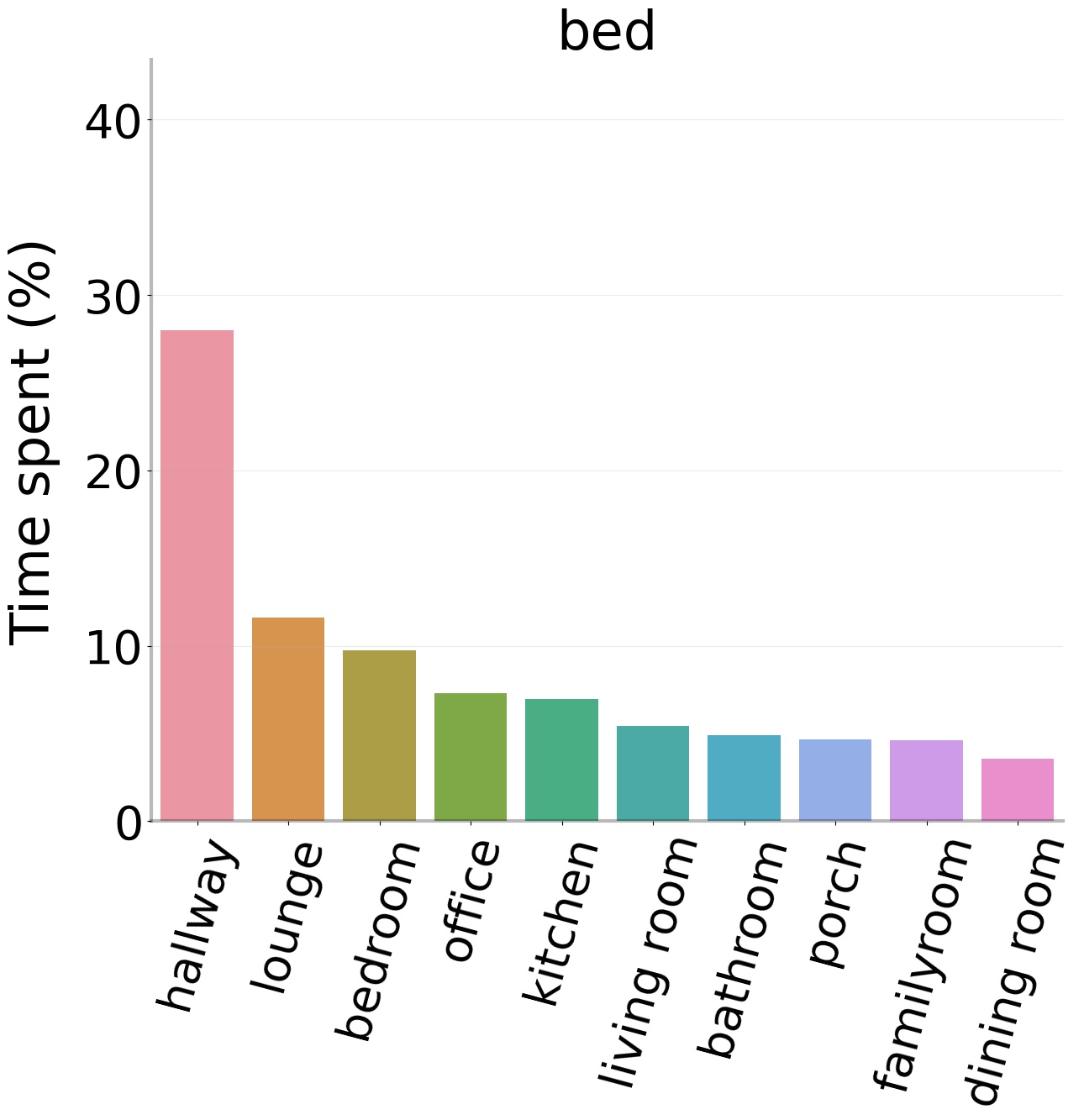}
            %\caption{IL on $40k$ Human demos}
        \end{subfigure}
    \end{minipage}
    \hfill
    \begin{minipage}[a]{0.32\textwidth}
        \begin{subfigure}{\textwidth}
            \centering
            \includegraphics[width=\textwidth]{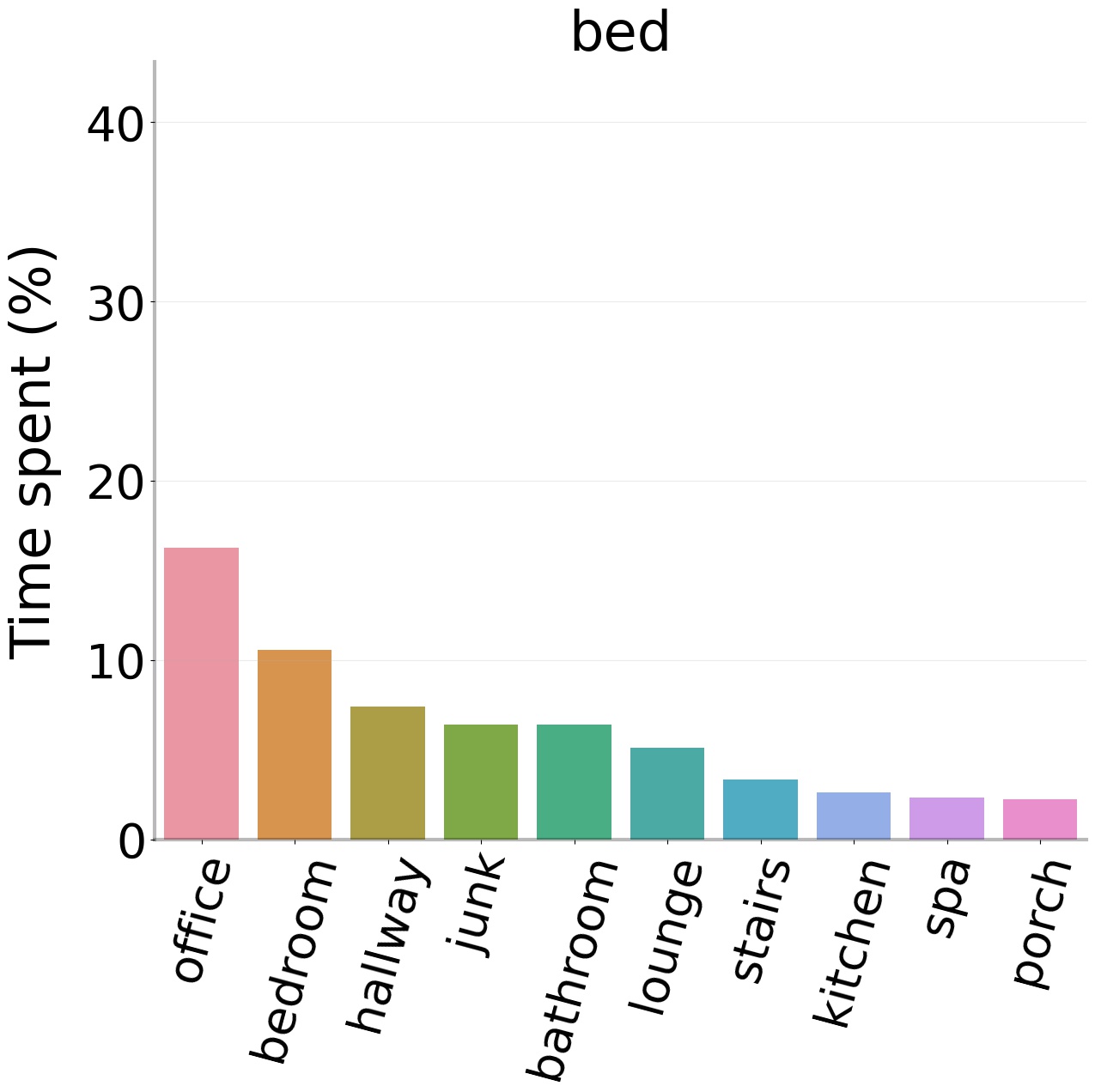}
            %\caption{RL}
        \end{subfigure}
    \end{minipage}

    \begin{minipage}[a]{0.32\textwidth}
        \begin{subfigure}{\textwidth}
            \centering
            \includegraphics[width=\textwidth]{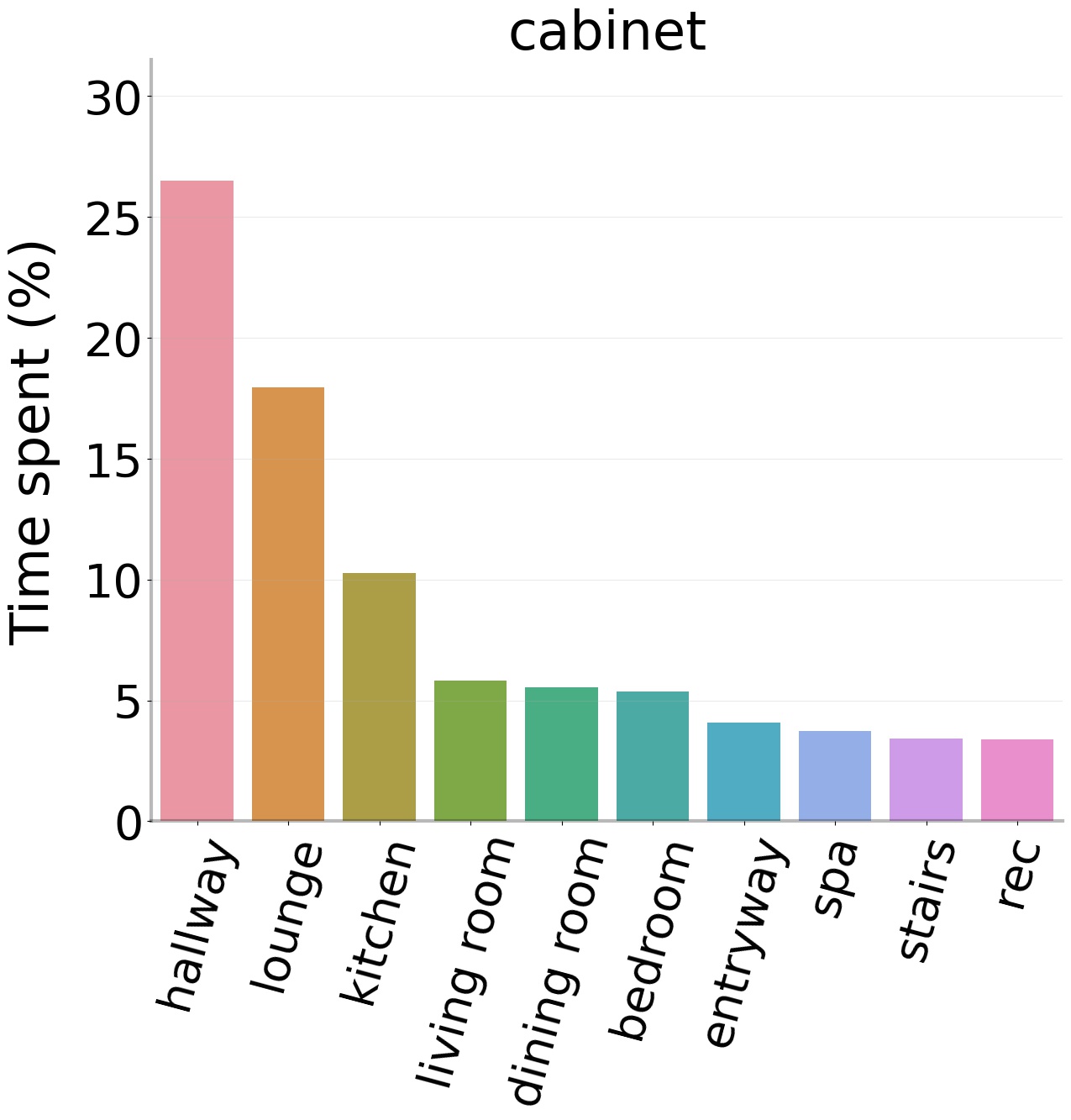}
            \caption{Humans}
        \end{subfigure}
    \end{minipage}
    \hfill
    \begin{minipage}[a]{0.32\textwidth}
        \begin{subfigure}{\textwidth}
            \centering
            \includegraphics[width=\textwidth]{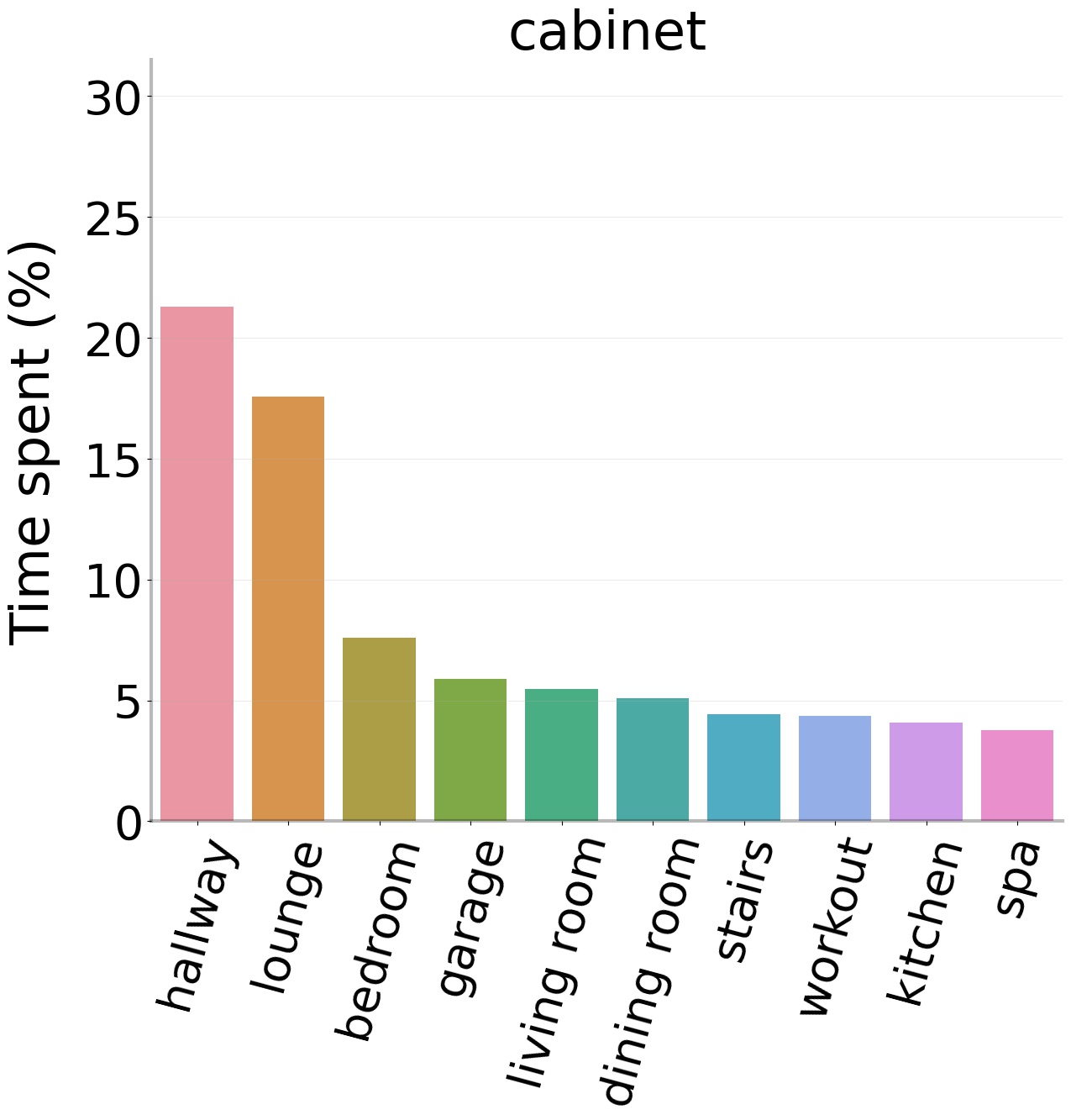}
            \caption{IL on $40k$ Human demos}
        \end{subfigure}
    \end{minipage}
    \hfill
    \begin{minipage}[a]{0.32\textwidth}
        \begin{subfigure}{\textwidth}
            \centering
            \includegraphics[width=\textwidth]{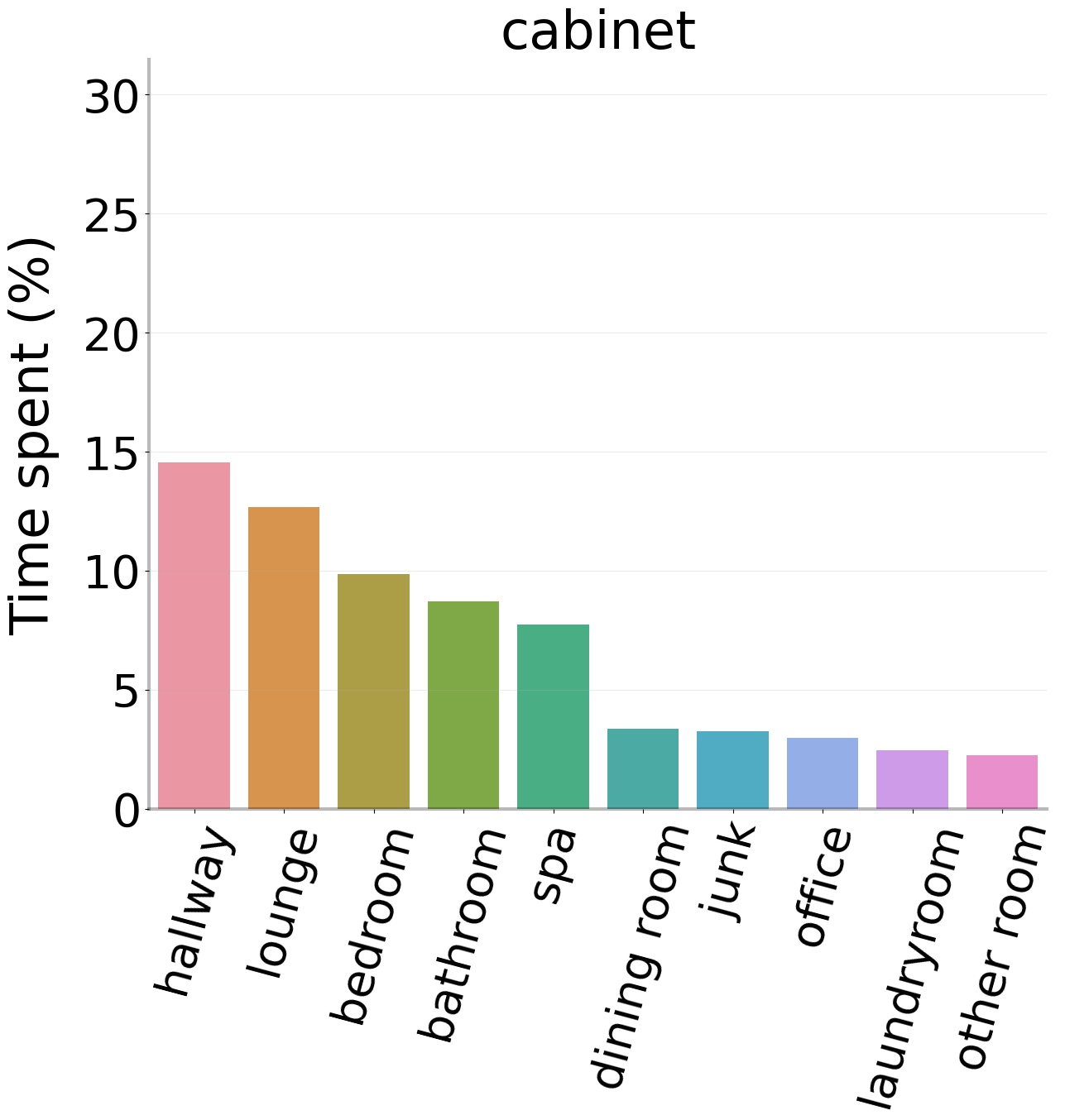}
            \caption{RL}
        \end{subfigure}
    \end{minipage}

    \caption{Comparison of per room time spent for all MP3D goal categories on \textsc{val} split for human demonstrations~\vs IL agents trained on human demos~\vs RL agents. The plot shows the top 10 rooms ordered by the maximum time spent in each room.}
    \label{fig:pRTS_per_object}
    
\end{figure*}

\begin{figure*}[t]\ContinuedFloat
    \centering
    \begin{minipage}[a]{0.32\textwidth}
        \begin{subfigure}{\textwidth}
            \centering
            \includegraphics[width=\textwidth]{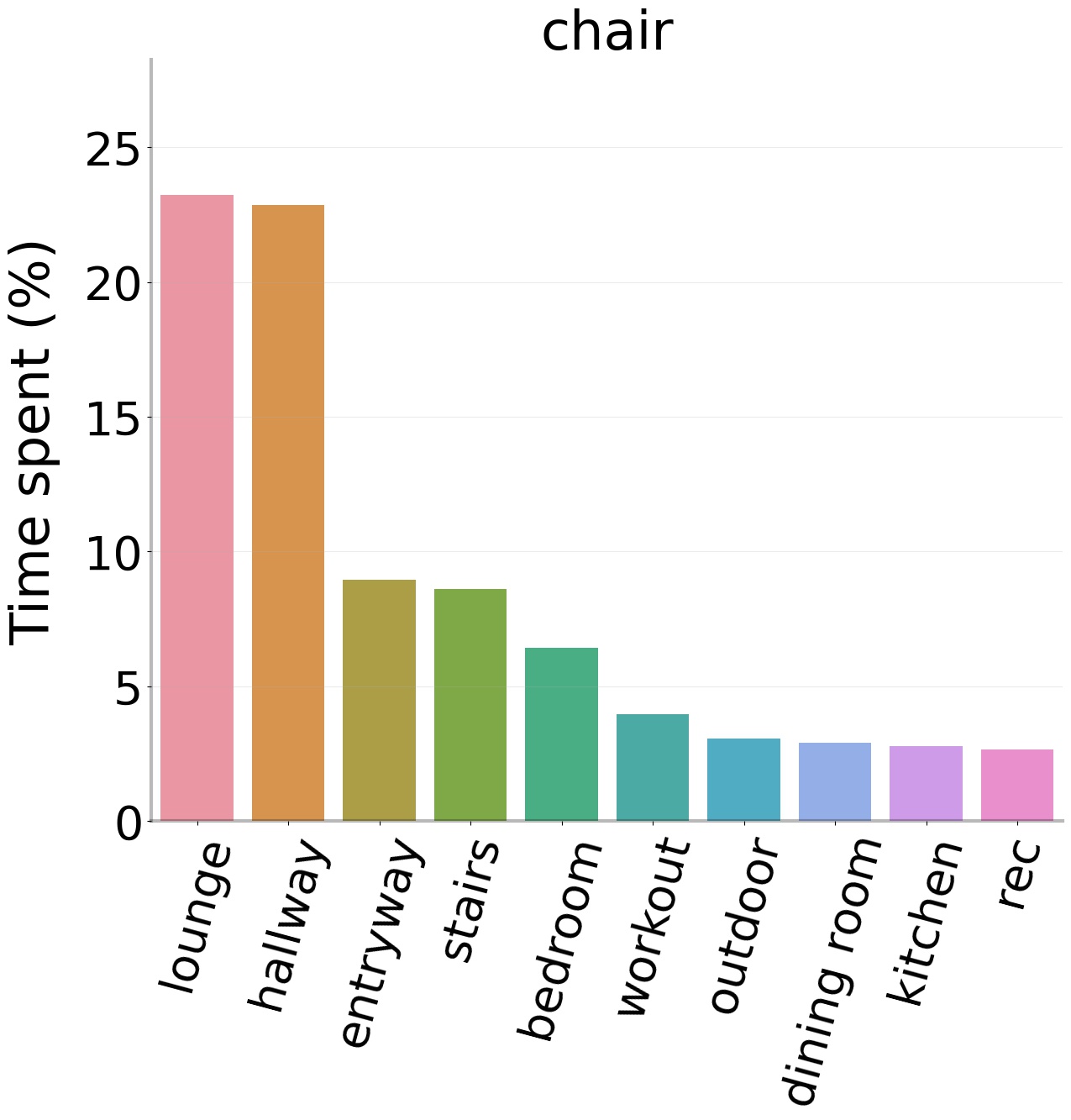}
            %\caption{Humans}
        \end{subfigure}
    \end{minipage}
    \hfill
    \begin{minipage}[a]{0.32\textwidth}
        \begin{subfigure}{\textwidth}
            \centering
            \includegraphics[width=\textwidth]{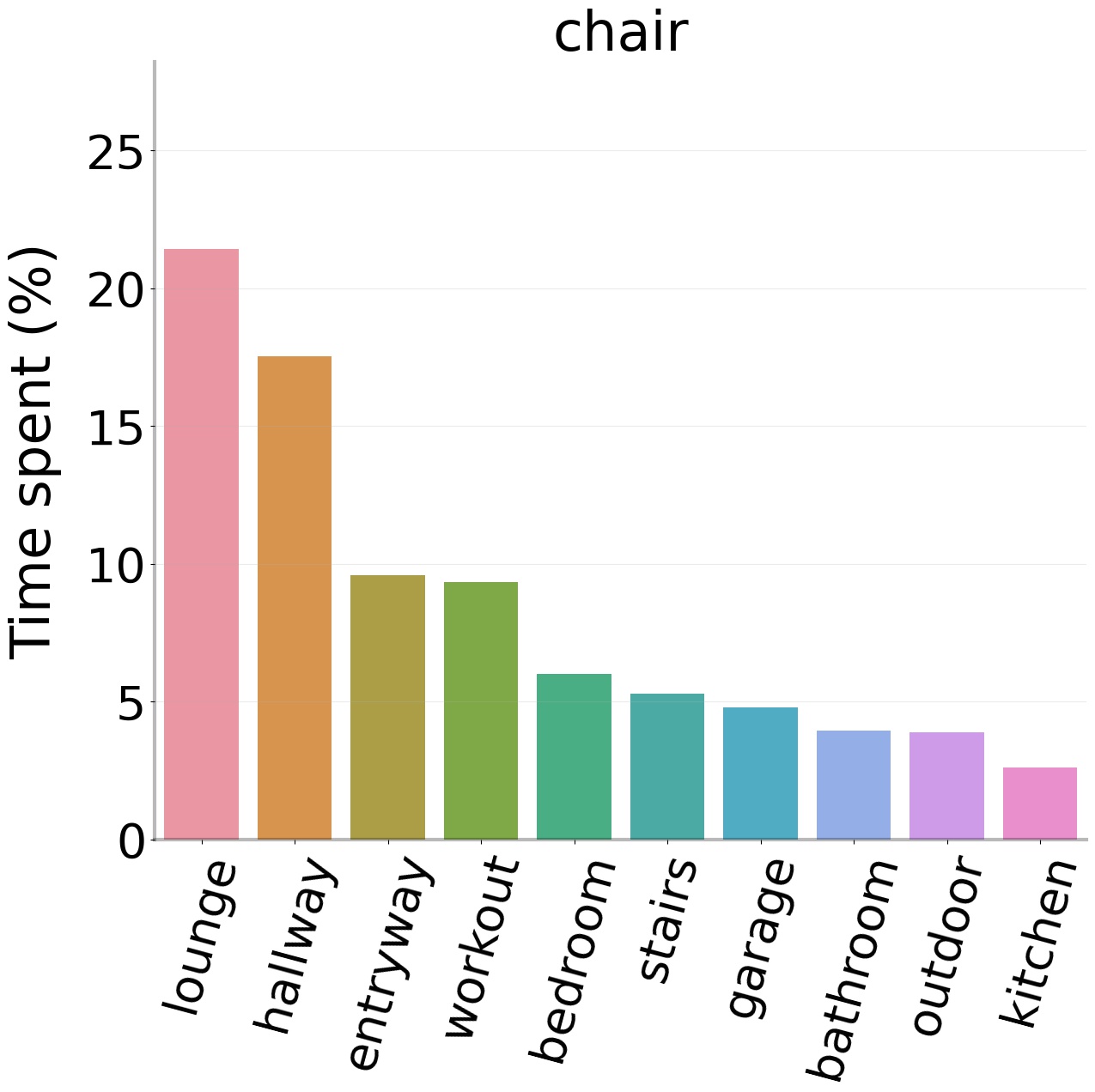}
            %\caption{IL on $40k$ Human demos}
        \end{subfigure}
    \end{minipage}
    \hfill
    \begin{minipage}[a]{0.32\textwidth}
        \begin{subfigure}{\textwidth}
            \centering
            \includegraphics[width=\textwidth]{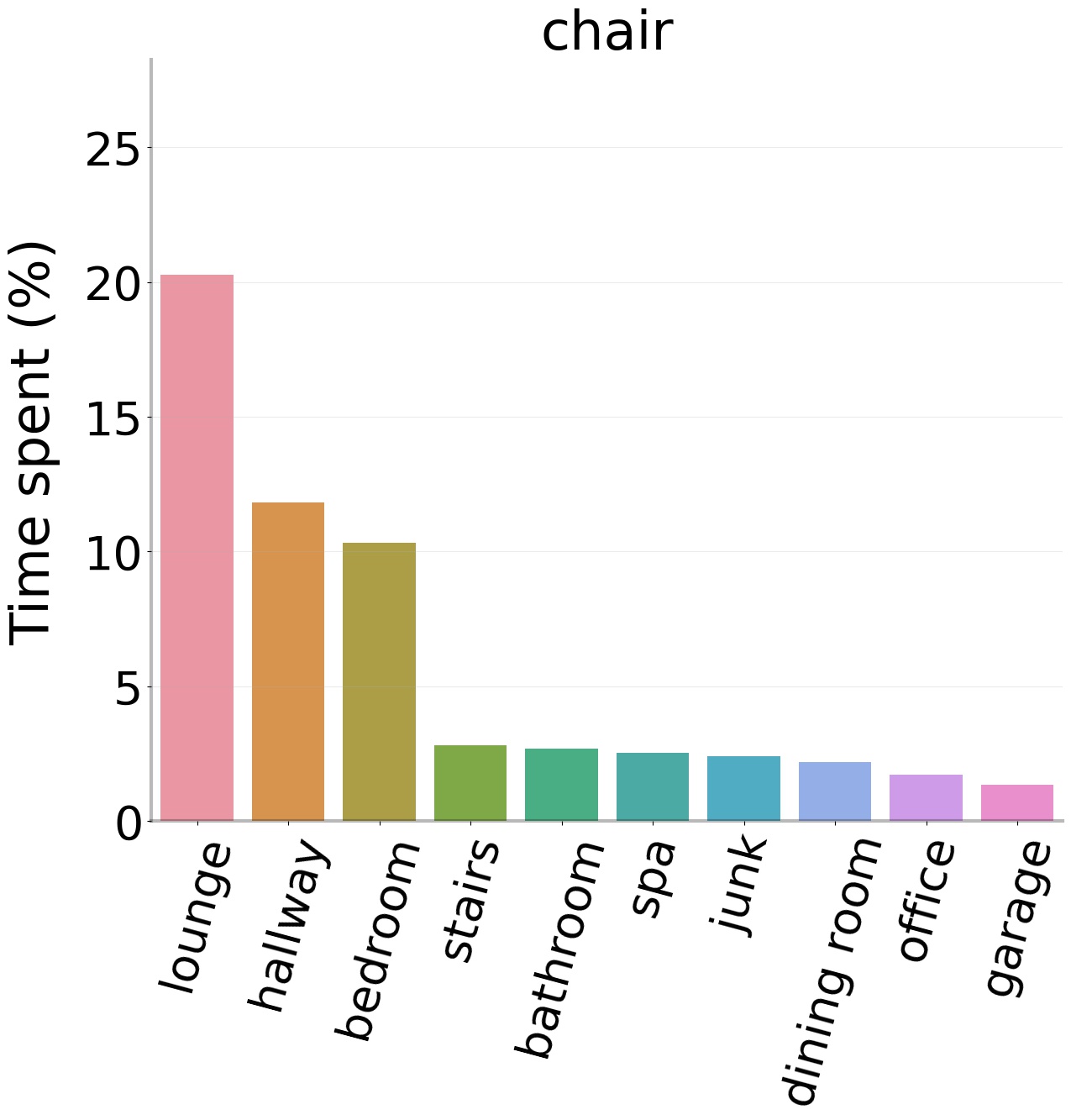}
            %\caption{RL}
        \end{subfigure}
    \end{minipage}

    \begin{minipage}[a]{0.32\textwidth}
        \begin{subfigure}{\textwidth}
            \centering
            \includegraphics[width=\textwidth]{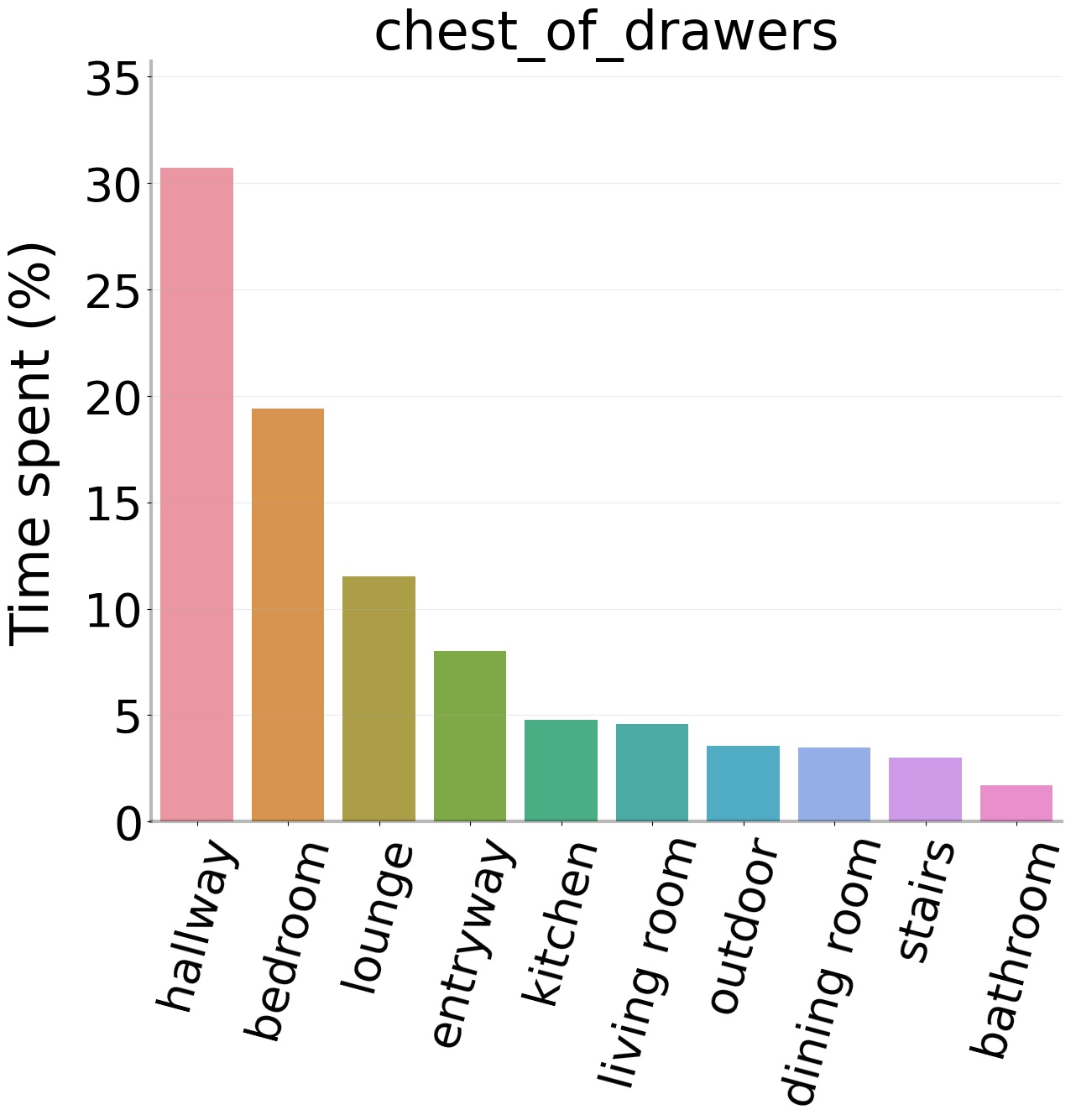}
            %\caption{Humans}
        \end{subfigure}
    \end{minipage}
    \hfill
    \begin{minipage}[a]{0.32\textwidth}
        \begin{subfigure}{\textwidth}
            \centering
            \includegraphics[width=\textwidth]{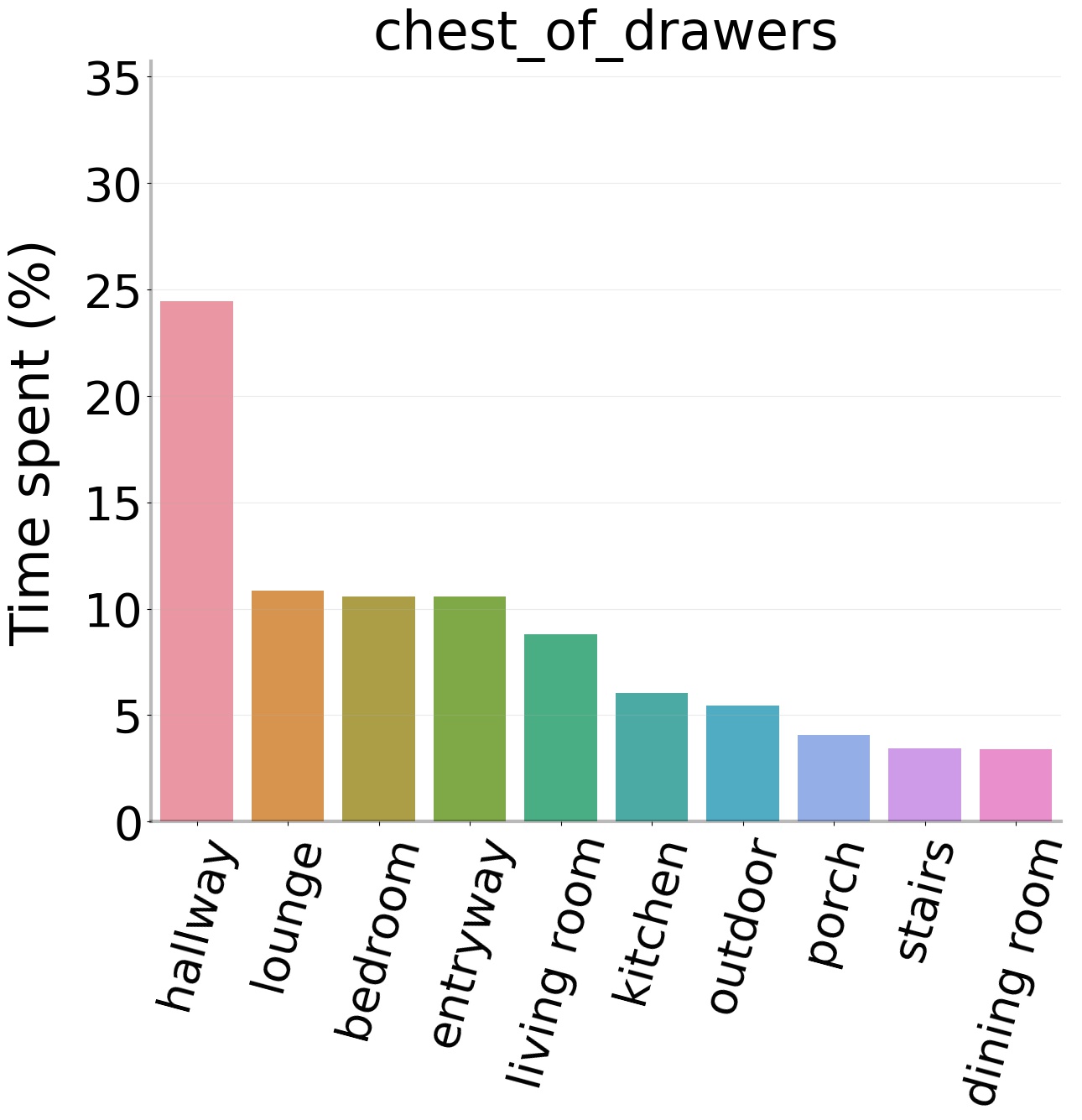}
            %\caption{IL on $40k$ Human demos}
        \end{subfigure}
    \end{minipage}
    \hfill
    \begin{minipage}[a]{0.32\textwidth}
        \begin{subfigure}{\textwidth}
            \centering
            \includegraphics[width=\textwidth]{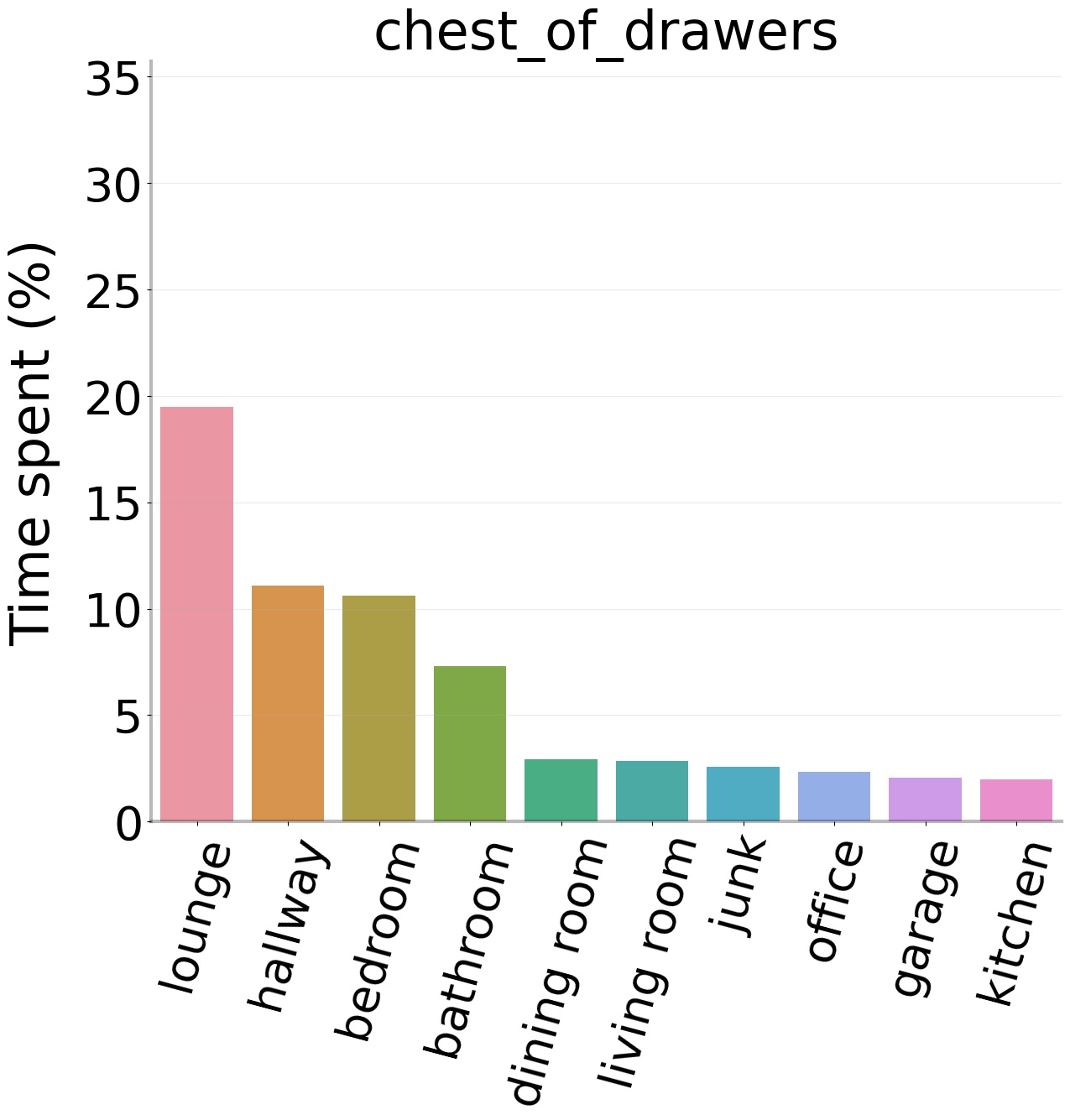}
            %\caption{RL}
        \end{subfigure}
    \end{minipage}

    \begin{minipage}[a]{0.32\textwidth}
        \begin{subfigure}{\textwidth}
            \centering
            \includegraphics[width=\textwidth]{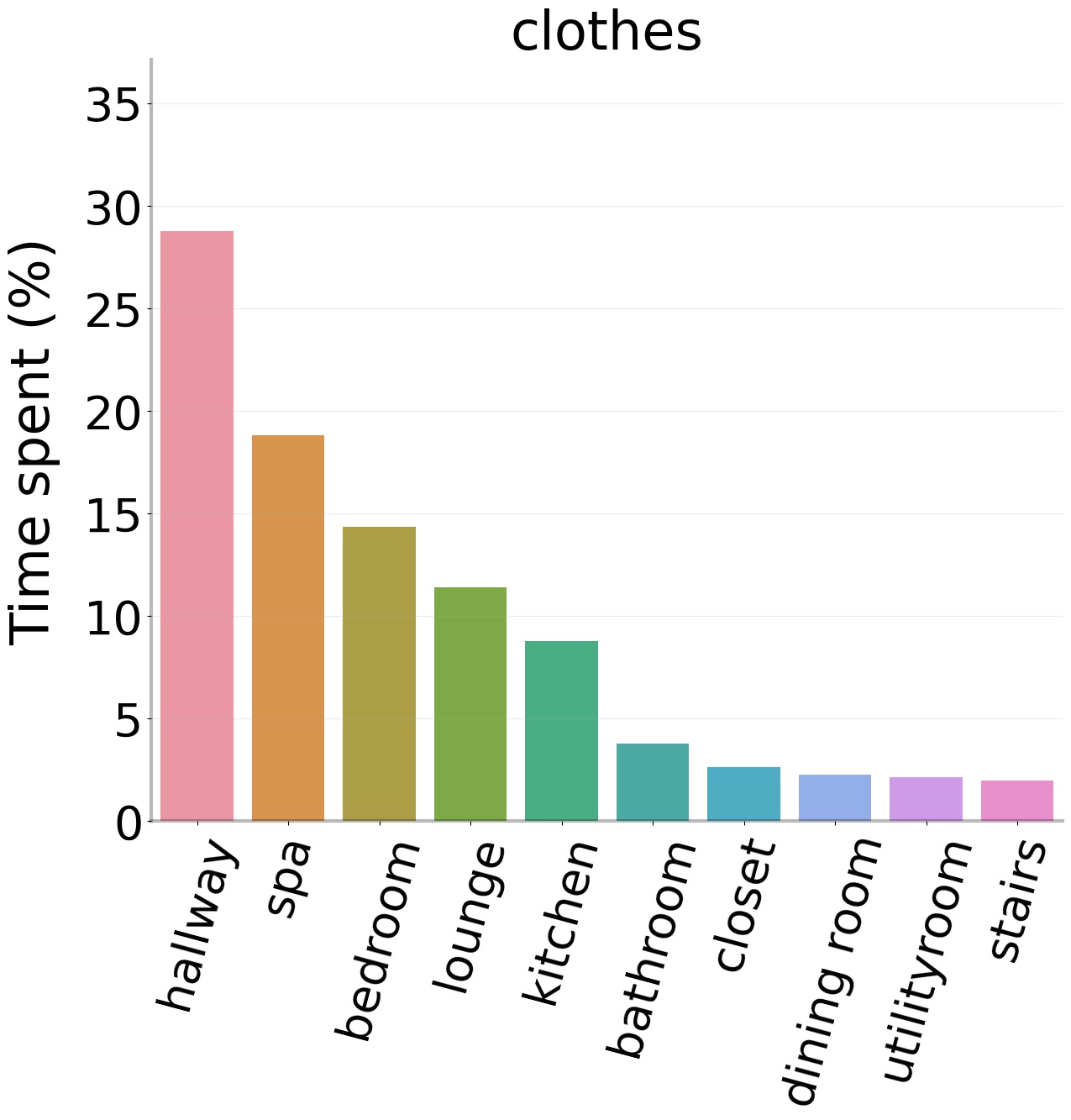}
            \caption{Humans}
        \end{subfigure}
    \end{minipage}
    \hfill
    \begin{minipage}[a]{0.32\textwidth}
        \begin{subfigure}{\textwidth}
            \centering
            \includegraphics[width=\textwidth]{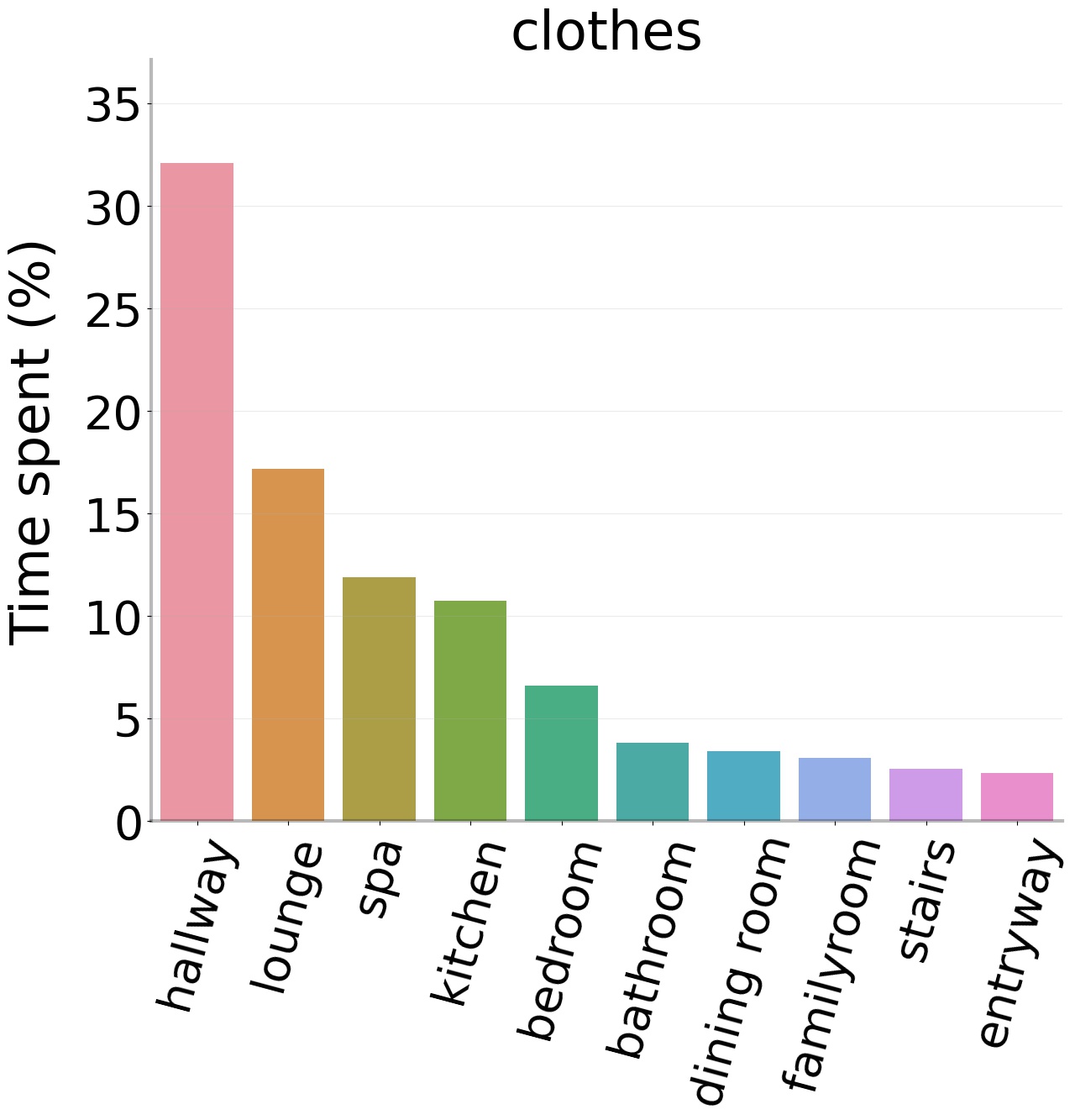}
            \caption{IL on $40k$ Human demos}
        \end{subfigure}
    \end{minipage}
    \hfill
    \begin{minipage}[a]{0.32\textwidth}
        \begin{subfigure}{\textwidth}
            \centering
            \includegraphics[width=\textwidth]{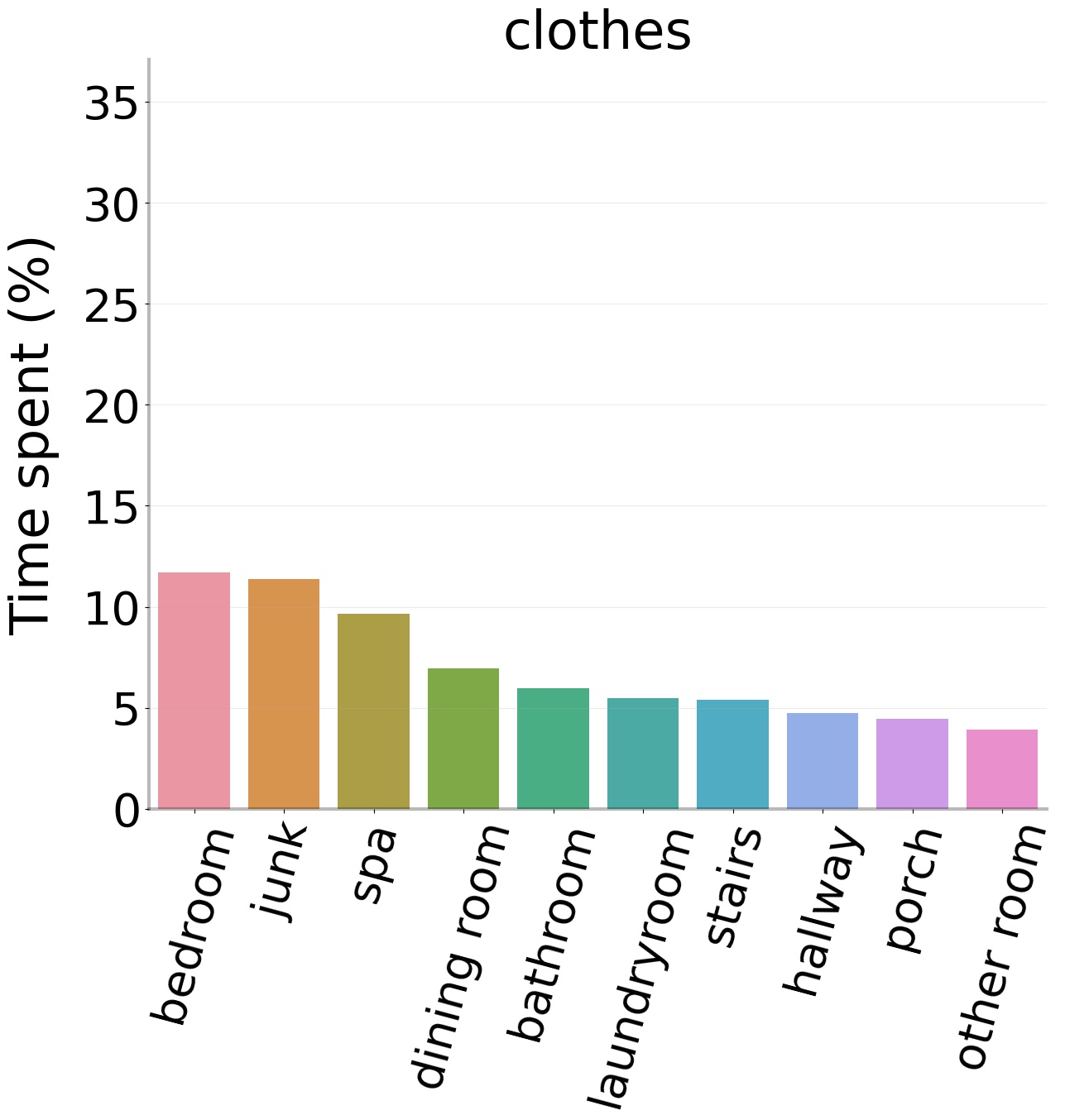}
            \caption{RL}
        \end{subfigure}
    \end{minipage}

    \caption{Comparison of per room time spent for all MP3D goal categories on \textsc{val} split for human demonstrations~\vs IL agents trained on human demos~\vs RL agents. The plot shows the top 10 rooms ordered by the maximum time spent in each room.}
    \label{fig:pRTS_per_object_2}
    
\end{figure*}

\begin{figure*}[t]\ContinuedFloat
    %\centering
    \begin{minipage}[a]{0.32\textwidth}
        \begin{subfigure}{\textwidth}
            \centering
            \includegraphics[width=\textwidth]{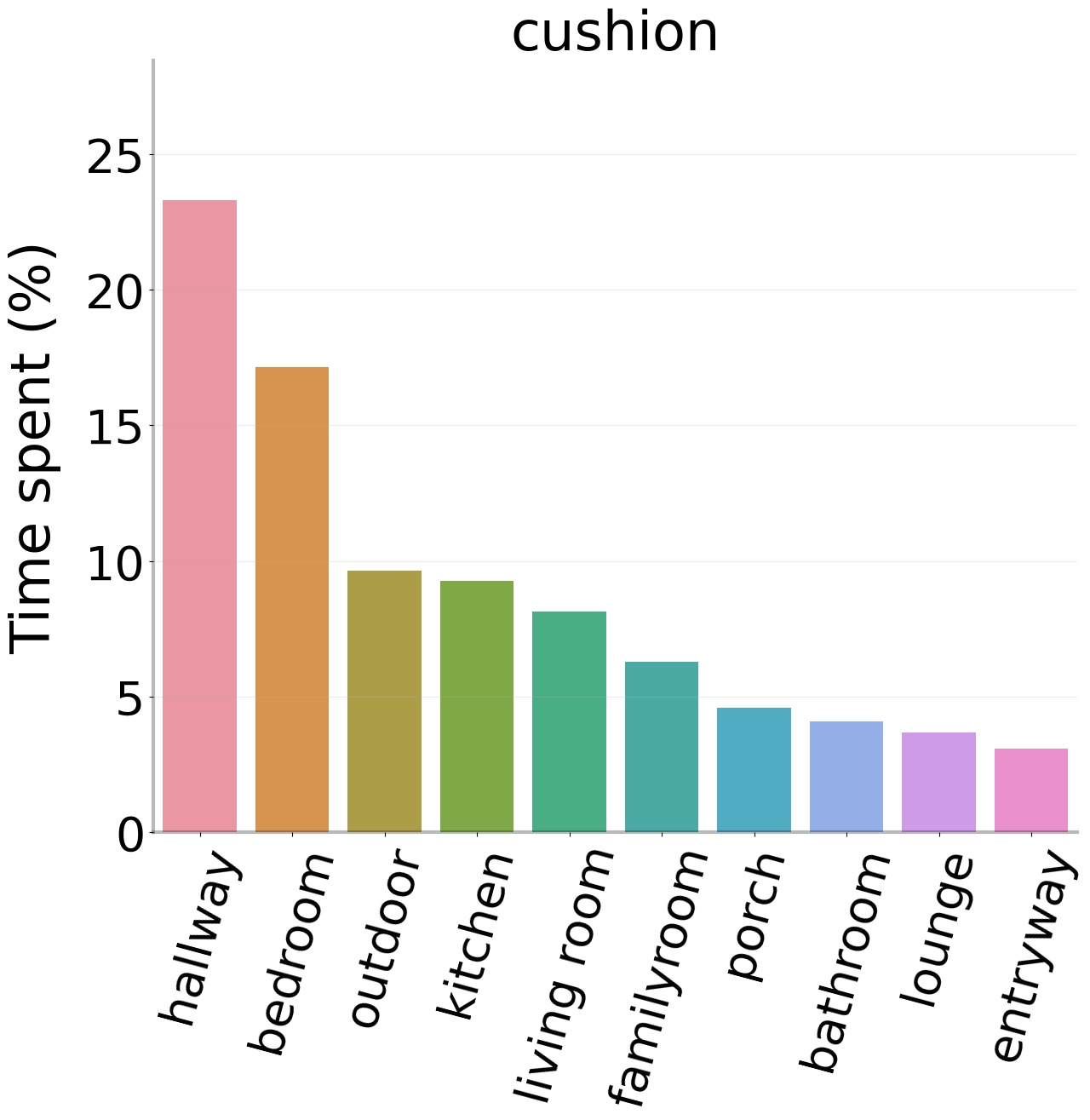}
            %\caption{Humans}
        \end{subfigure}
    \end{minipage}
    \hfill
    \begin{minipage}[a]{0.32\textwidth}
        \begin{subfigure}{\textwidth}
            \centering
            \includegraphics[width=\textwidth]{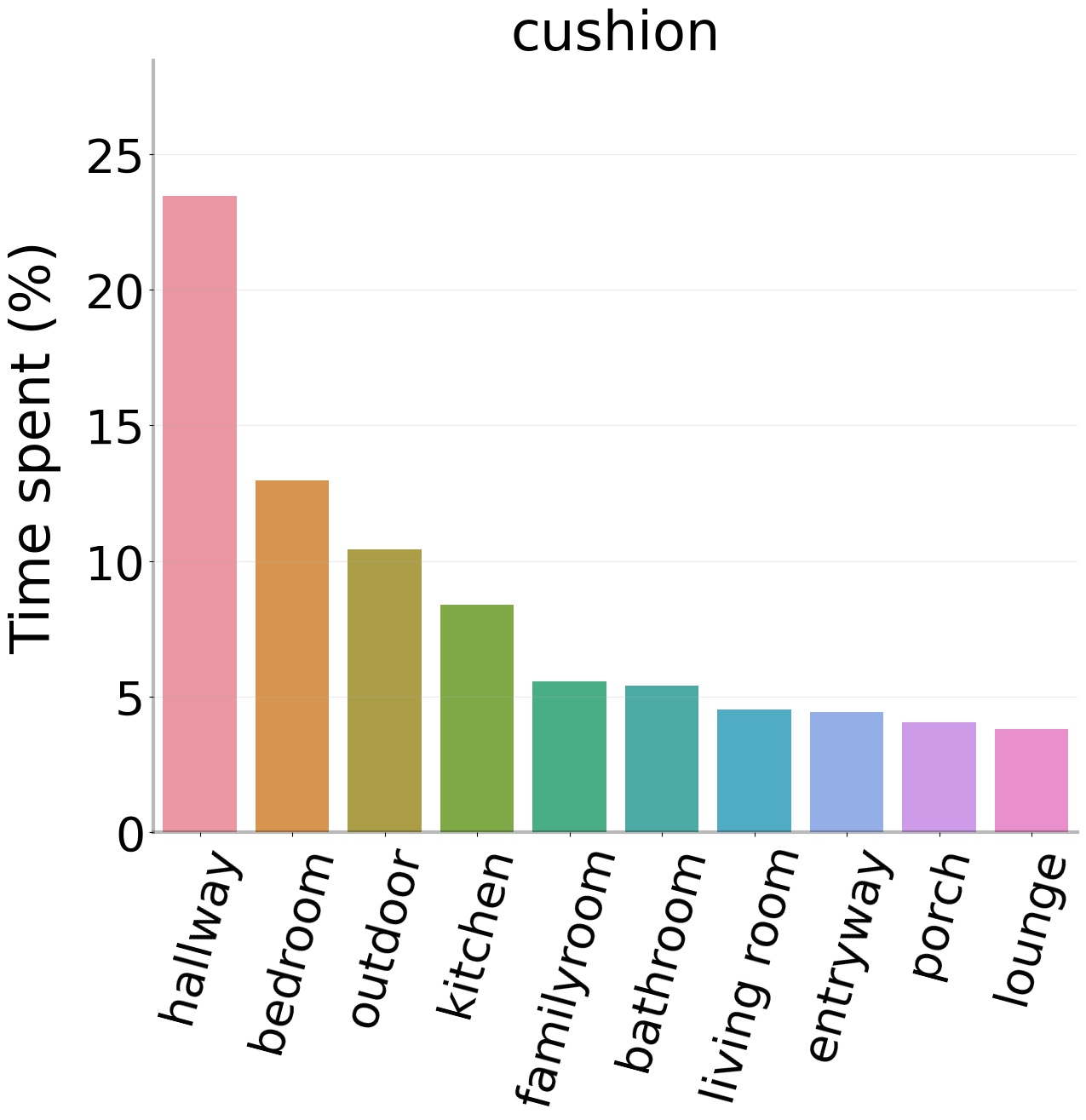}
            %\caption{IL on $40k$ Human demos}
        \end{subfigure}
    \end{minipage}
    \hfill
    \begin{minipage}[a]{0.32\textwidth}
        \begin{subfigure}{\textwidth}
            \centering
            \includegraphics[width=\textwidth]{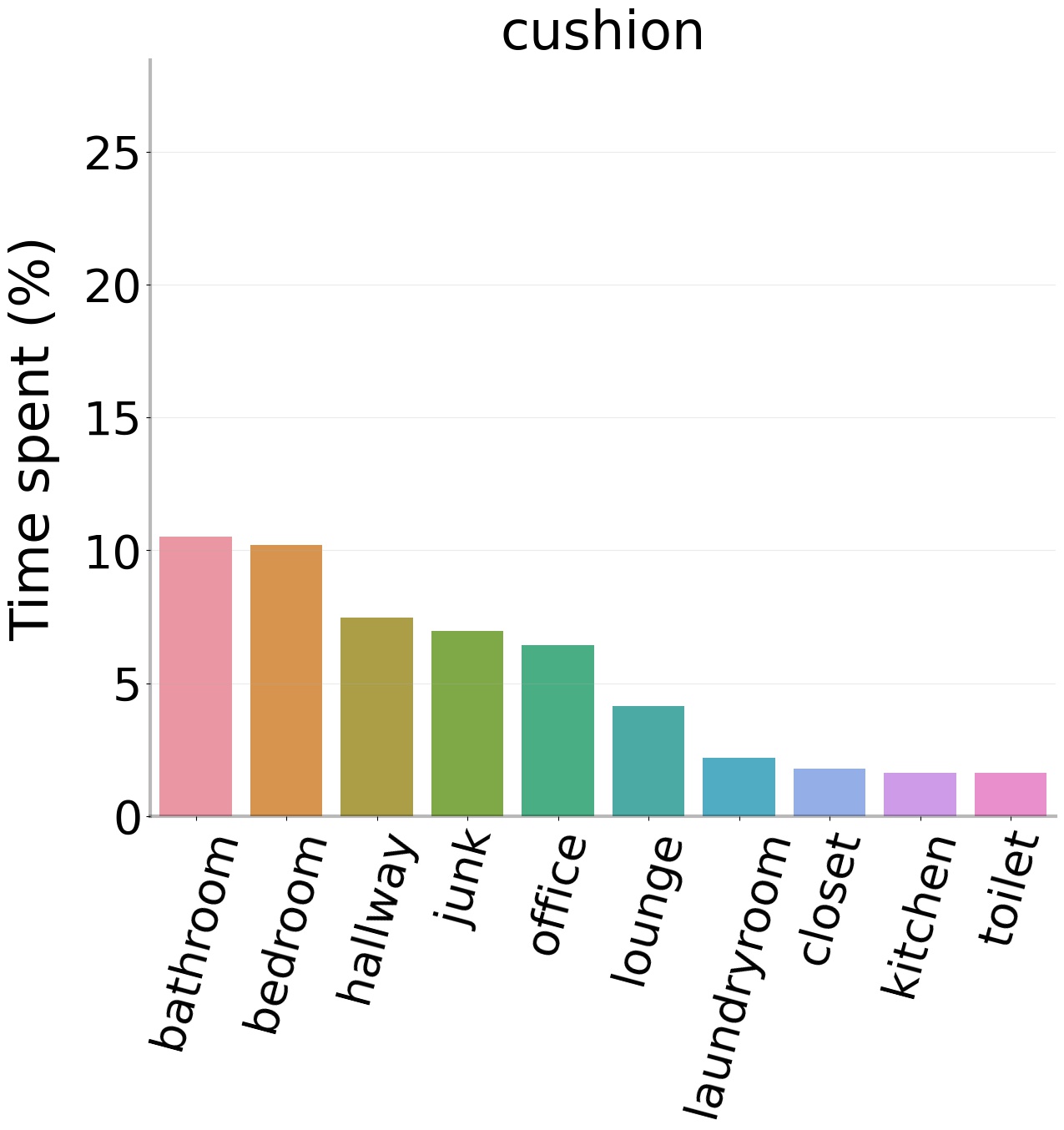}
            %\caption{RL}
        \end{subfigure}
    \end{minipage}

    \begin{minipage}[a]{0.32\textwidth}
        \begin{subfigure}{\textwidth}
            \centering
            \includegraphics[width=\textwidth]{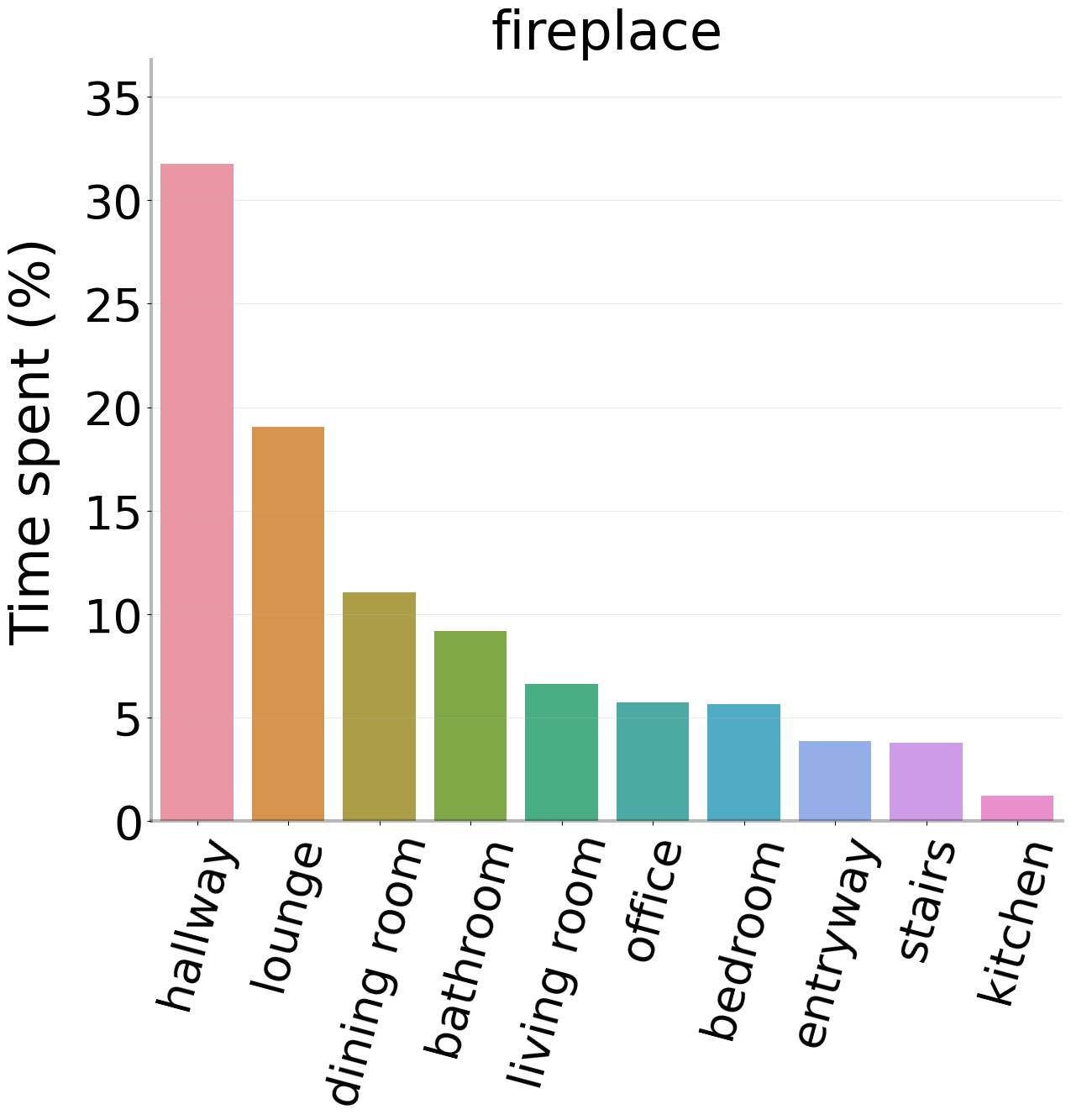}
            %\caption{Humans}
        \end{subfigure}
    \end{minipage}
    \hfill
    \begin{minipage}[a]{0.32\textwidth}
        \begin{subfigure}{\textwidth}
            \centering
            \includegraphics[width=\textwidth]{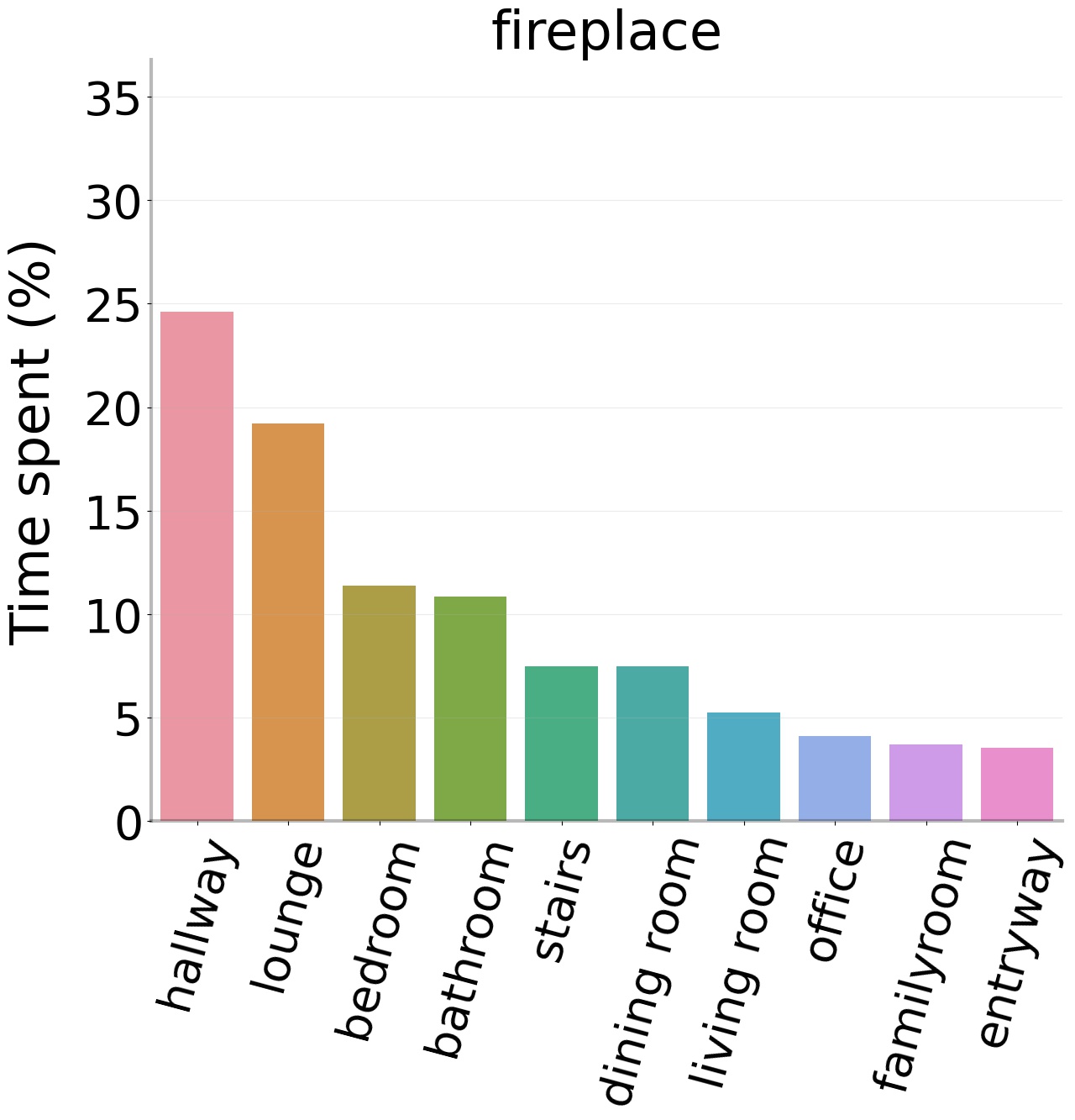}
            %\caption{IL on $40k$ Human demos}
        \end{subfigure}
    \end{minipage}
    \hfill
    \begin{minipage}[a]{0.32\textwidth}
        \begin{subfigure}{\textwidth}
            \centering
            \includegraphics[width=\textwidth]{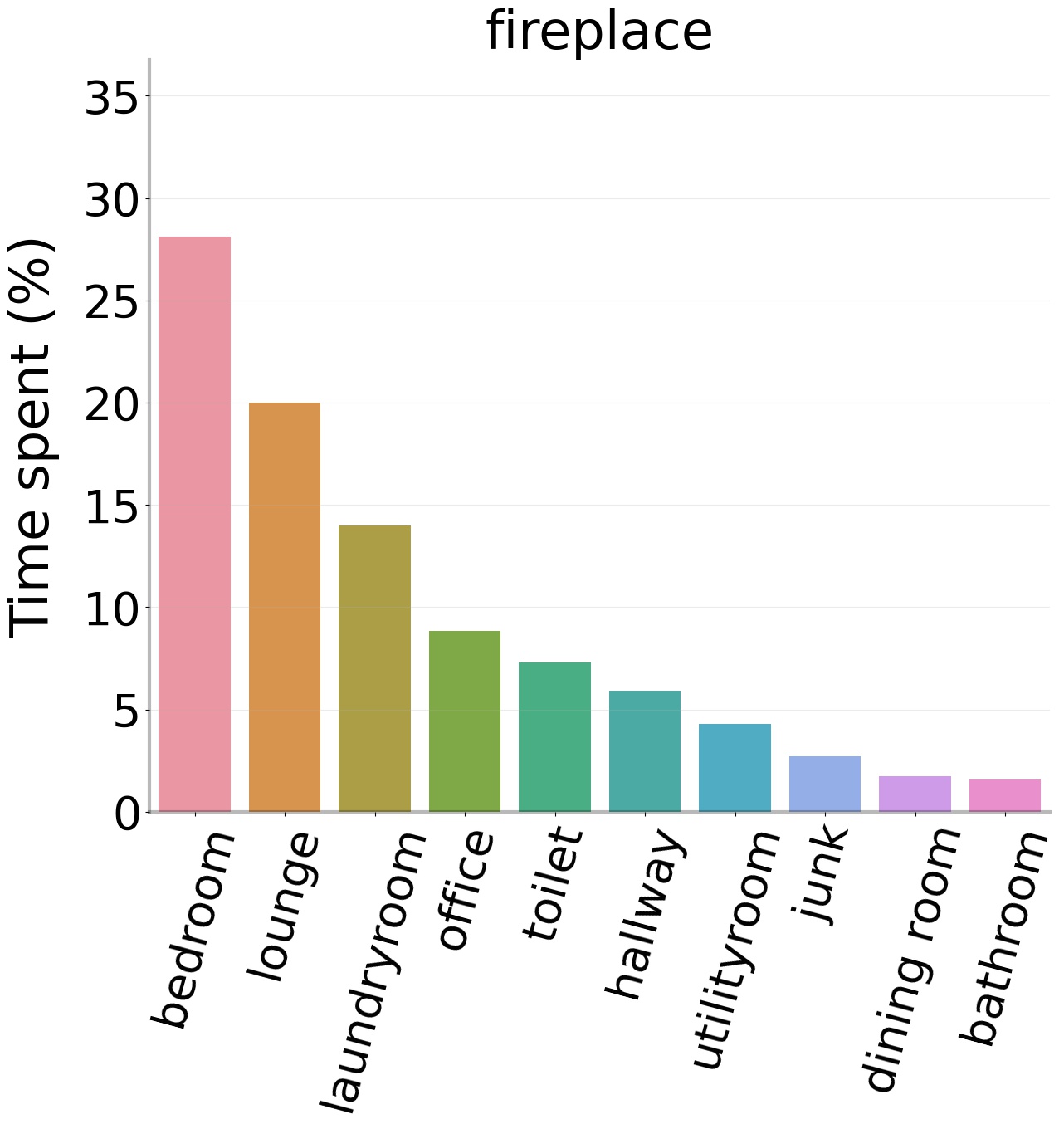}
            %\caption{RL}
        \end{subfigure}
    \end{minipage}

    \begin{minipage}[a]{0.32\textwidth}
        \begin{subfigure}{\textwidth}
            \centering
            \includegraphics[width=\textwidth]{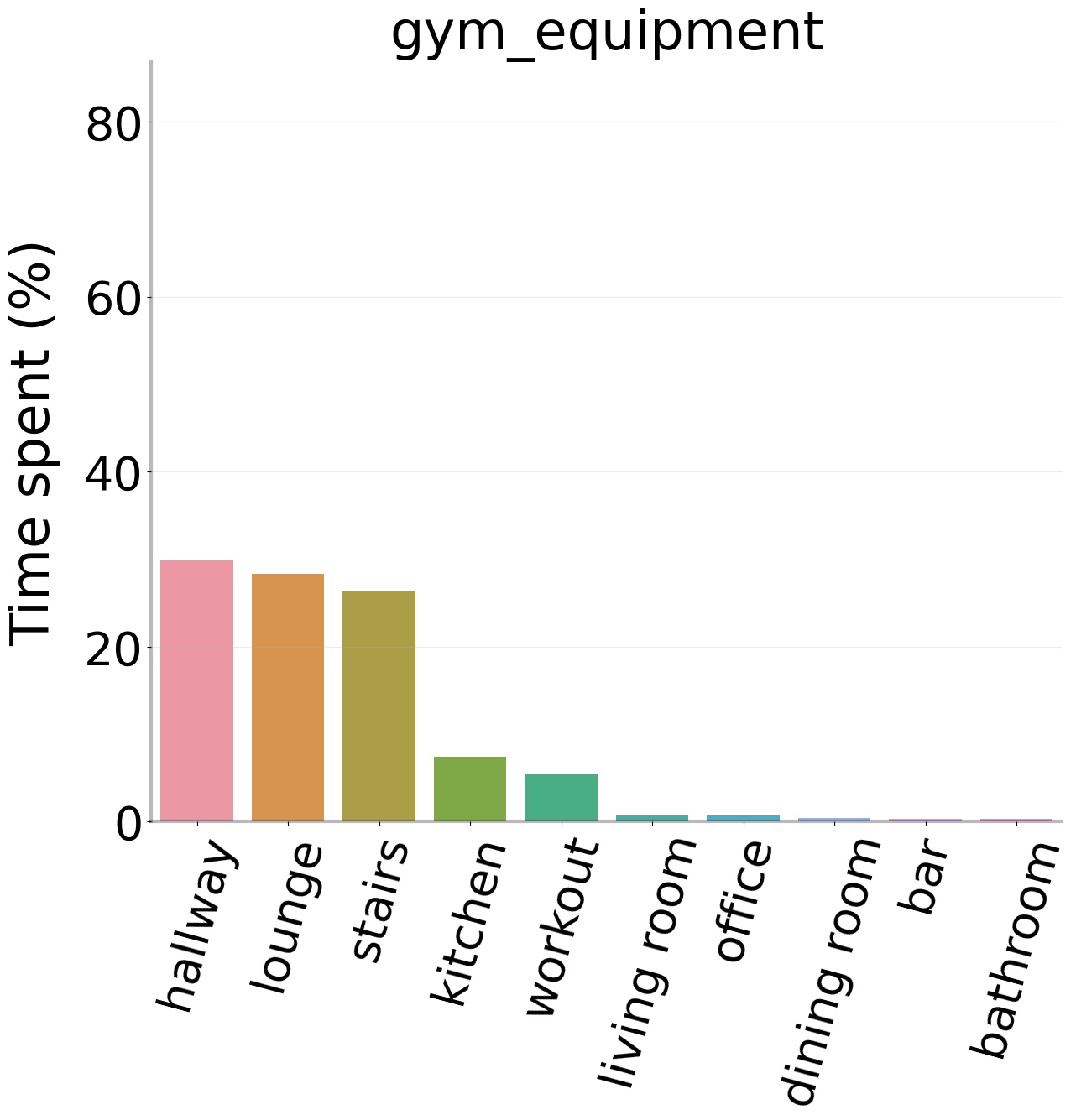}
            \caption{Humans}
        \end{subfigure}
    \end{minipage}
    \hfill
    \begin{minipage}[a]{0.32\textwidth}
        \begin{subfigure}{\textwidth}
            \centering
            \includegraphics[width=\textwidth]{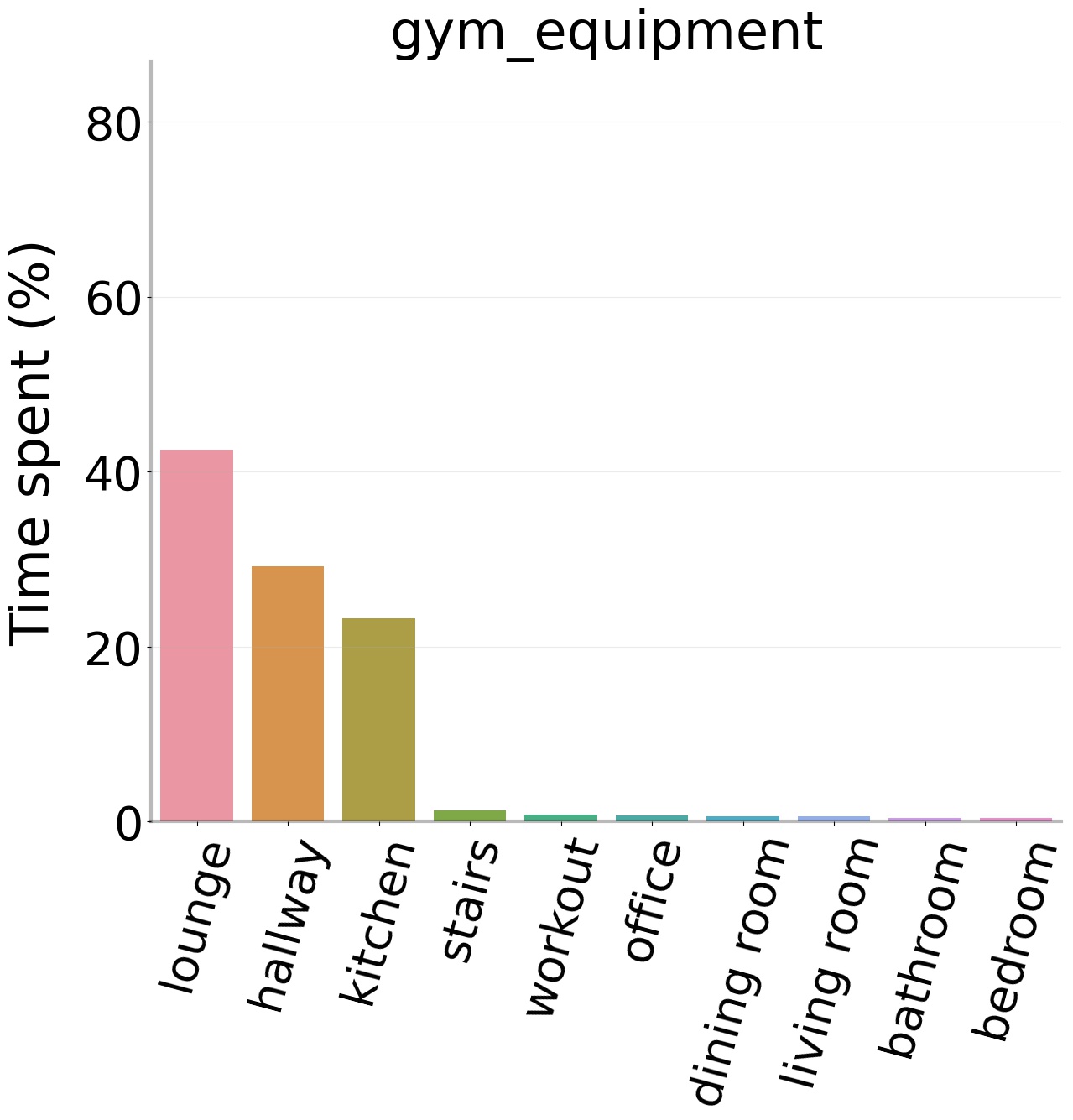}
            \caption{IL on $40k$ Human demos}
        \end{subfigure}
    \end{minipage}
    \hfill
    \begin{minipage}[a]{0.32\textwidth}
        \begin{subfigure}{\textwidth}
            \centering
            \includegraphics[width=\textwidth]{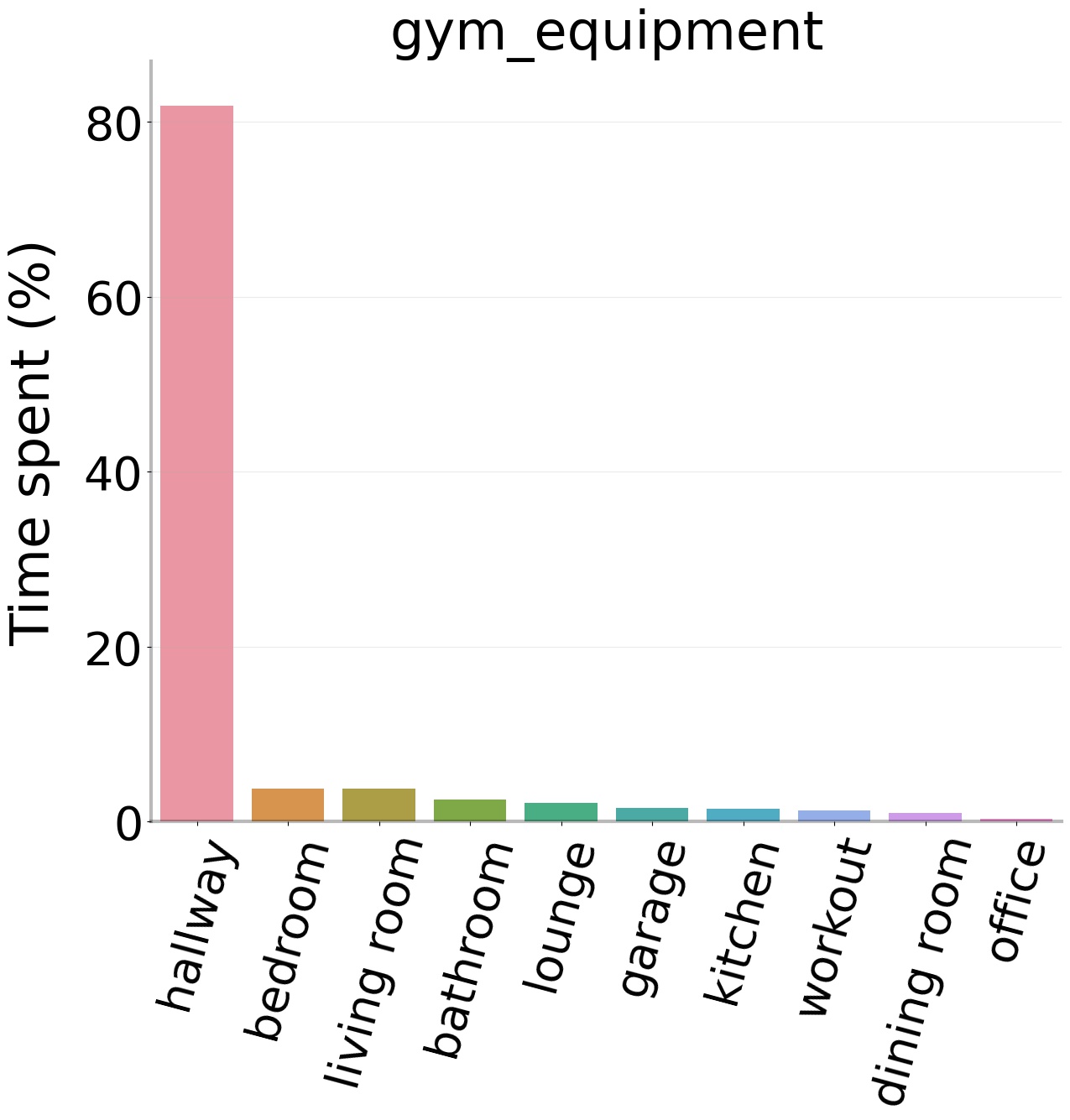}
            \caption{RL}
        \end{subfigure}
    \end{minipage}

    \caption{Comparison of per room time spent for all MP3D goal categories on \textsc{val} split for human demonstrations~\vs IL agents trained on human demos~\vs RL agents. The plot shows the top 10 rooms ordered by the maximum time spent in each room.}
    \label{fig:pRTS_per_object_3}
    
\end{figure*}

\begin{figure*}[t]\ContinuedFloat
    \centering
    \begin{minipage}[a]{0.32\textwidth}
        \begin{subfigure}{\textwidth}
            \centering
            \includegraphics[width=\textwidth]{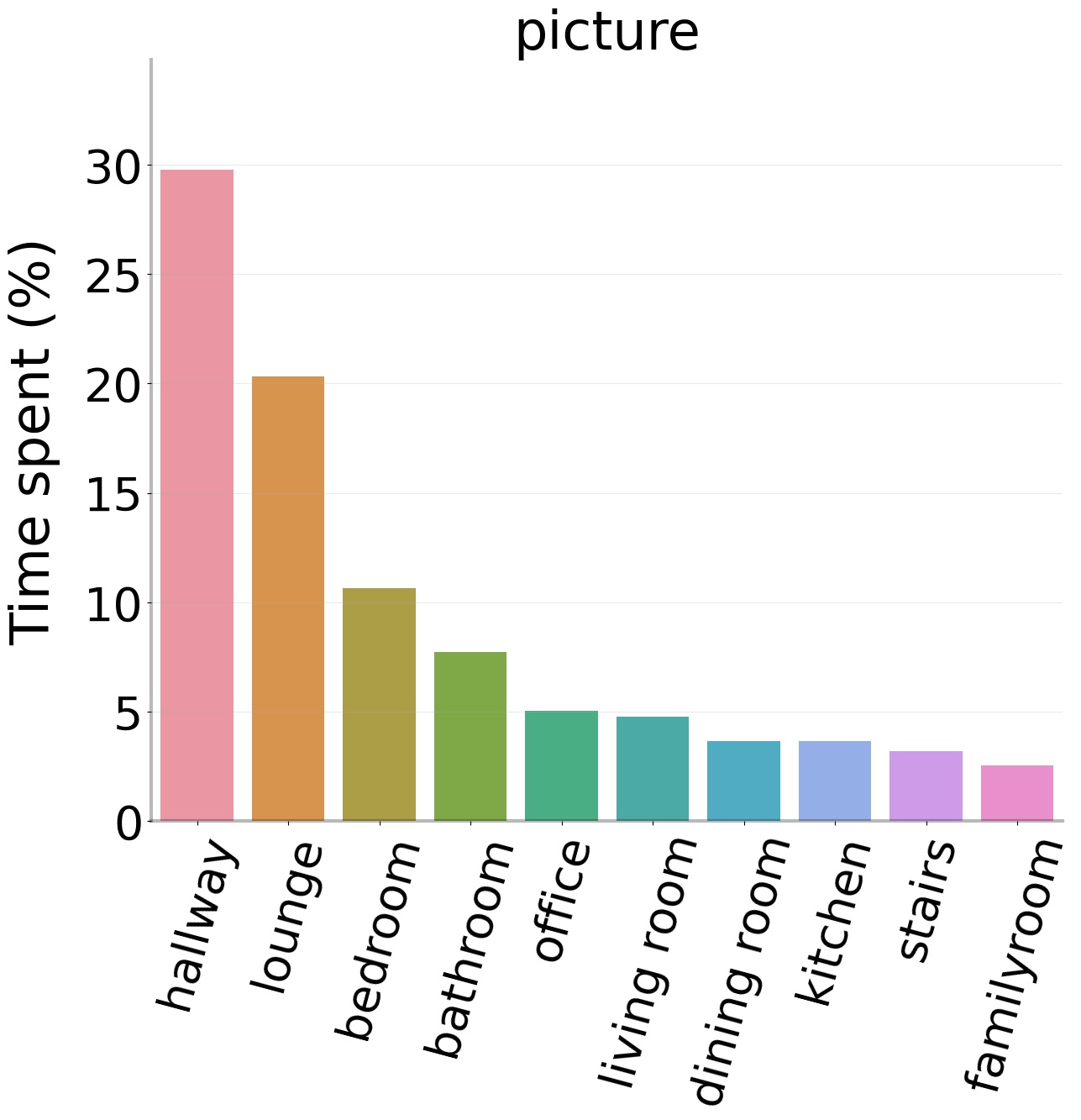}
            %\caption{Humans}
        \end{subfigure}
    \end{minipage}
    \hfill
    \begin{minipage}[a]{0.32\textwidth}
        \begin{subfigure}{\textwidth}
            \centering
            \includegraphics[width=\textwidth]{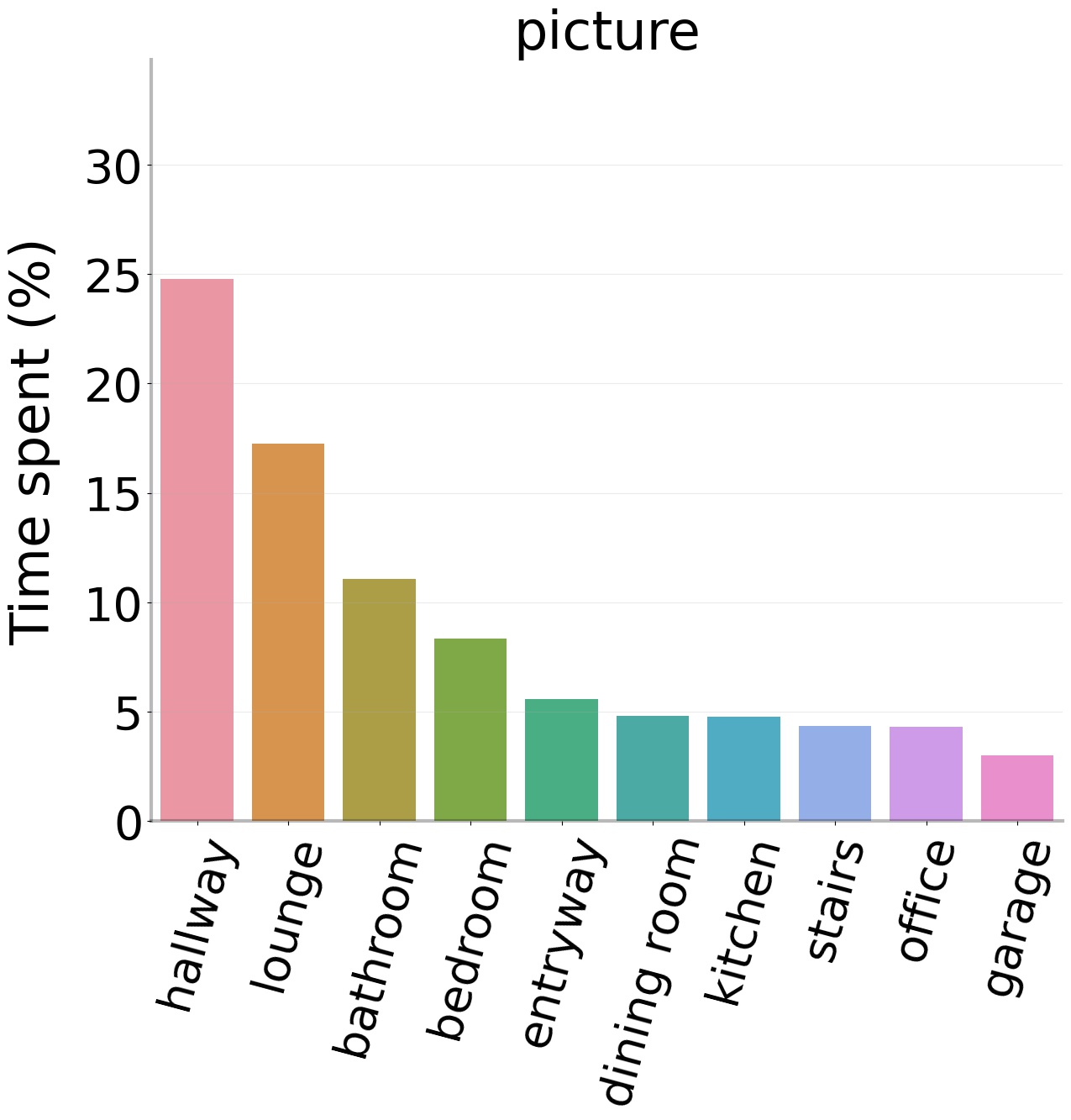}
            %\caption{IL on $40k$ Human demos}
        \end{subfigure}
    \end{minipage}
    \hfill
    \begin{minipage}[a]{0.32\textwidth}
        \begin{subfigure}{\textwidth}
            \centering
            \includegraphics[width=\textwidth]{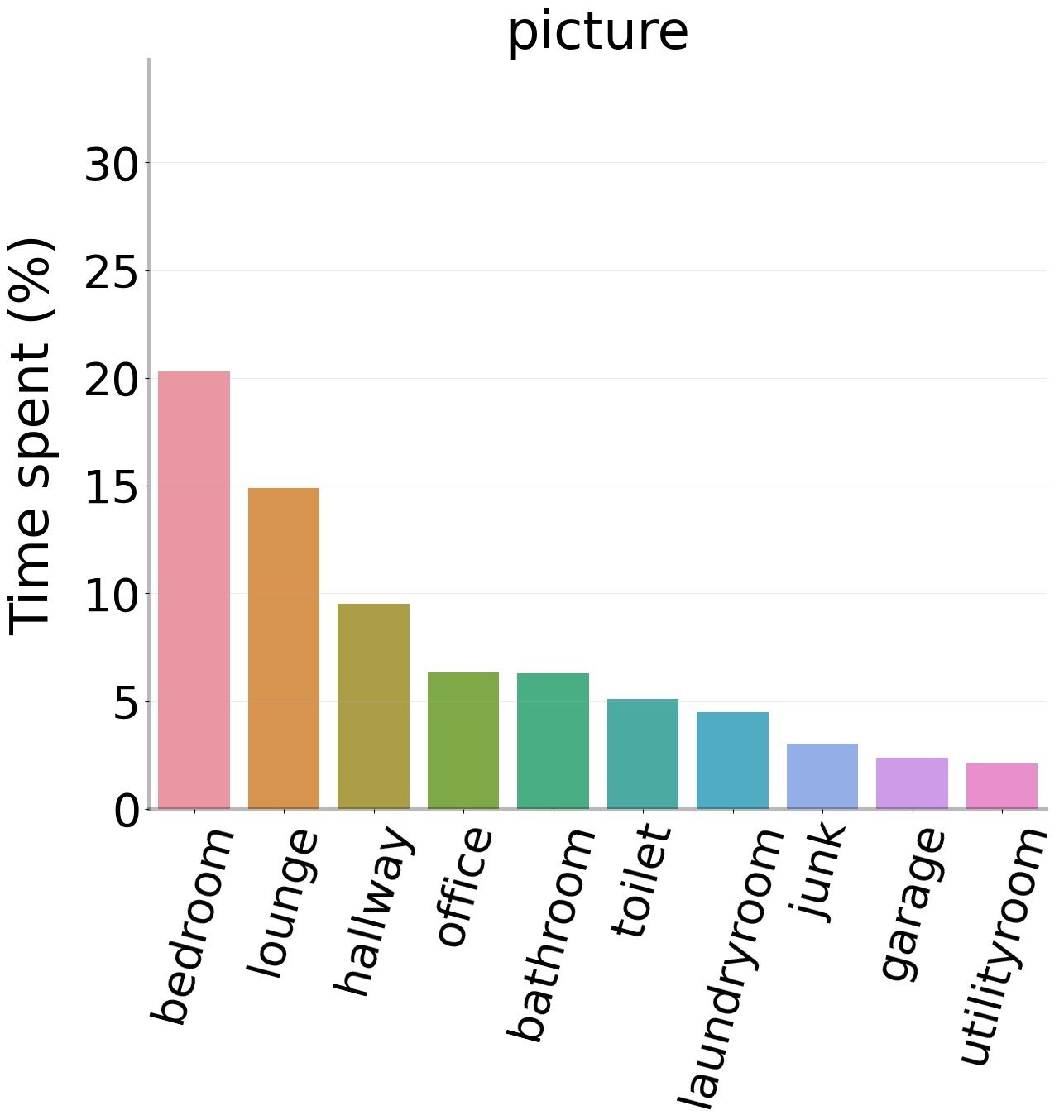}
            %\caption{RL}
        \end{subfigure}
    \end{minipage}

    \begin{minipage}[a]{0.32\textwidth}
        \begin{subfigure}{\textwidth}
            \centering
            \includegraphics[width=\textwidth]{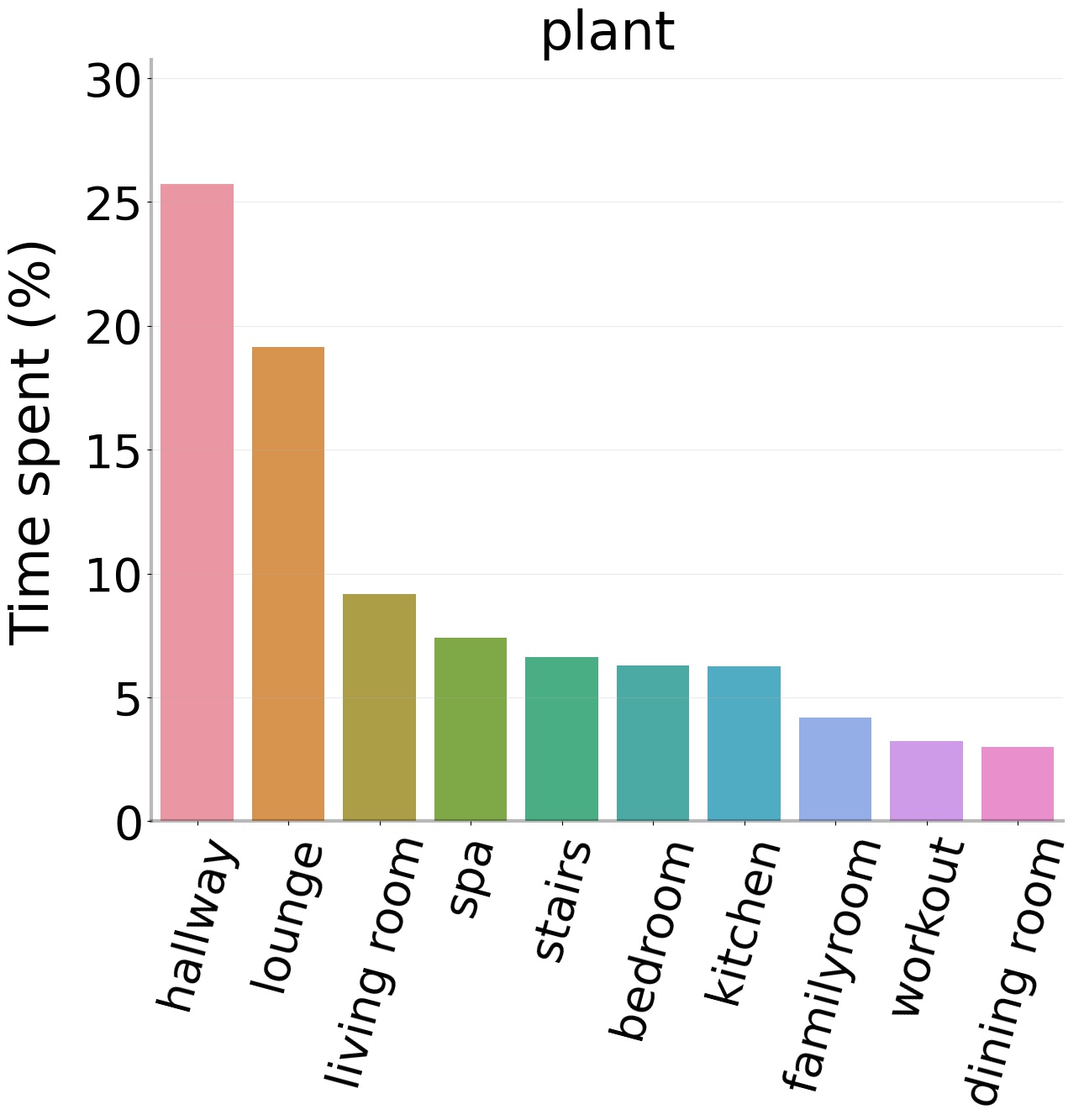}
            %\caption{Humans}
        \end{subfigure}
    \end{minipage}
    \hfill
    \begin{minipage}[a]{0.32\textwidth}
        \begin{subfigure}{\textwidth}
            \centering
            \includegraphics[width=\textwidth]{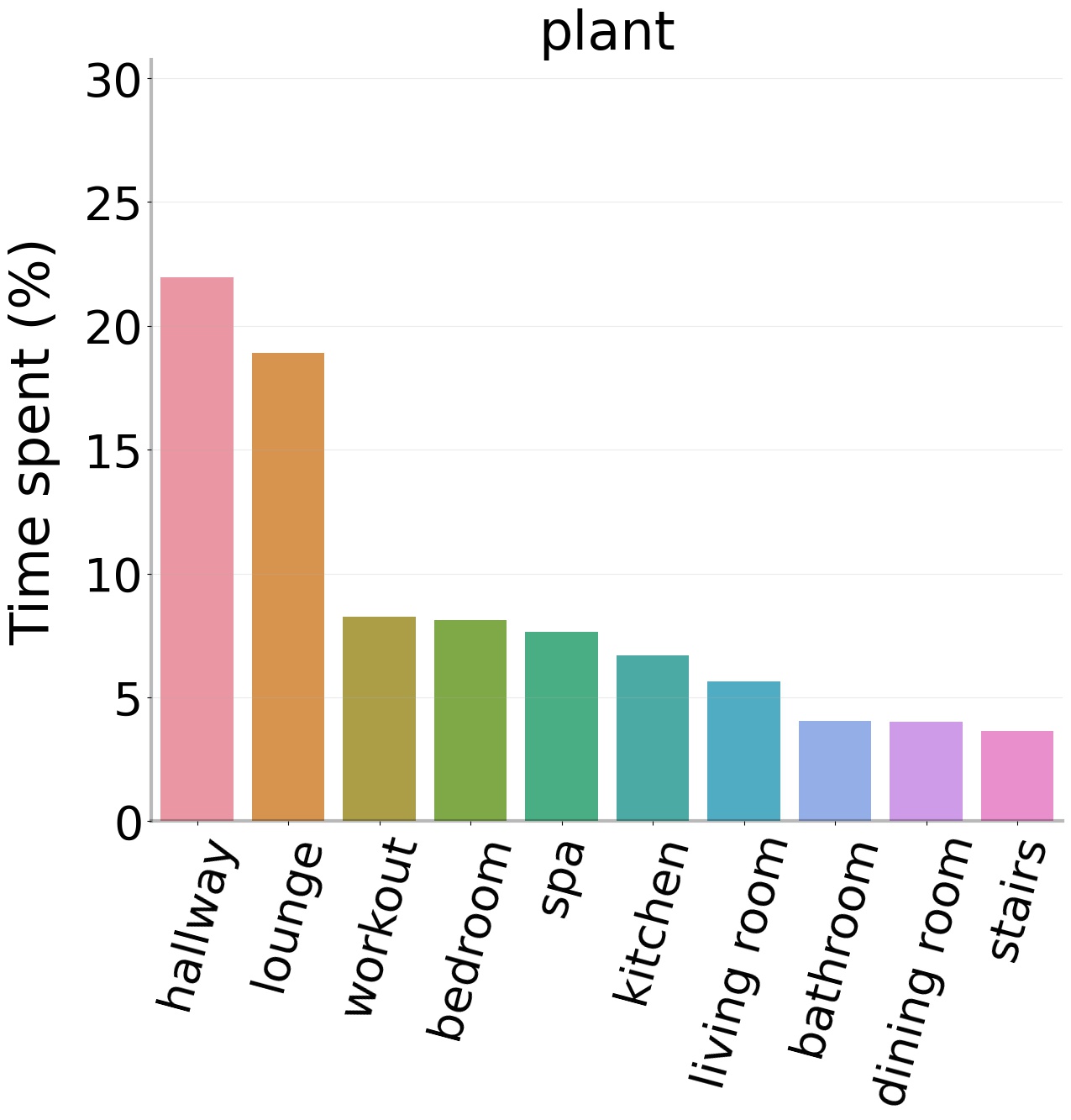}
            %\caption{IL on $40k$ Human demos}
        \end{subfigure}
    \end{minipage}
    \hfill
    \begin{minipage}[a]{0.32\textwidth}
        \begin{subfigure}{\textwidth}
            \centering
            \includegraphics[width=\textwidth]{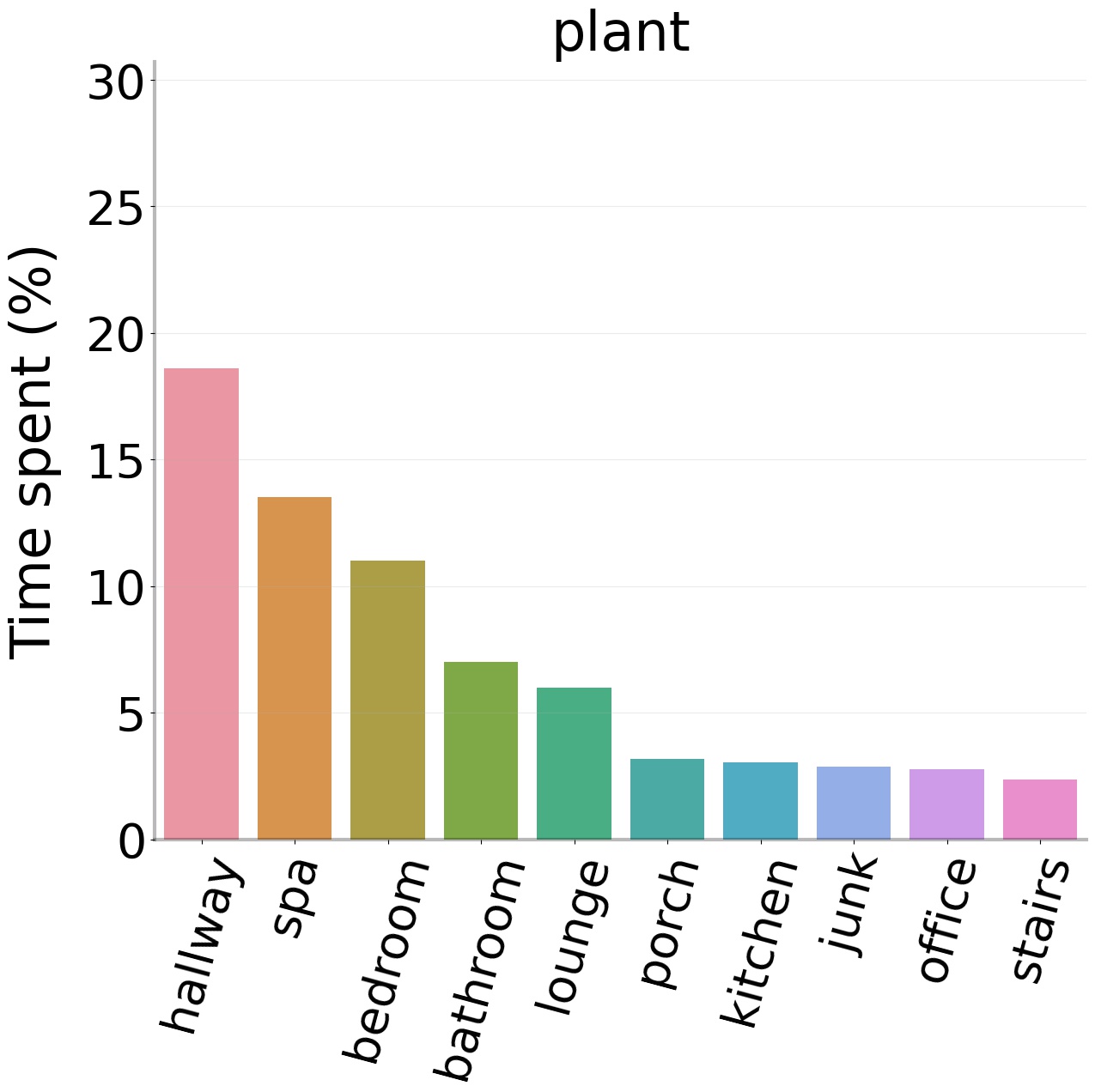}
            %\caption{RL}
        \end{subfigure}
    \end{minipage}

    \begin{minipage}[a]{0.32\textwidth}
        \begin{subfigure}{\textwidth}
            \centering
            \includegraphics[width=\textwidth]{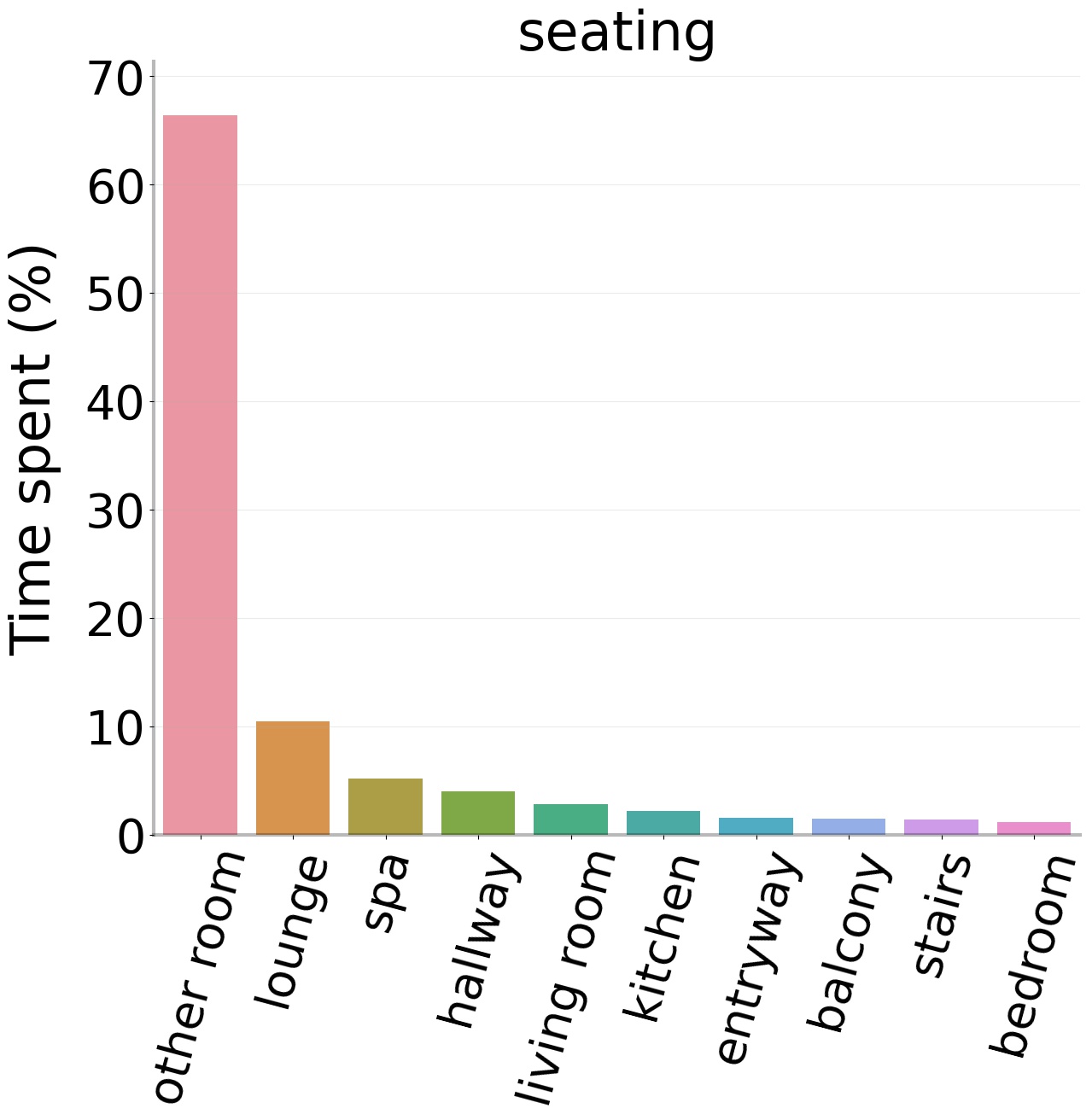}
            \caption{Humans}
        \end{subfigure}
    \end{minipage}
    \hfill
    \begin{minipage}[a]{0.32\textwidth}
        \begin{subfigure}{\textwidth}
            \centering
            \includegraphics[width=\textwidth]{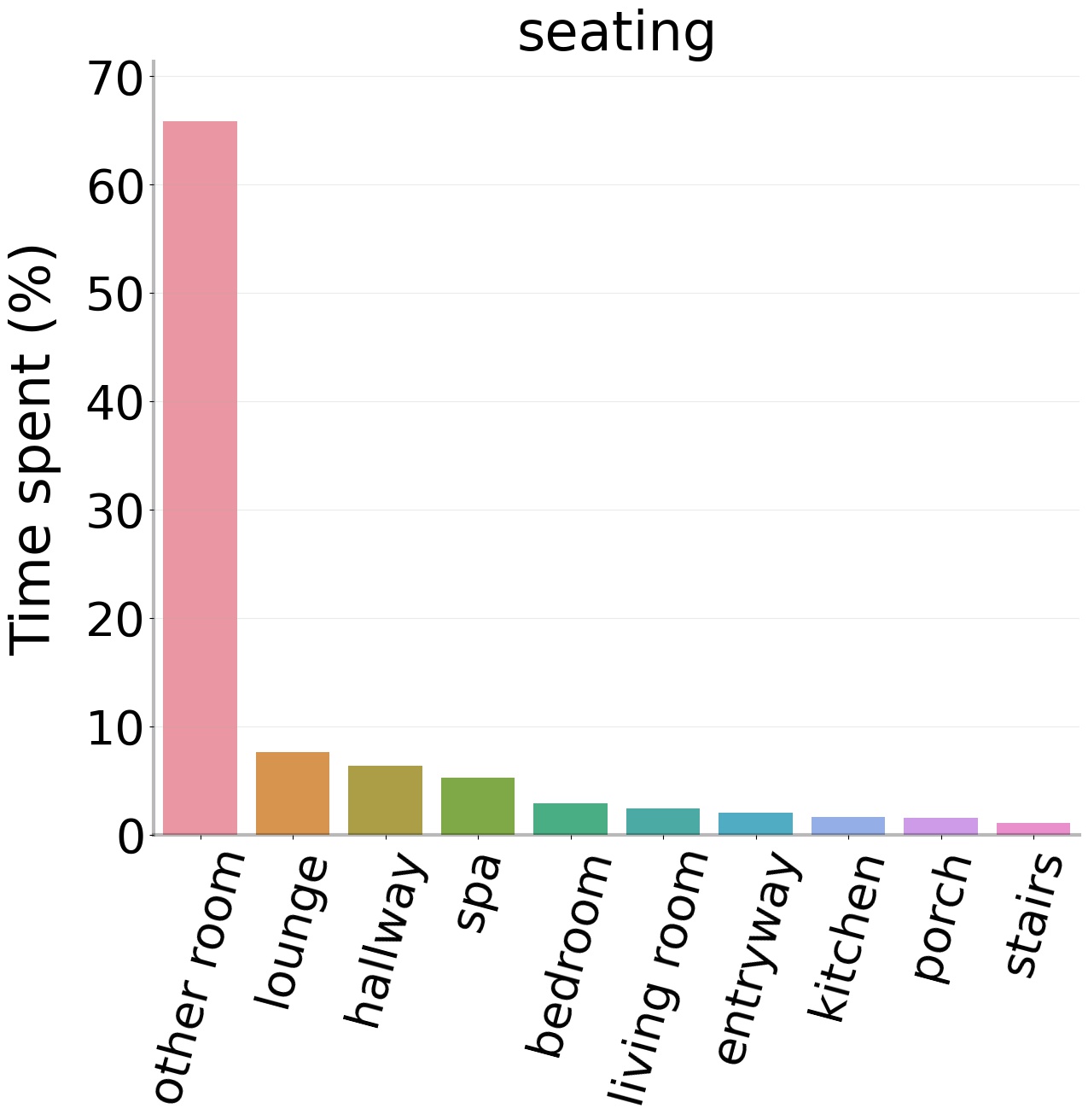}
            \caption{IL on $40k$ Human demos}
        \end{subfigure}
    \end{minipage}
    \hfill
    \begin{minipage}[a]{0.32\textwidth}
        \begin{subfigure}{\textwidth}
            \centering
            \includegraphics[width=\textwidth]{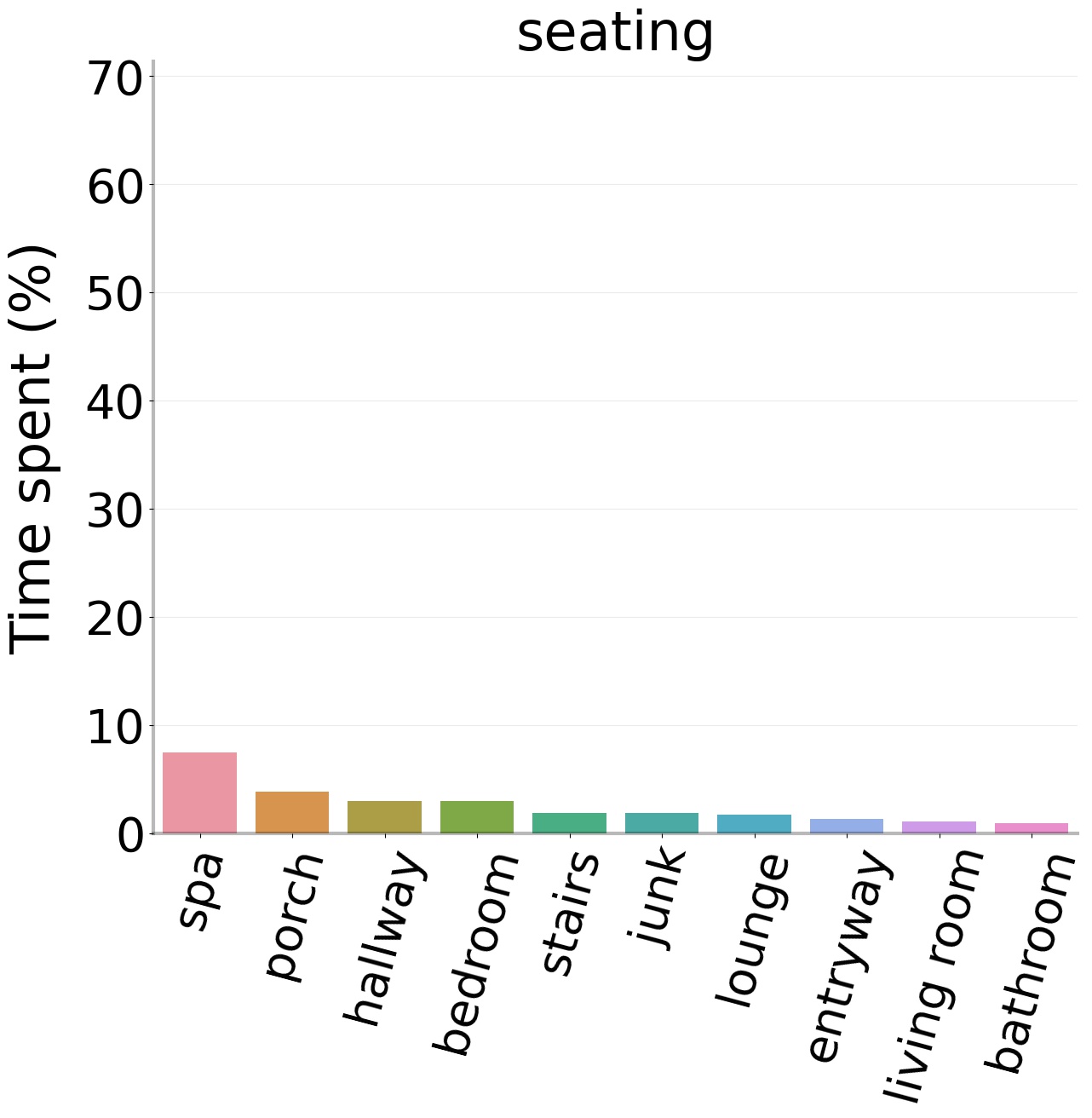}
            \caption{RL}
        \end{subfigure}
    \end{minipage}

    \caption{Comparison of per room time spent for all MP3D goal categories on \textsc{val} split for human demonstrations~\vs IL agents trained on human demos~\vs RL agents. The plot shows the top 10 rooms ordered by the maximum time spent in each room.}
    \label{fig:pRTS_per_object_4}
    
\end{figure*}

\begin{figure*}[t]\ContinuedFloat
    \centering
    \begin{minipage}[a]{0.32\textwidth}
        \begin{subfigure}{\textwidth}
            \centering
            \includegraphics[width=\textwidth]{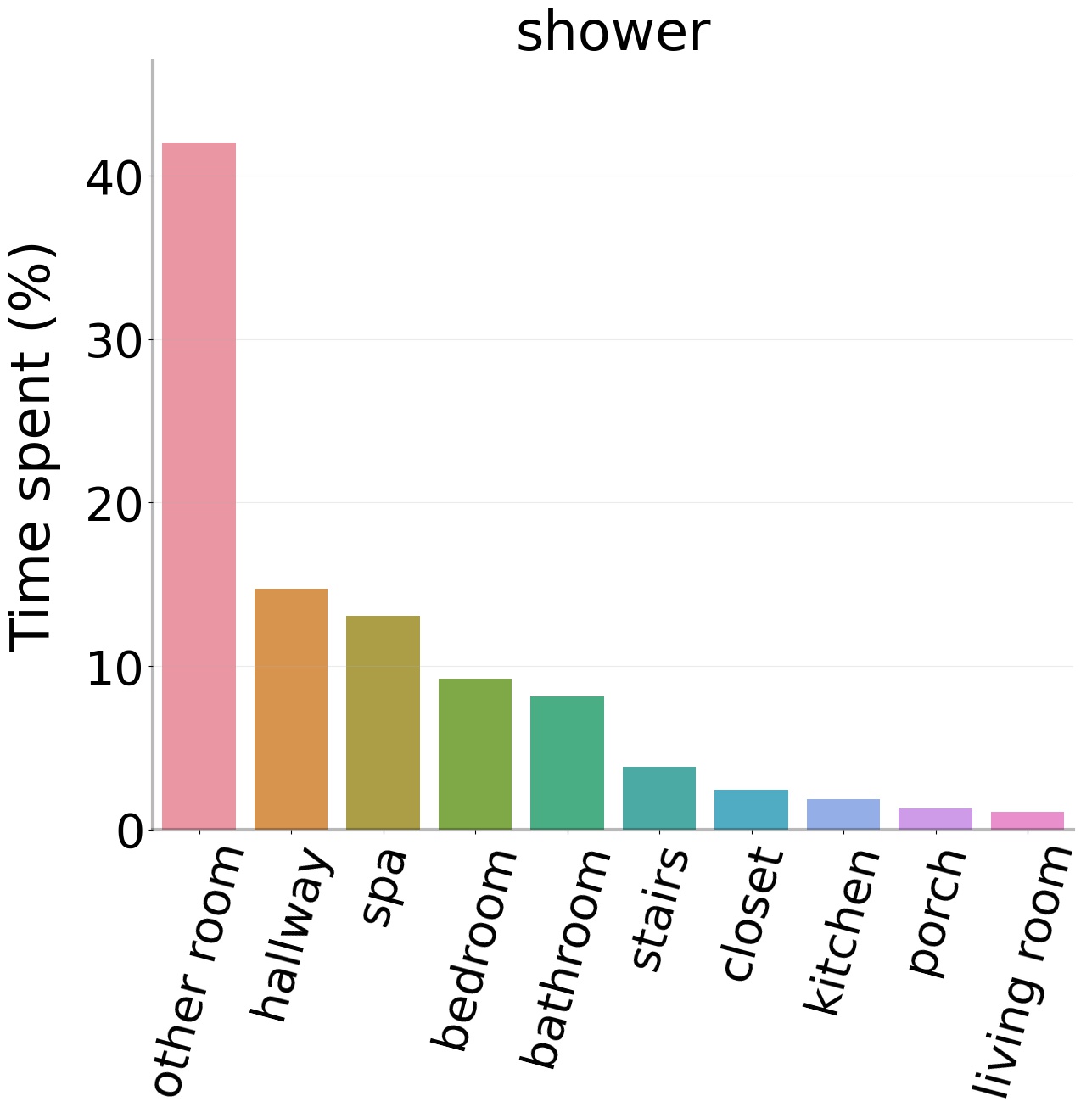}
            %\caption{Humans}
        \end{subfigure}
    \end{minipage}
    \hfill
    \begin{minipage}[a]{0.32\textwidth}
        \begin{subfigure}{\textwidth}
            \centering
            \includegraphics[width=\textwidth]{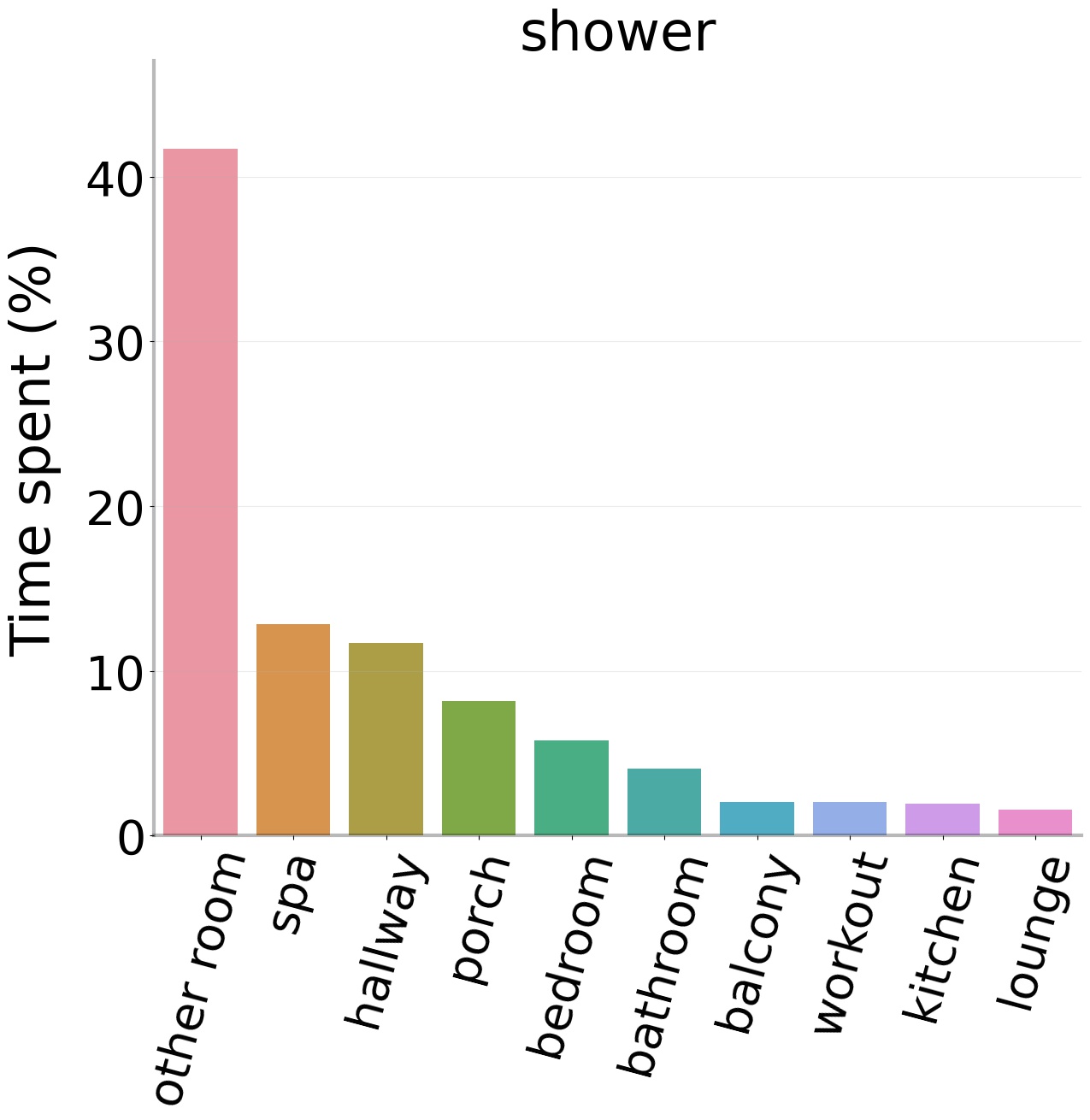}
            %\caption{IL on $40k$ Human demos}
        \end{subfigure}
    \end{minipage}
    \hfill
    \begin{minipage}[a]{0.32\textwidth}
        \begin{subfigure}{\textwidth}
            \centering
            \includegraphics[width=\textwidth]{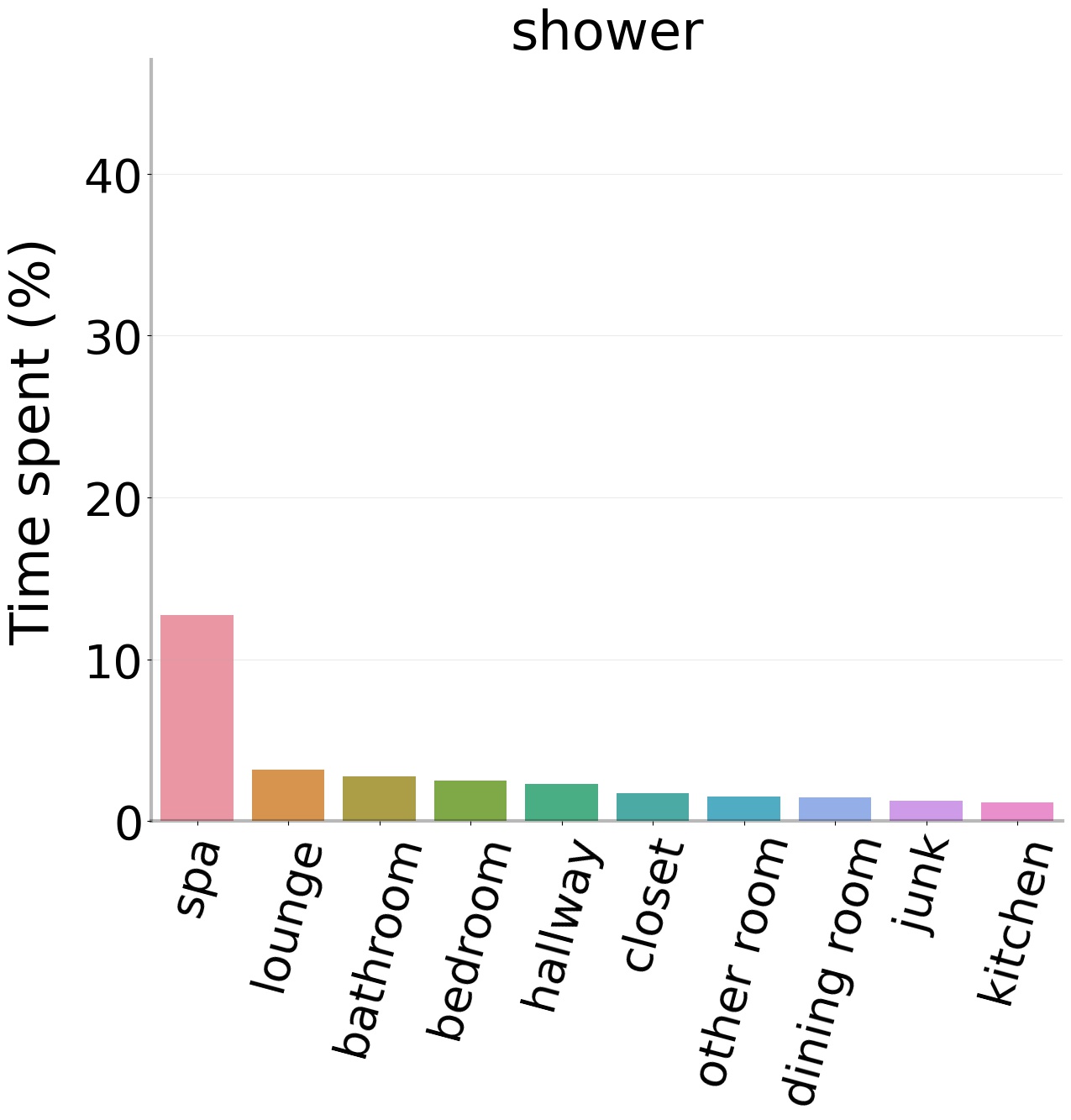}
            %\caption{RL}
        \end{subfigure}
    \end{minipage}

    \begin{minipage}[a]{0.32\textwidth}
        \begin{subfigure}{\textwidth}
            \centering
            \includegraphics[width=\textwidth]{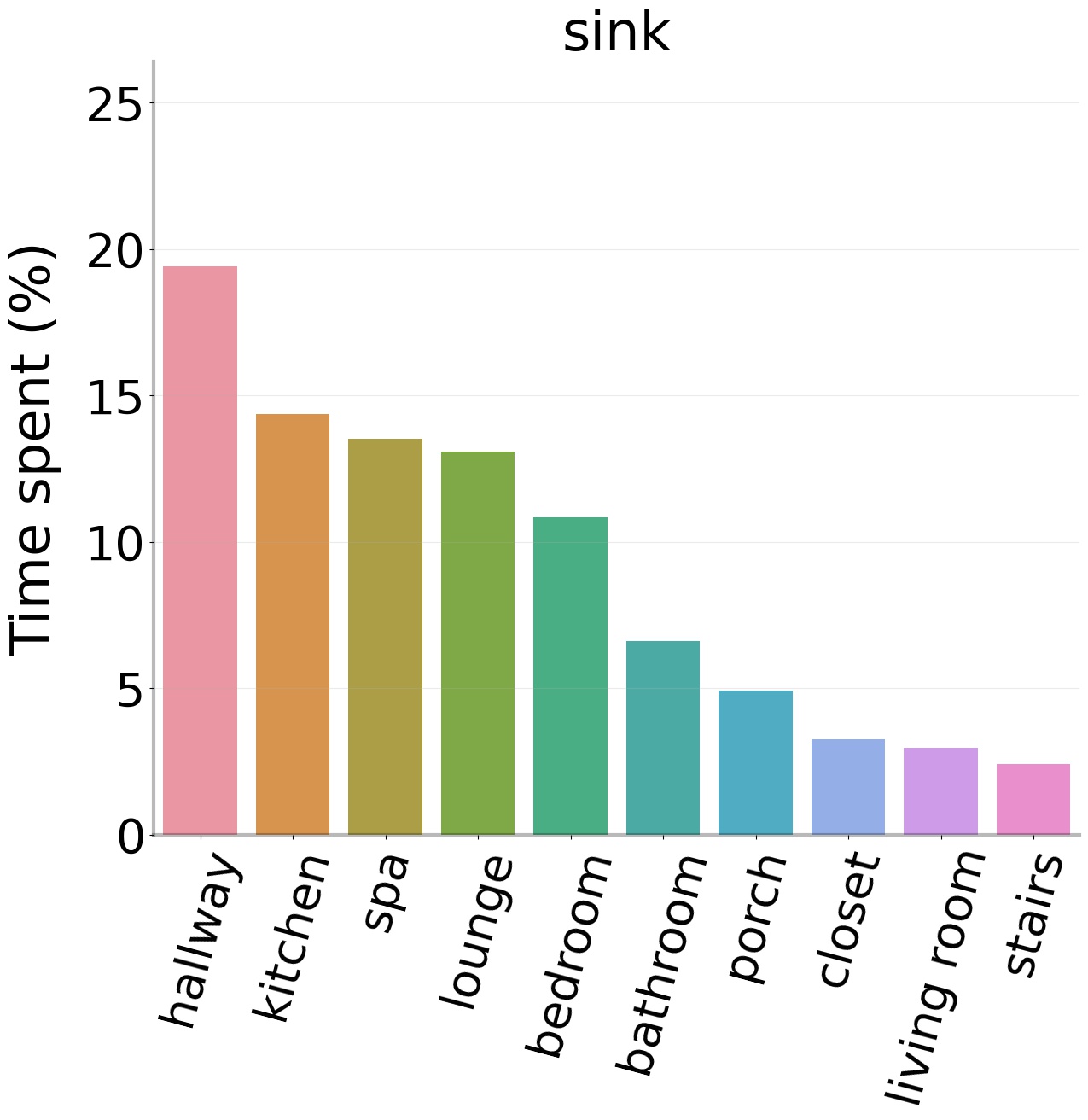}
            %\caption{Humans}
        \end{subfigure}
    \end{minipage}
    \hfill
    \begin{minipage}[a]{0.32\textwidth}
        \begin{subfigure}{\textwidth}
            \centering
            \includegraphics[width=\textwidth]{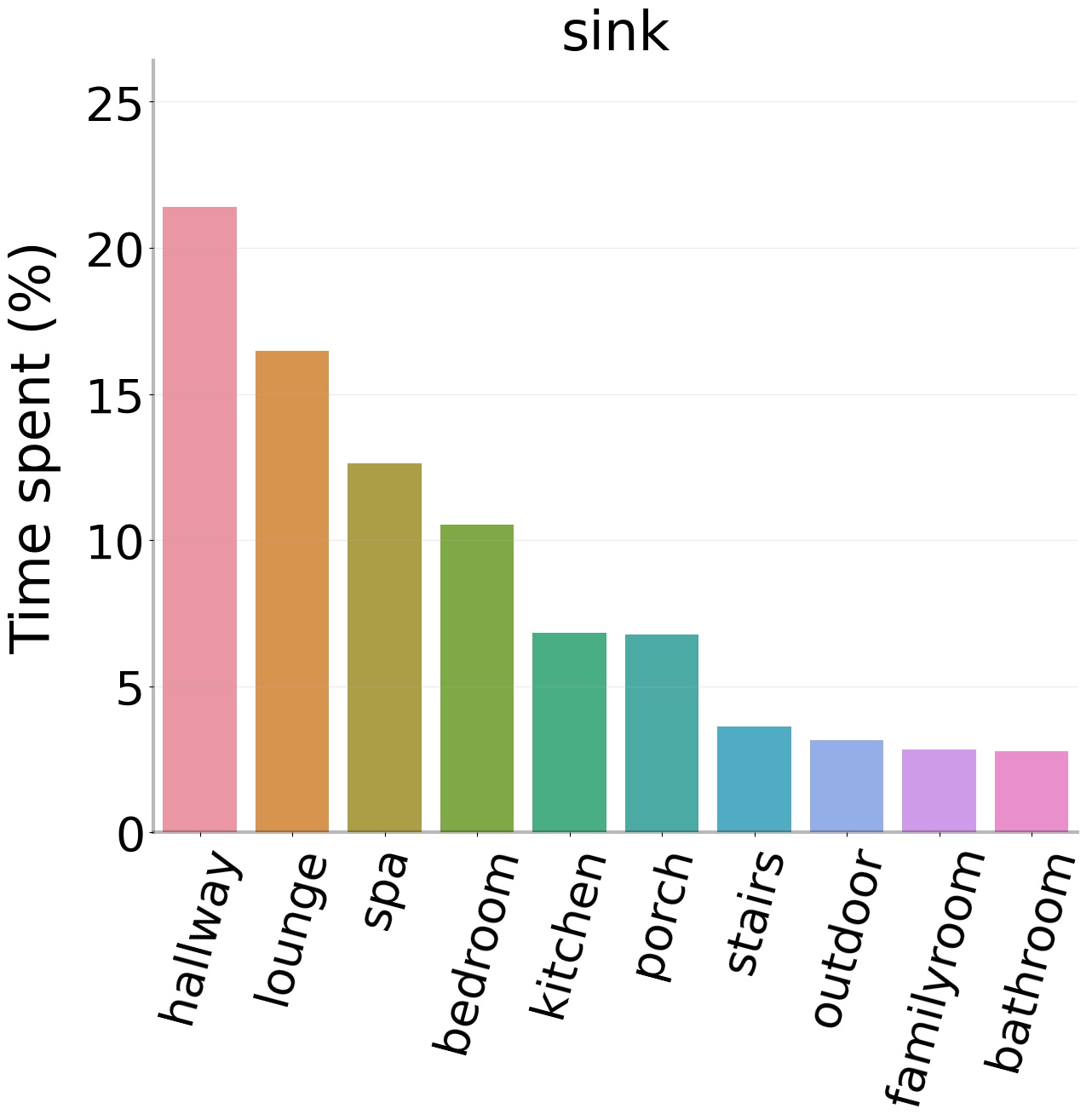}
            %\caption{IL on $40k$ Human demos}
        \end{subfigure}
    \end{minipage}
    \hfill
    \begin{minipage}[a]{0.32\textwidth}
        \begin{subfigure}{\textwidth}
            \centering
            \includegraphics[width=\textwidth]{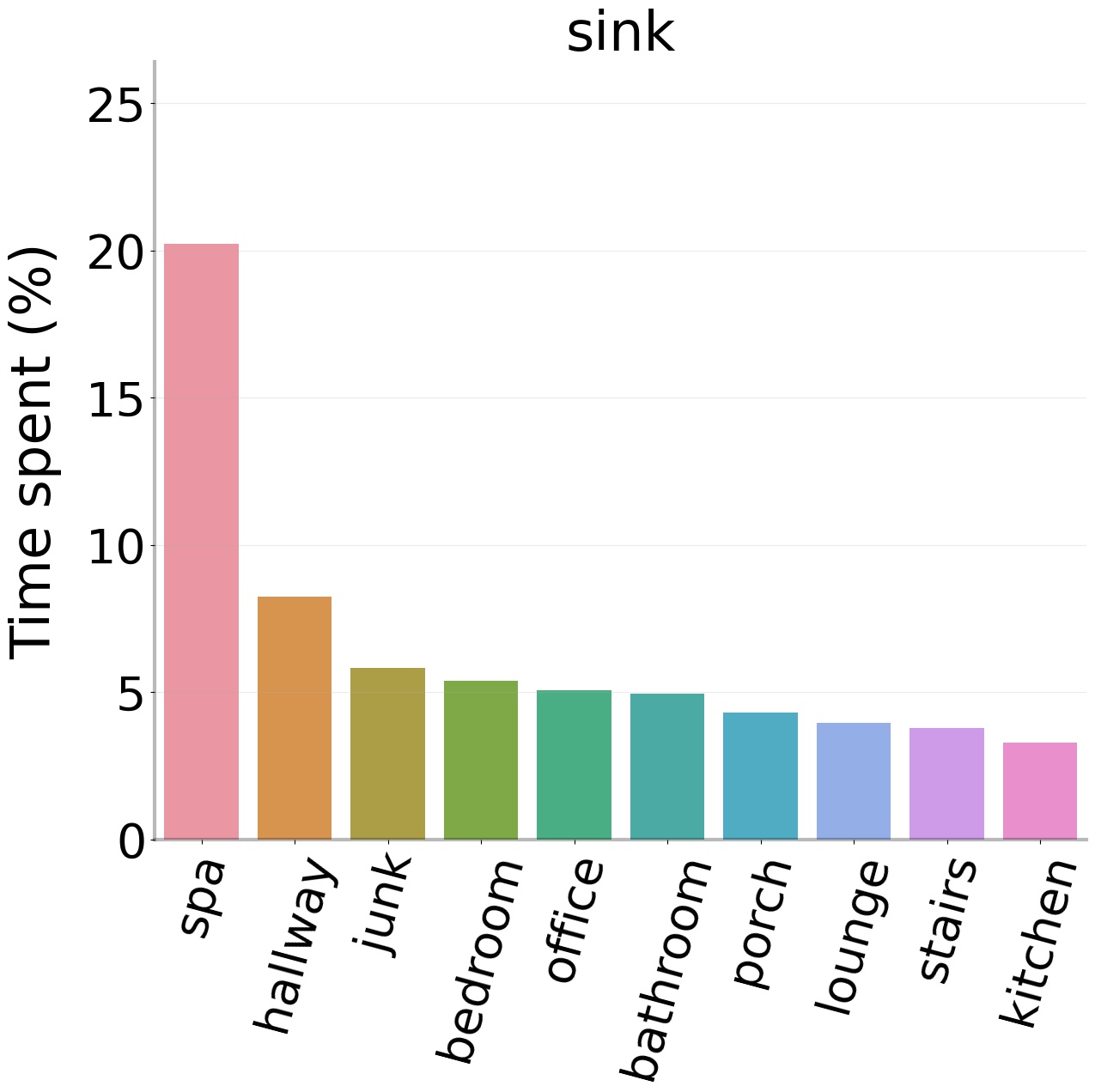}
            %\caption{RL}
        \end{subfigure}
    \end{minipage}

    \begin{minipage}[a]{0.32\textwidth}
        \begin{subfigure}{\textwidth}
            \centering
            \includegraphics[width=\textwidth]{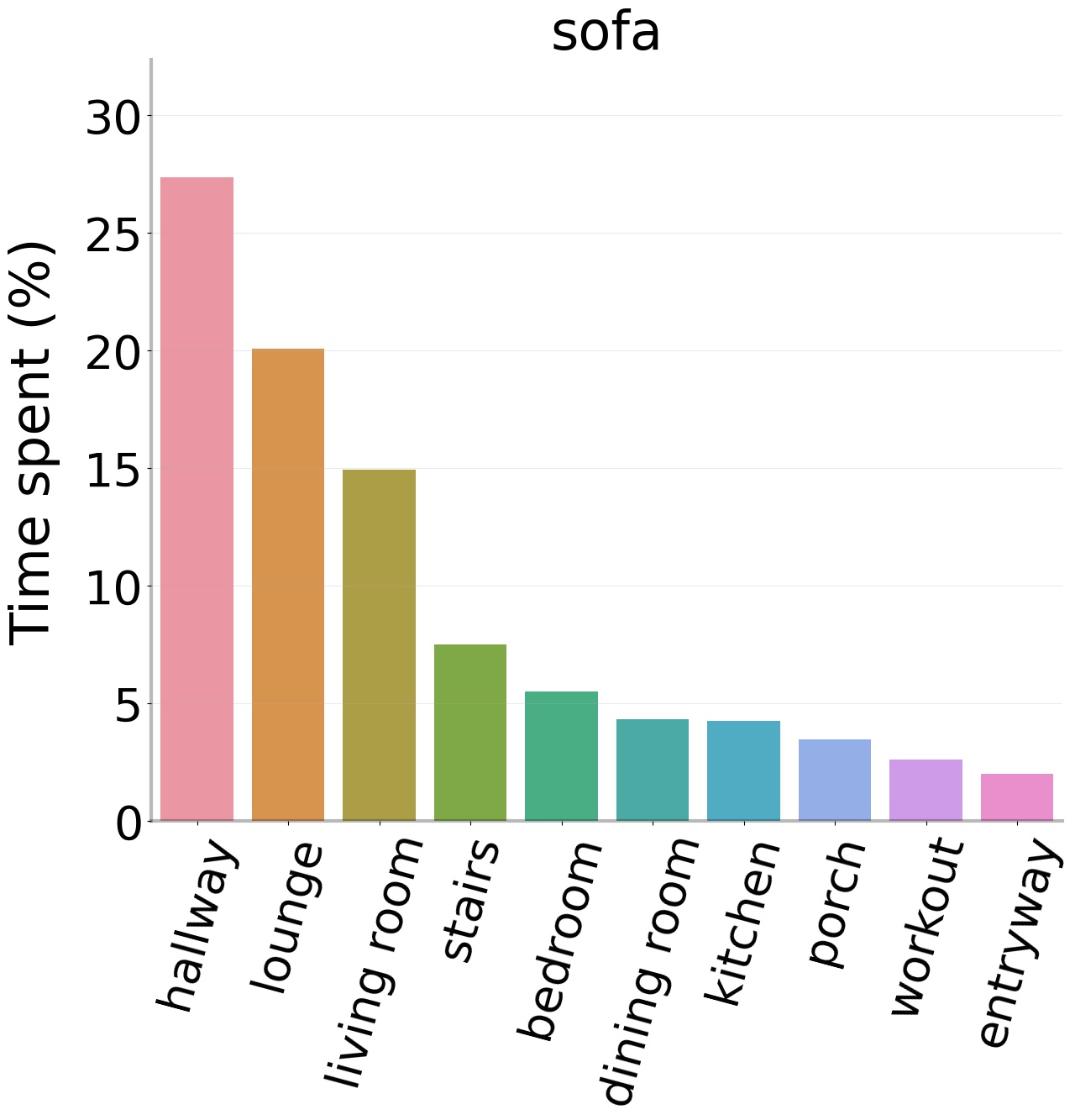}
            \caption{Humans}
        \end{subfigure}
    \end{minipage}
    \hfill
    \begin{minipage}[a]{0.32\textwidth}
        \begin{subfigure}{\textwidth}
            \centering
            \includegraphics[width=\textwidth]{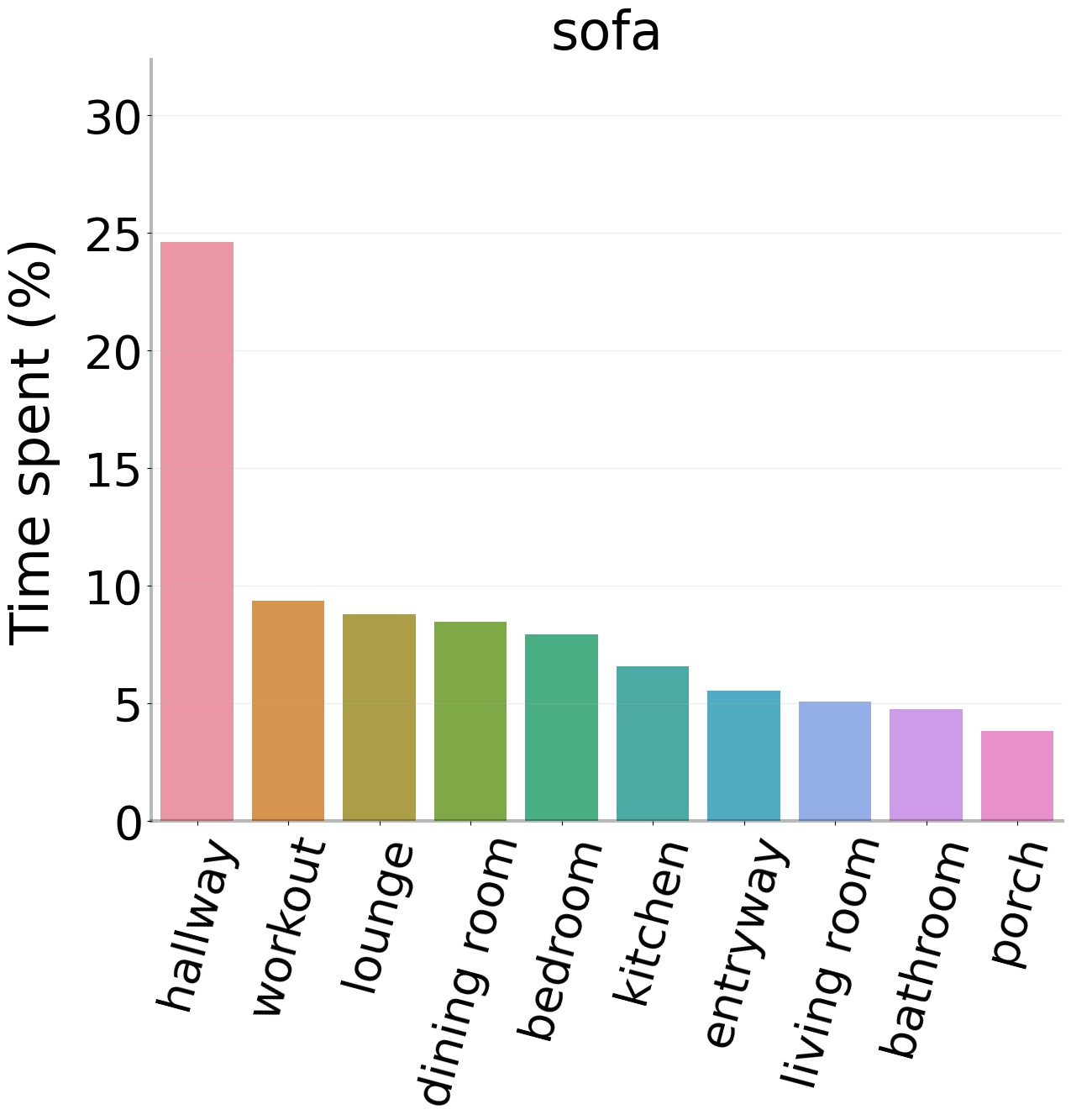}
            \caption{IL on $40k$ Human demos}
        \end{subfigure}
    \end{minipage}
    \hfill
    \begin{minipage}[a]{0.32\textwidth}
        \begin{subfigure}{\textwidth}
            \centering
            \includegraphics[width=\textwidth]{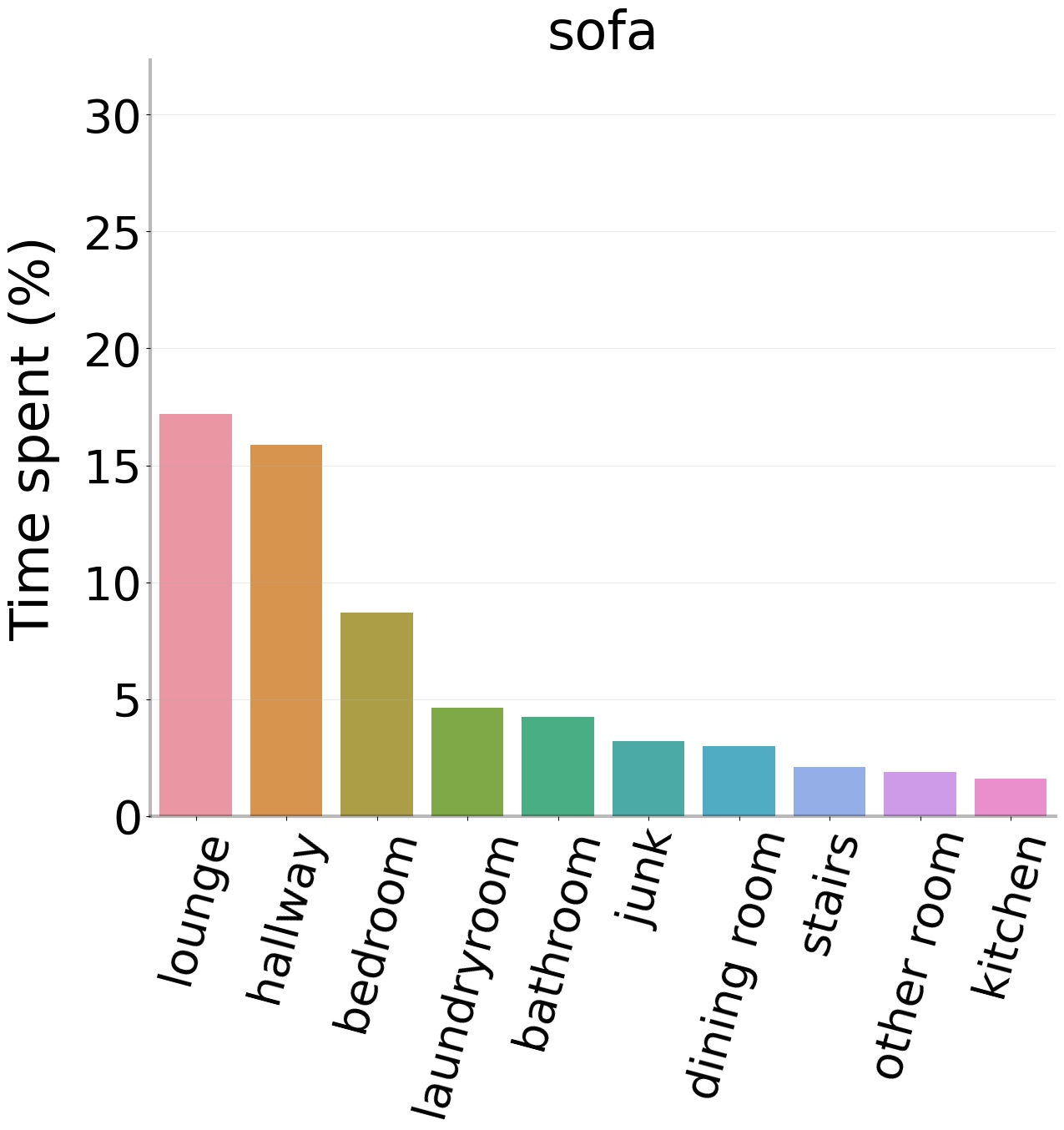}
            \caption{RL}
        \end{subfigure}
    \end{minipage}
    \caption{Comparison of per room time spent for all MP3D goal categories on \textsc{val} split for human demonstrations~\vs IL agents trained on human demos~\vs RL agents. The plot shows the top 10 rooms ordered by the maximum time spent in each room.}
    \label{fig:pRTS_per_object_5}
    
\end{figure*}

\begin{figure*}[t]\ContinuedFloat
    \centering
    \begin{minipage}[a]{0.32\textwidth}
        \begin{subfigure}{\textwidth}
            \centering
            \includegraphics[width=\textwidth]{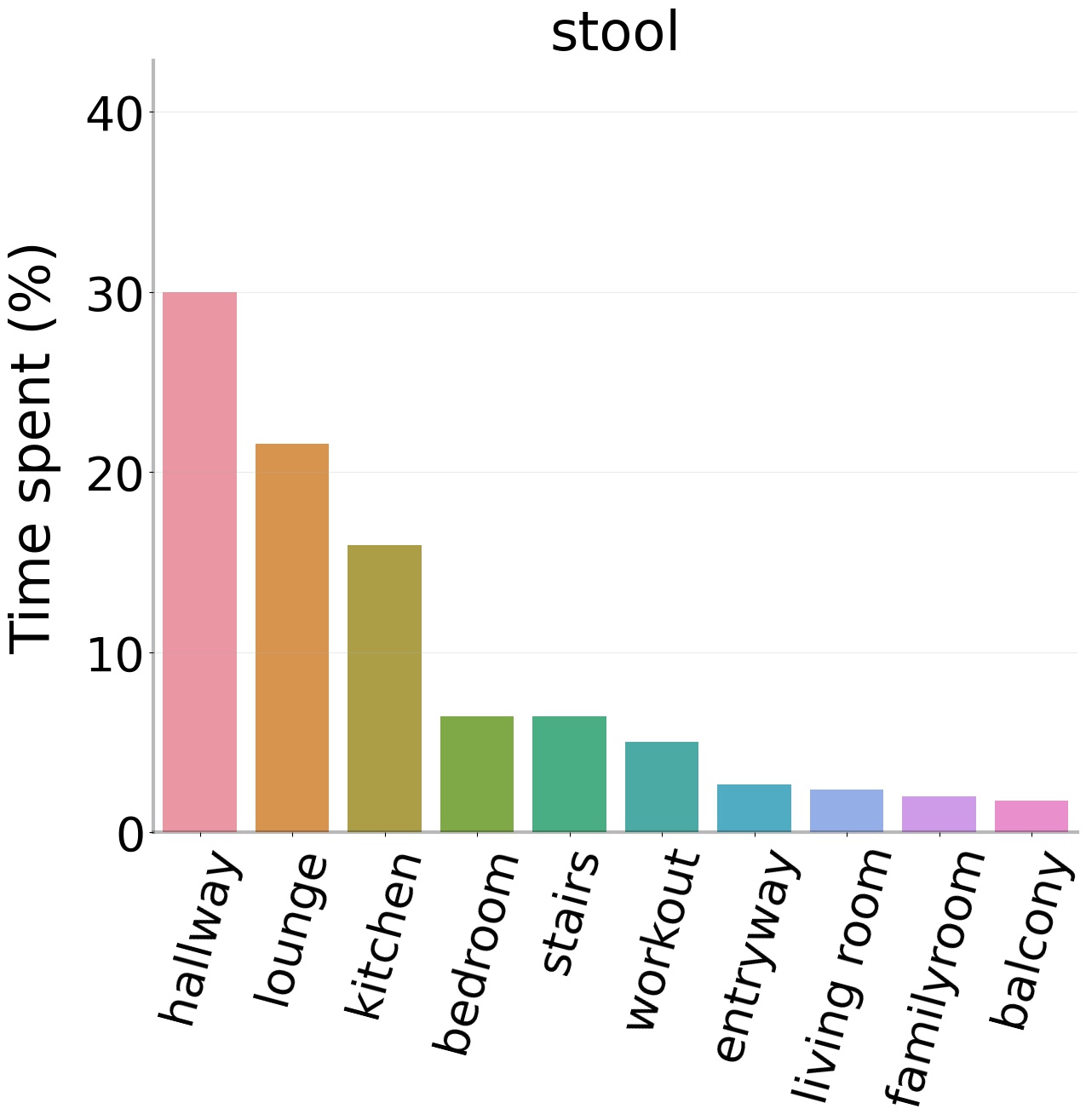}
            %\caption{Humans}
        \end{subfigure}
    \end{minipage}
    \hfill
    \begin{minipage}[a]{0.32\textwidth}
        \begin{subfigure}{\textwidth}
            \centering
            \includegraphics[width=\textwidth]{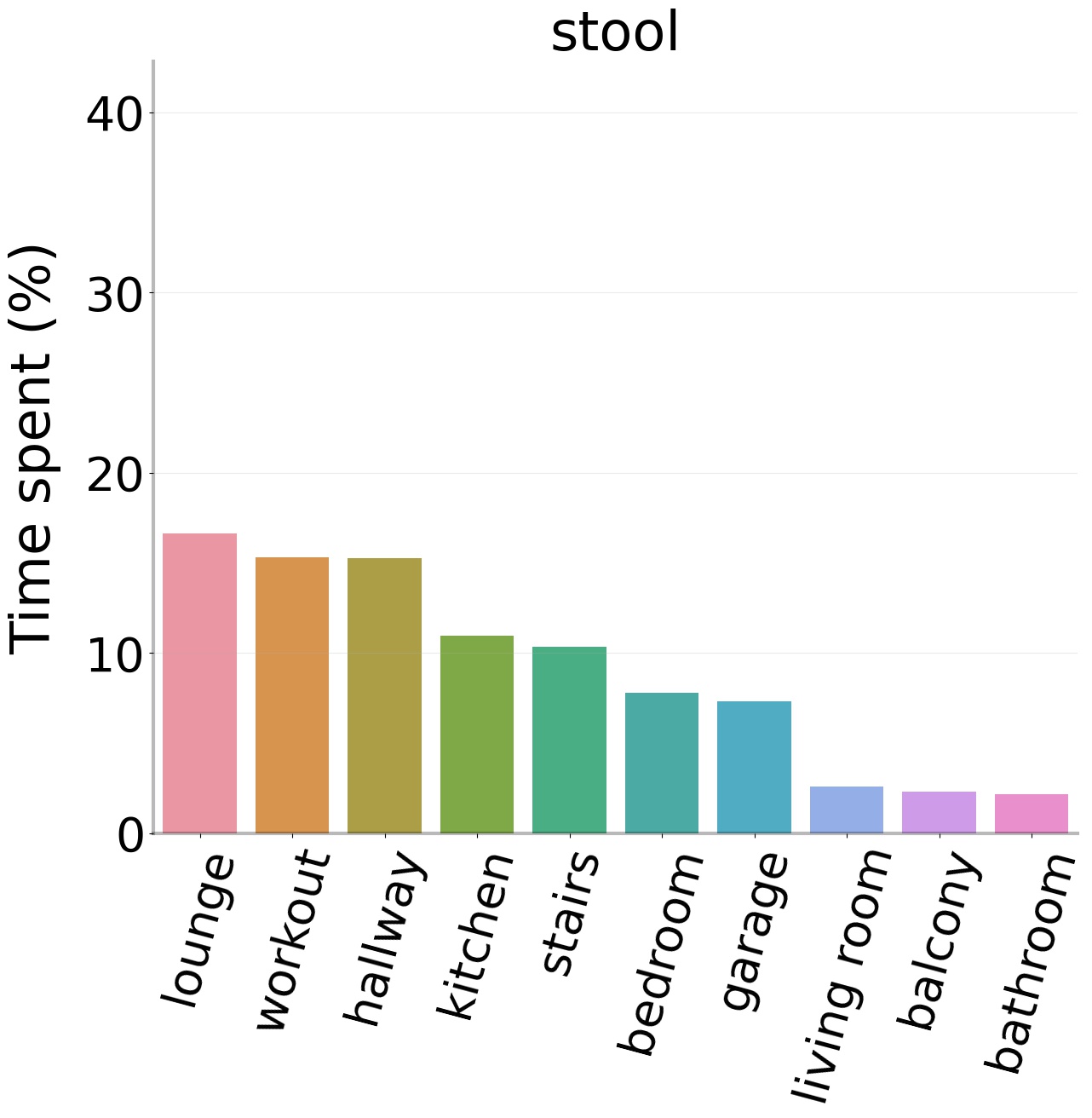}
            %\caption{IL on $40k$ Human demos}
        \end{subfigure}
    \end{minipage}
    \hfill
    \begin{minipage}[a]{0.32\textwidth}
        \begin{subfigure}{\textwidth}
            \centering
            \includegraphics[width=\textwidth]{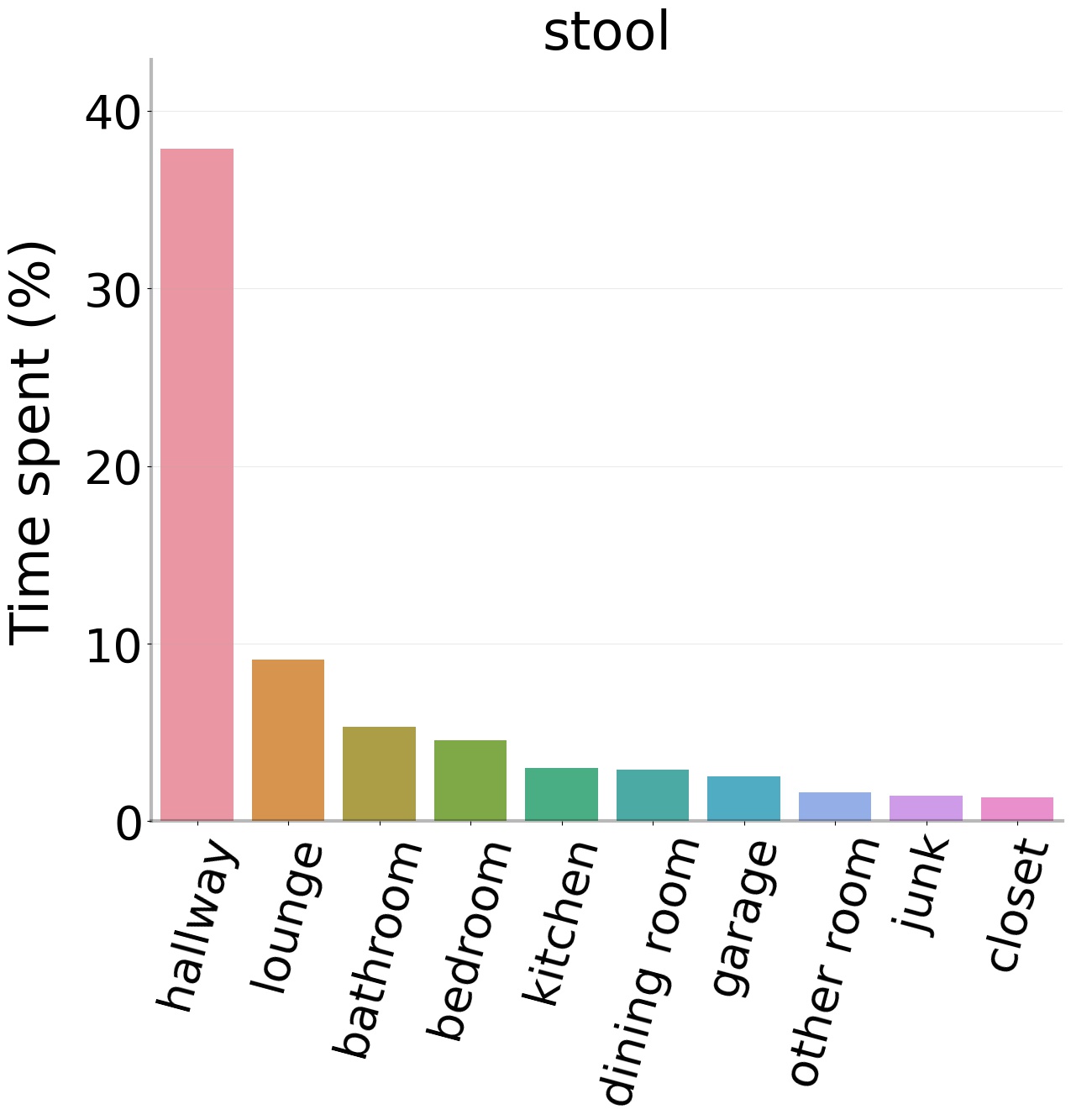}
            %\caption{RL}
        \end{subfigure}
    \end{minipage}

    \begin{minipage}[a]{0.32\textwidth}
        \begin{subfigure}{\textwidth}
            \centering
            \includegraphics[width=\textwidth]{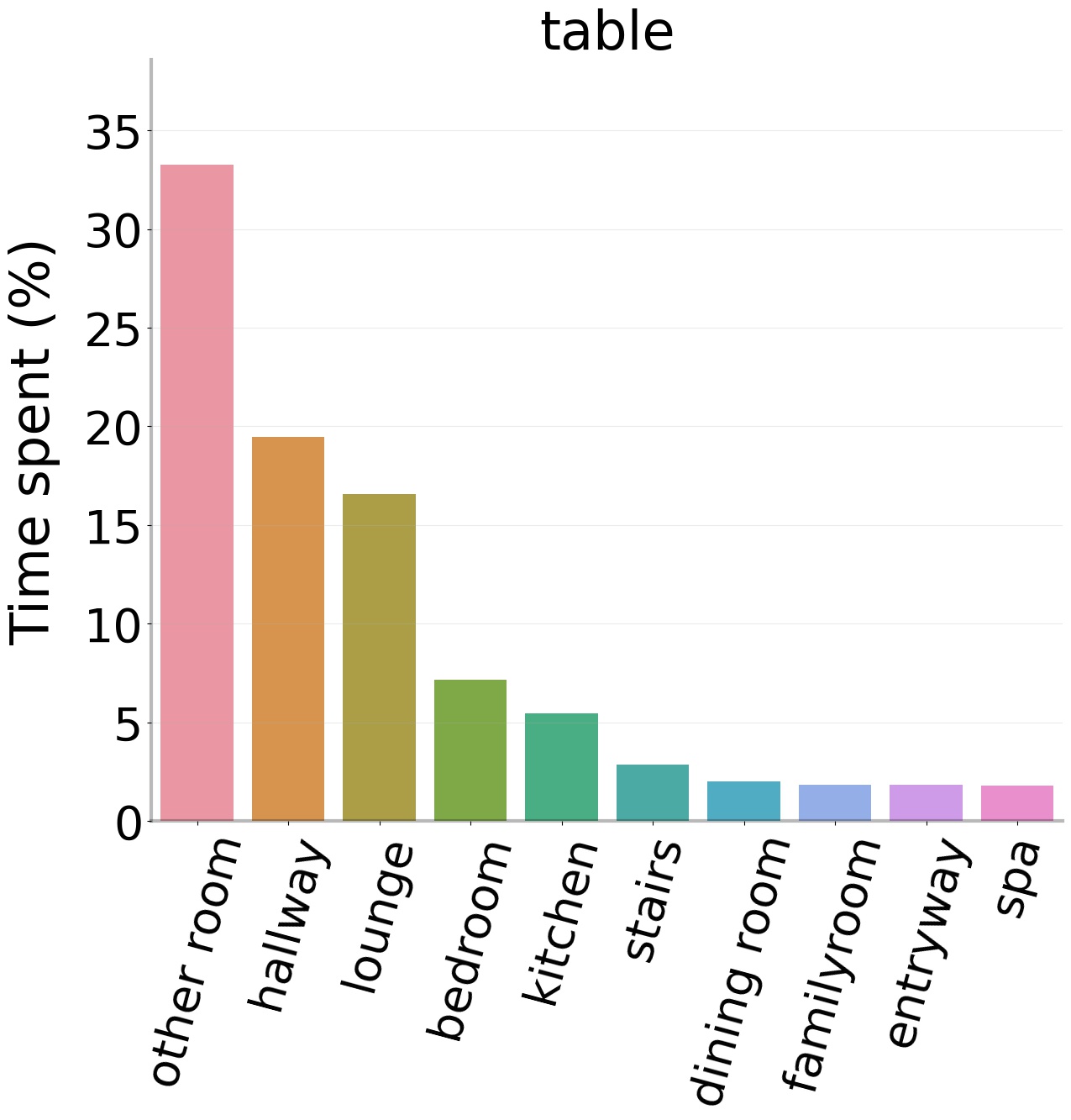}
            %\caption{Humans}
        \end{subfigure}
    \end{minipage}
    \hfill
    \begin{minipage}[a]{0.32\textwidth}
        \begin{subfigure}{\textwidth}
            \centering
            \includegraphics[width=\textwidth]{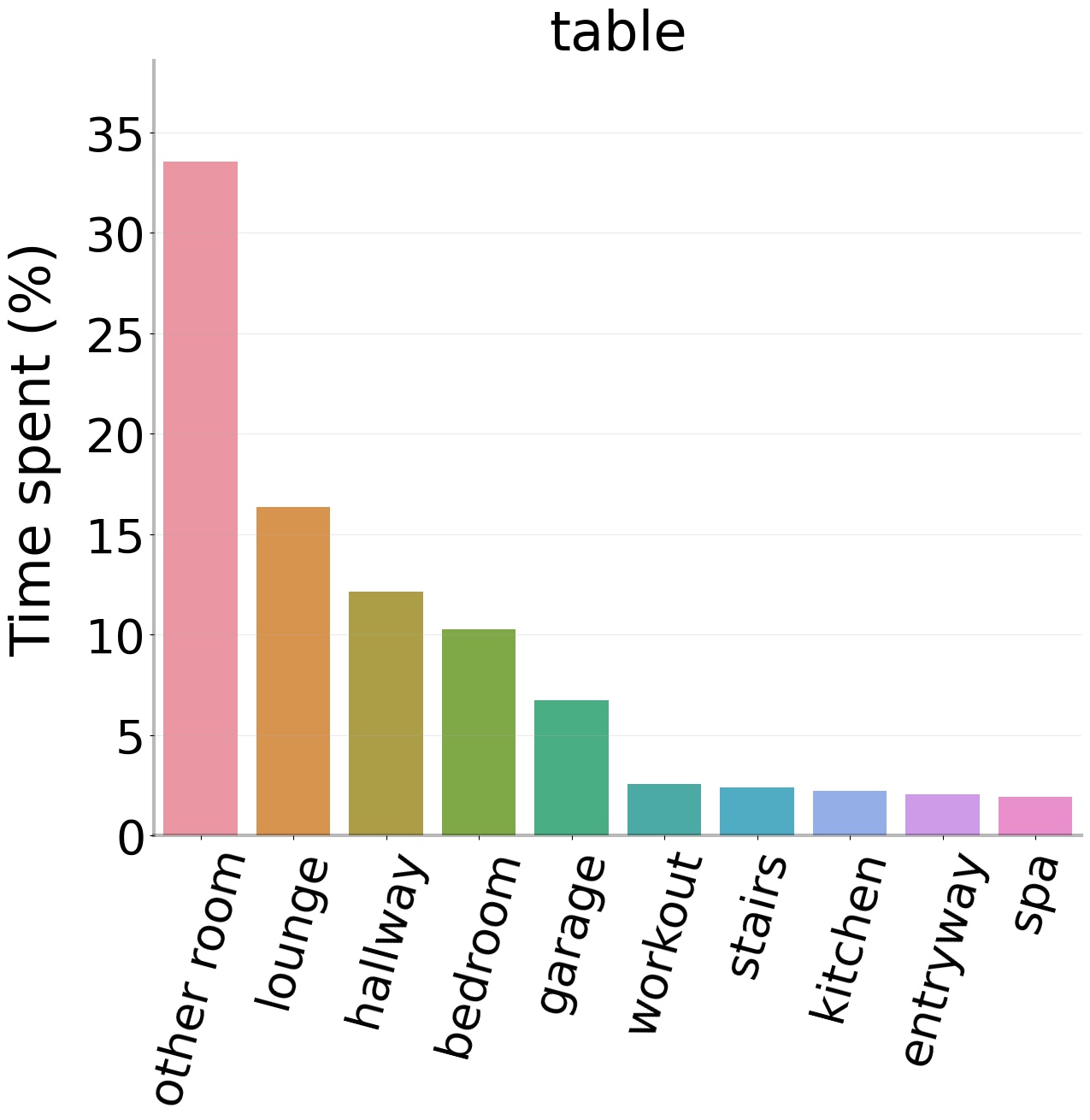}
            %\caption{IL on $40k$ Human demos}
        \end{subfigure}
    \end{minipage}
    \hfill
    \begin{minipage}[a]{0.32\textwidth}
        \begin{subfigure}{\textwidth}
            \centering
            \includegraphics[width=\textwidth]{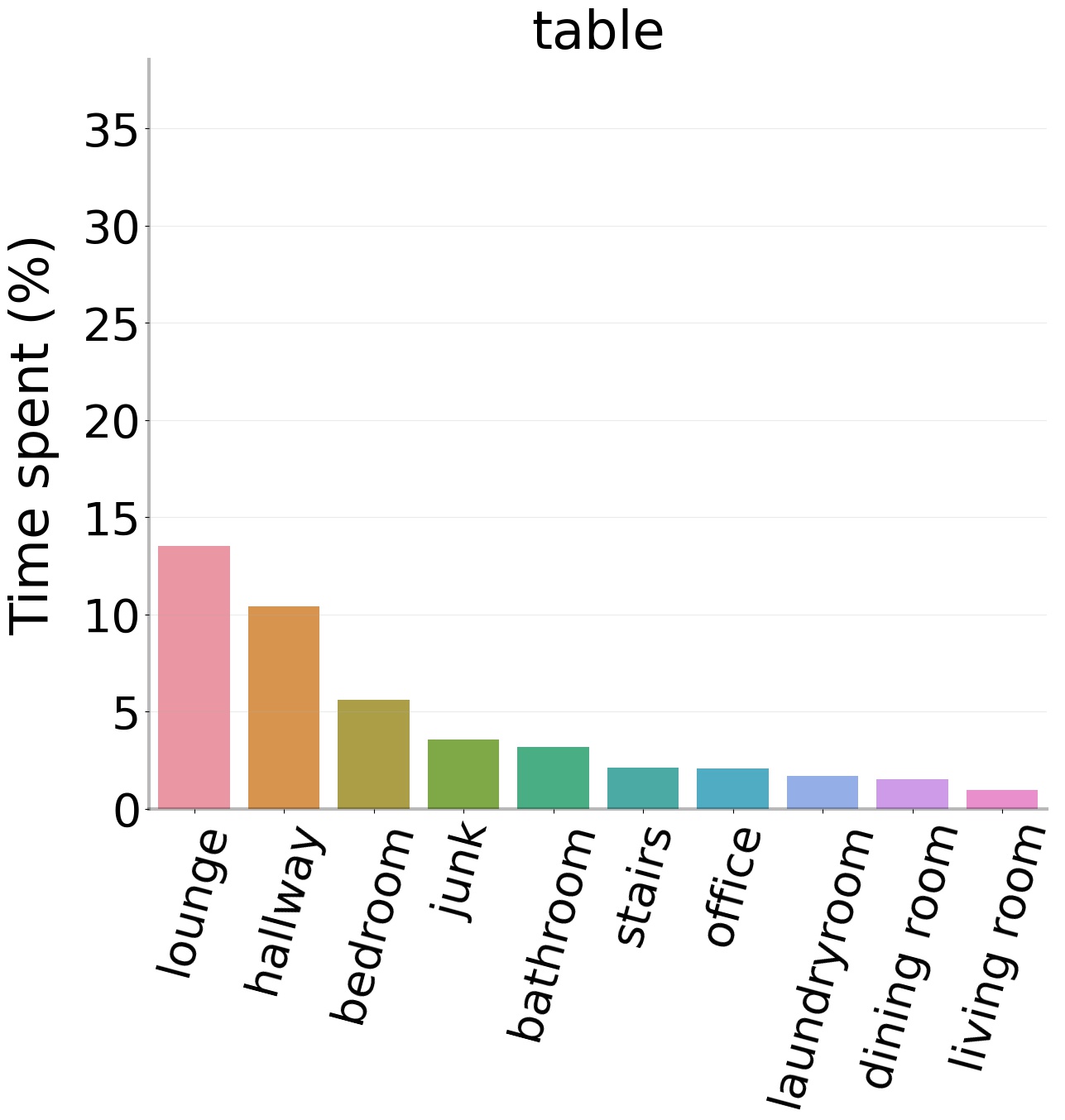}
            %\caption{RL}
        \end{subfigure}
    \end{minipage}

    \begin{minipage}[a]{0.32\textwidth}
        \begin{subfigure}{\textwidth}
            \centering
            \includegraphics[width=\textwidth]{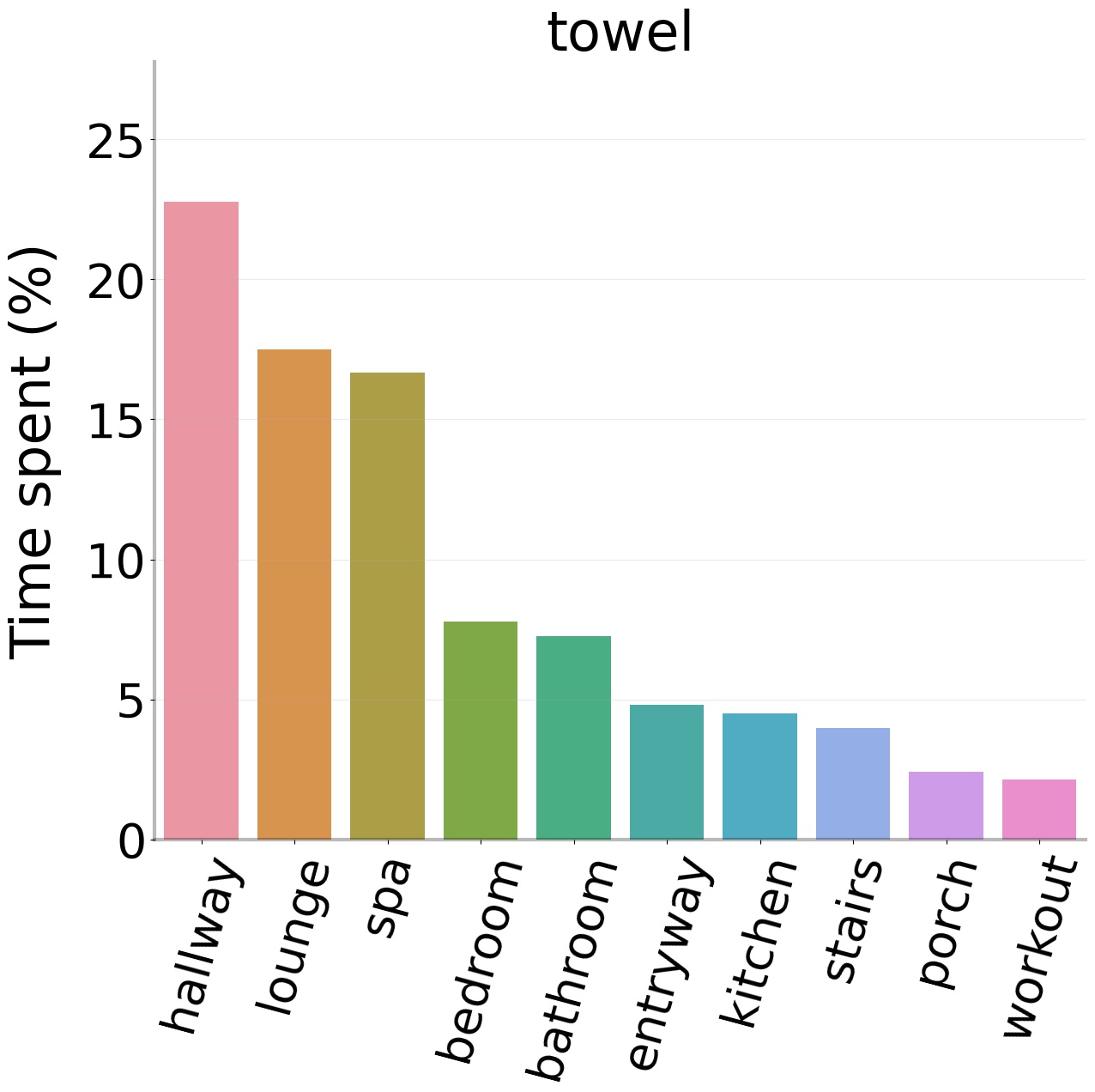}
            \caption{Humans}
        \end{subfigure}
    \end{minipage}
    \hfill
    \begin{minipage}[a]{0.32\textwidth}
        \begin{subfigure}{\textwidth}
            \centering
            \includegraphics[width=\textwidth]{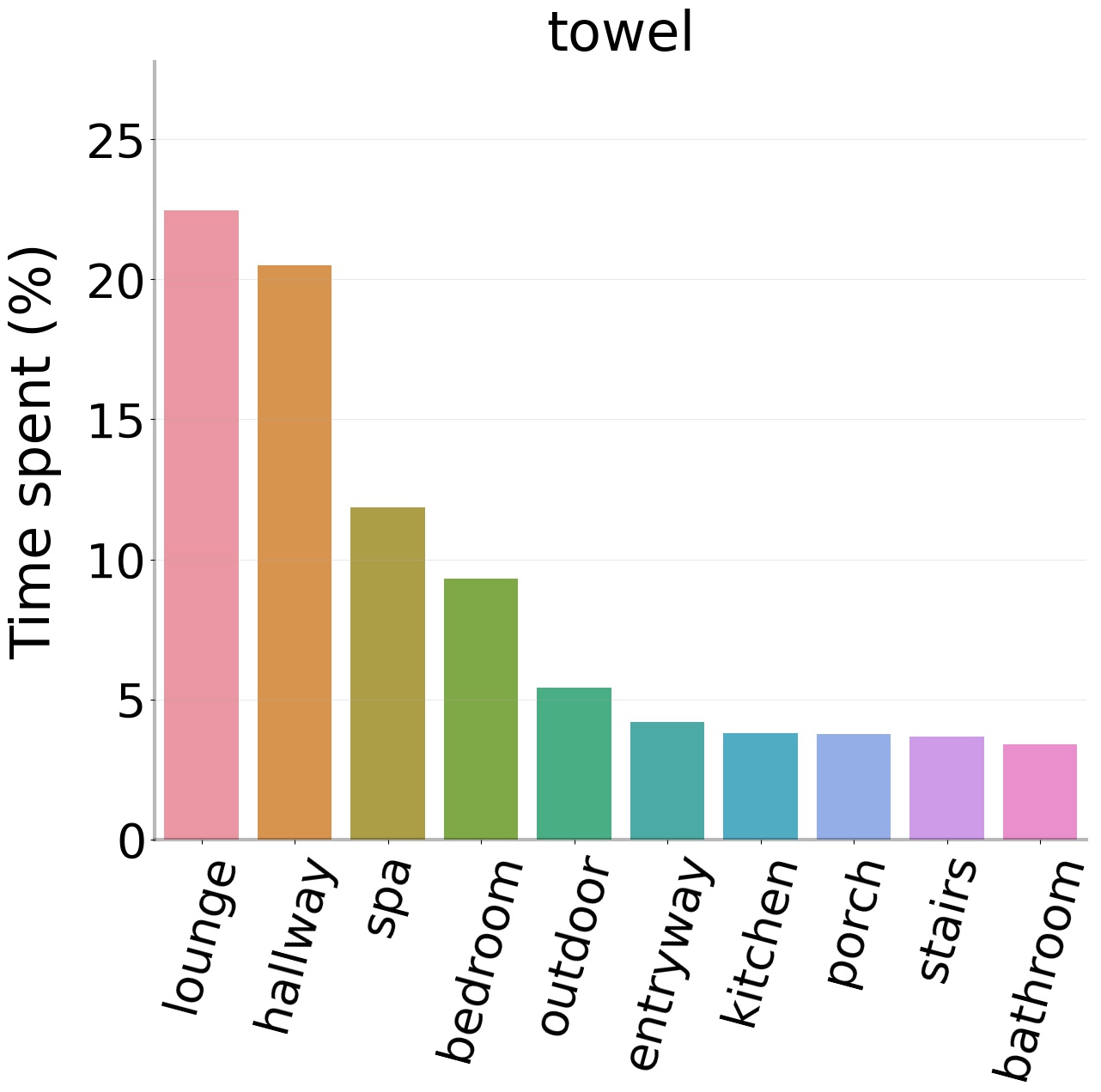}
            \caption{IL on $40k$ Human demos}
        \end{subfigure}
    \end{minipage}
    \hfill
    \begin{minipage}[a]{0.32\textwidth}
        \begin{subfigure}{\textwidth}
            \centering
            \includegraphics[width=\textwidth]{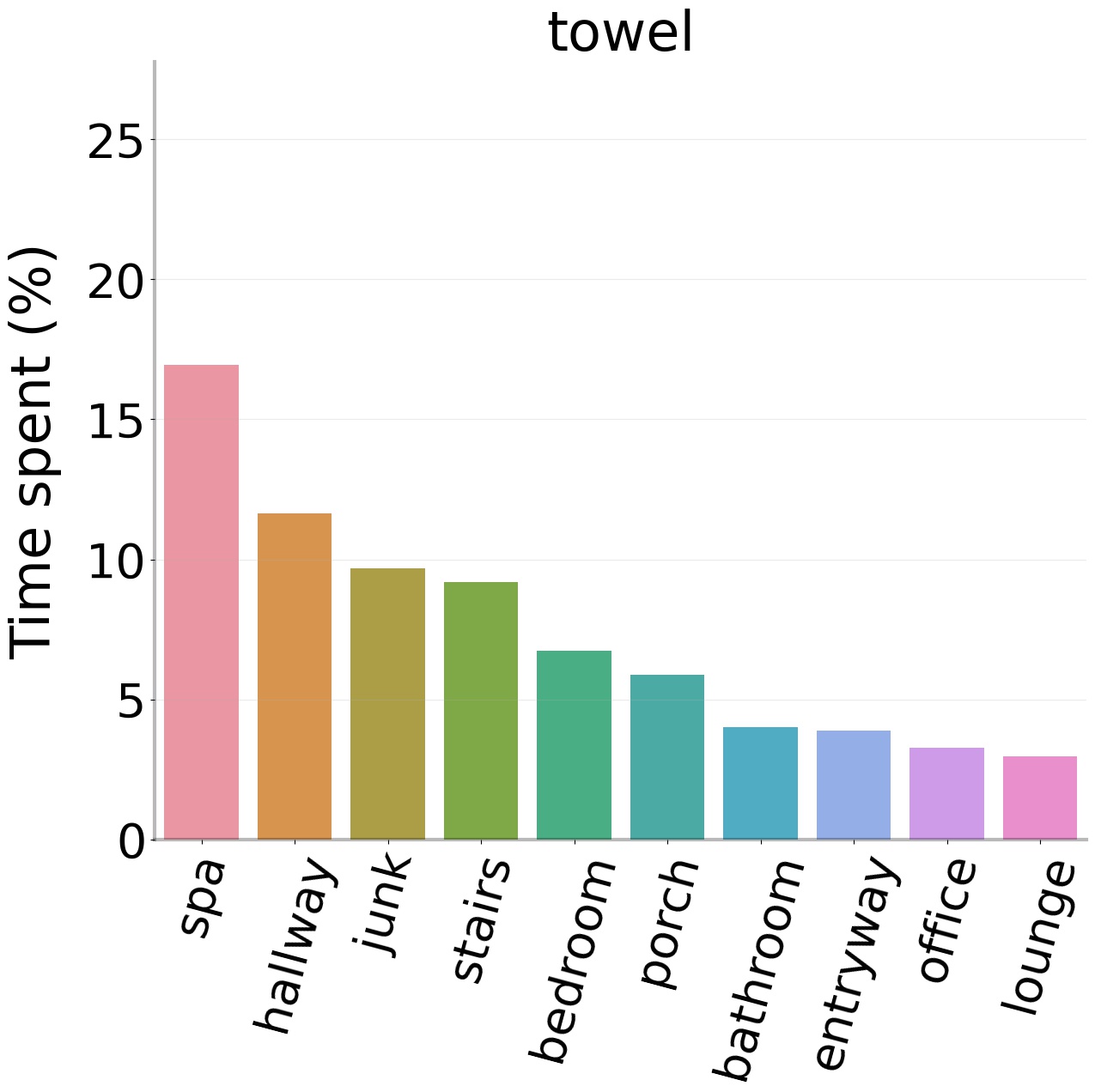}
            \caption{RL}
        \end{subfigure}
    \end{minipage}

    \caption{Comparison of per room time spent for all MP3D goal categories on \textsc{val} split for human demonstrations~\vs IL agents trained on human demos~\vs RL agents. The plot shows the top 10 rooms ordered by the maximum time spent in each room.}
    \label{fig:pRTS_per_object_6}
    
\end{figure*}

\begin{figure*}[t]\ContinuedFloat
    \centering
    \begin{minipage}[a]{0.32\textwidth}
        \begin{subfigure}{\textwidth}
            \centering
            \includegraphics[width=\textwidth]{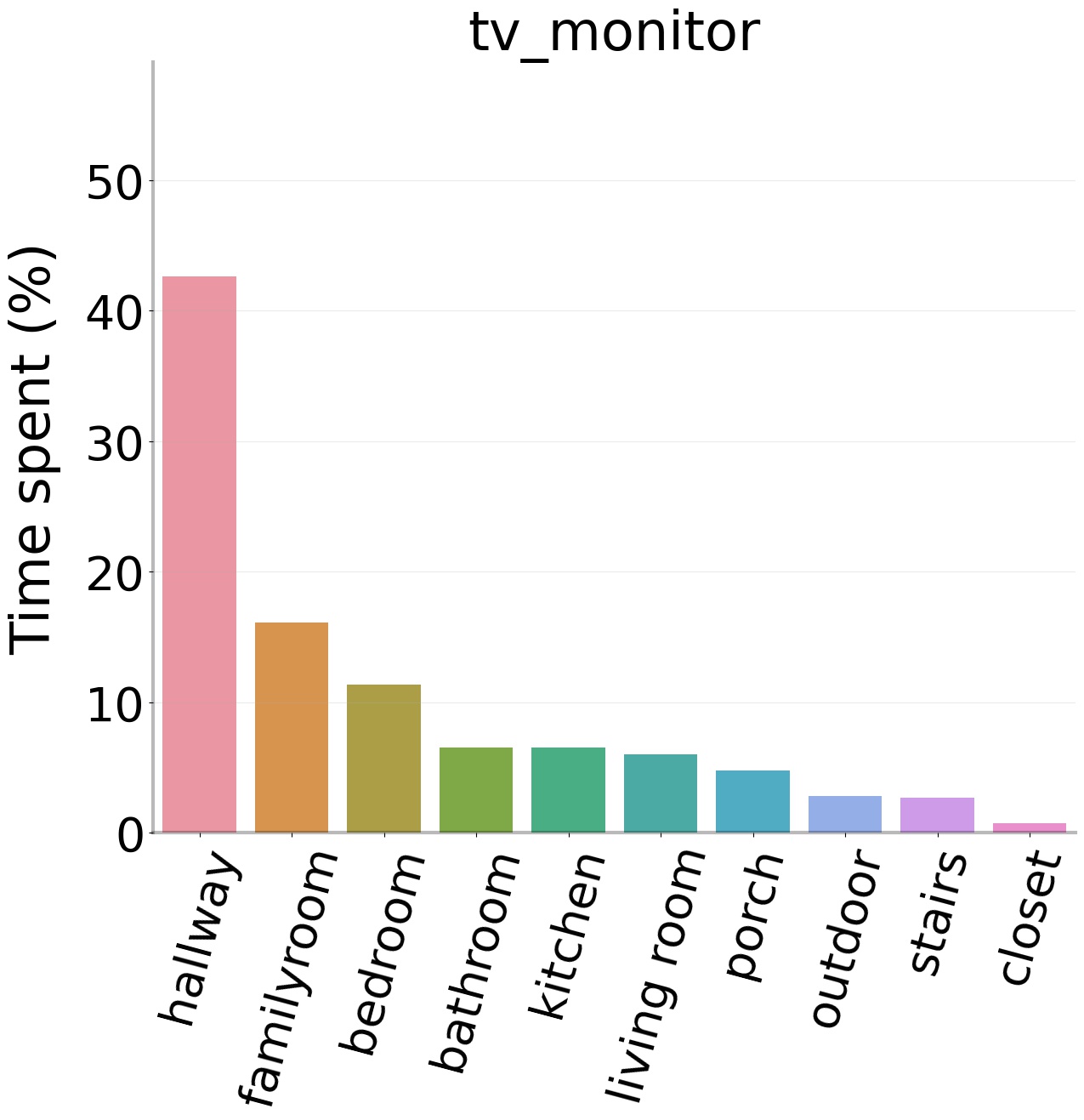}
            \caption{Humans}
        \end{subfigure}
    \end{minipage}
    \hfill
    \begin{minipage}[a]{0.32\textwidth}
        \begin{subfigure}{\textwidth}
            \centering
            \includegraphics[width=\textwidth]{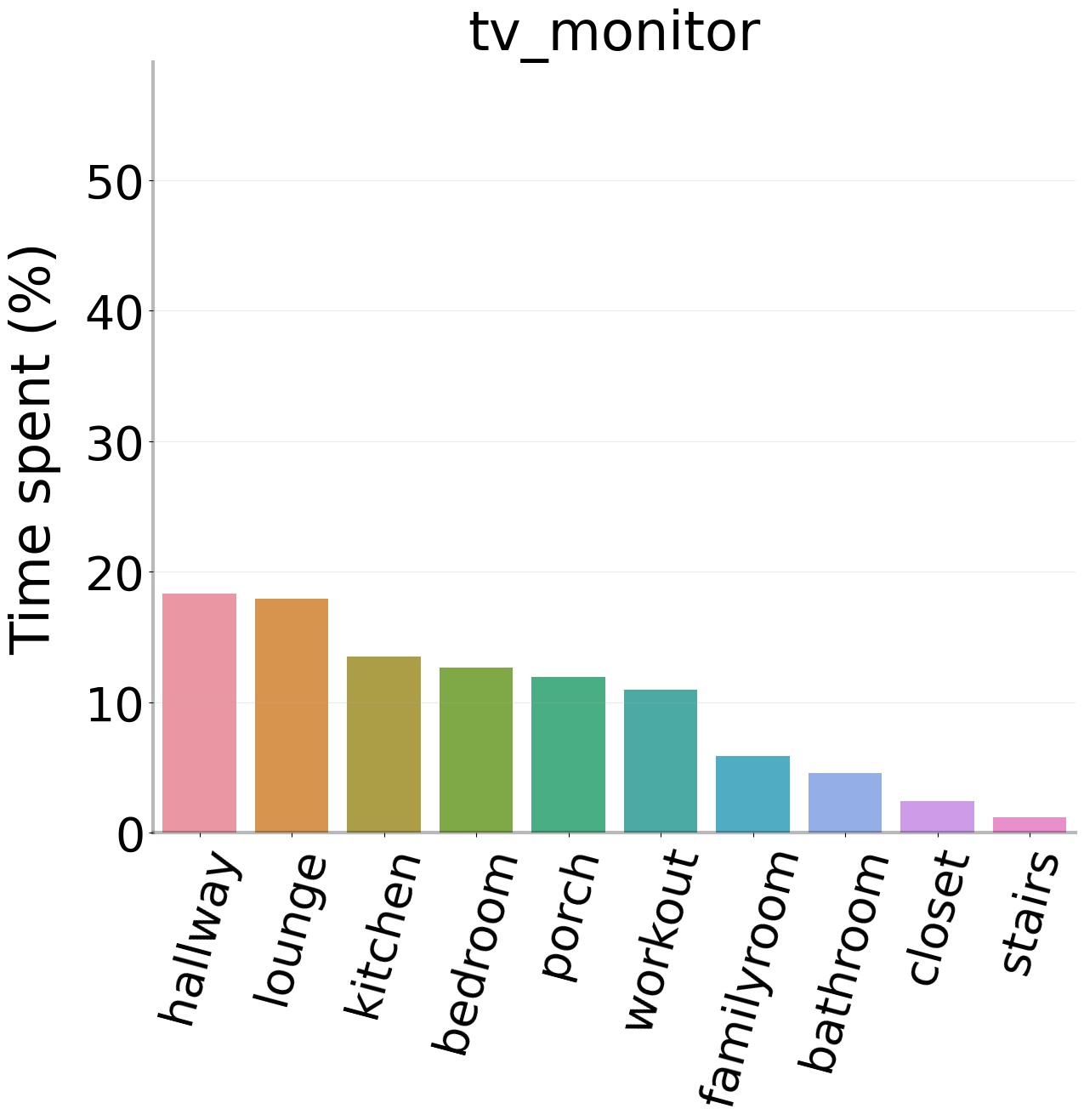}
            \caption{IL on $40k$ Human demos}
        \end{subfigure}
    \end{minipage}
    \hfill
    \begin{minipage}[a]{0.32\textwidth}
        \begin{subfigure}{\textwidth}
            \centering
            \includegraphics[width=\textwidth]{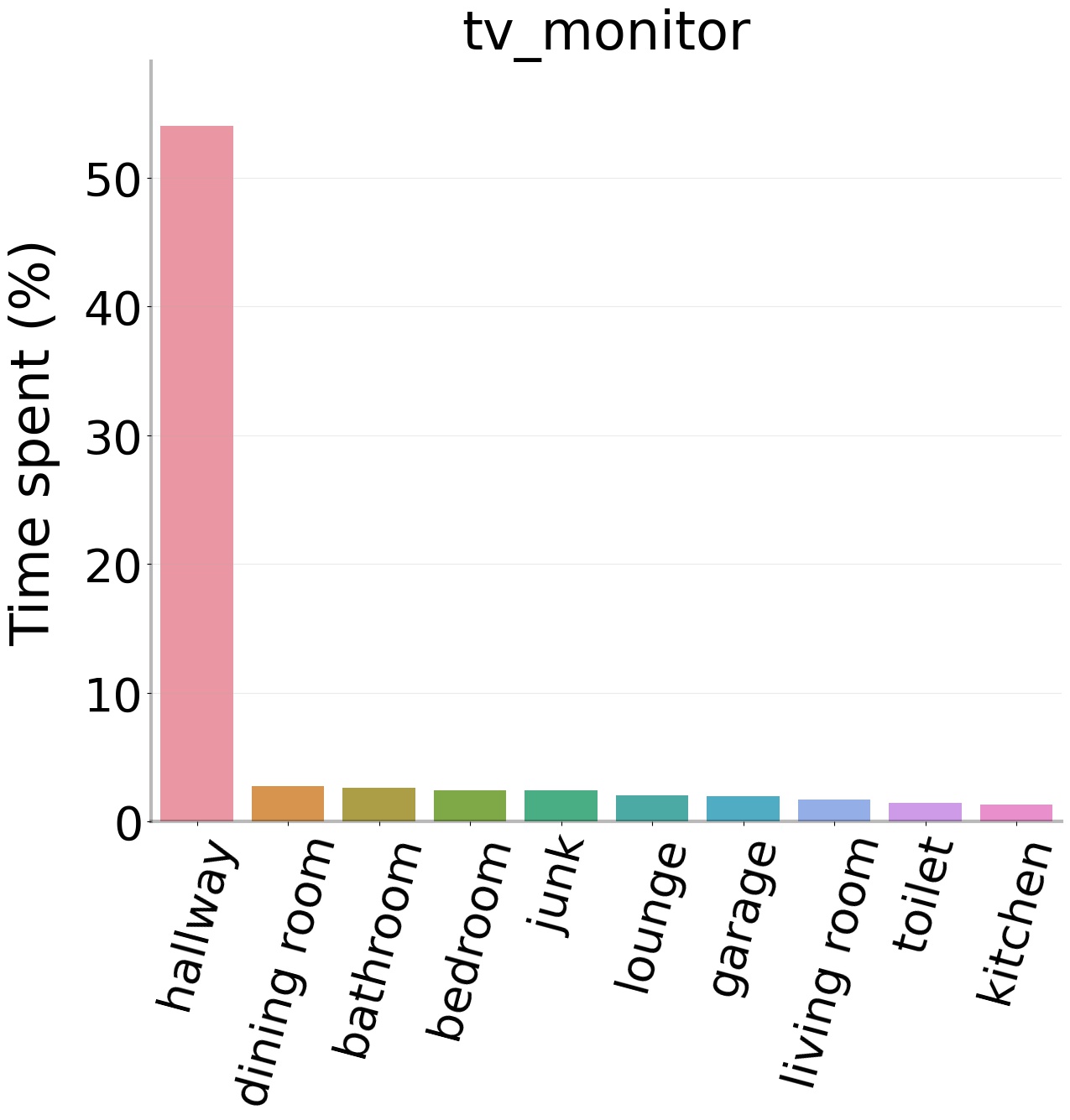}
            \caption{RL}
        \end{subfigure}
    \end{minipage}

    \caption{Comparison of per room time spent for all MP3D goal categories on \textsc{val} split for human demonstrations~\vs IL agents trained on human demos~\vs RL agents. The plot shows the top 10 rooms ordered by the maximum time spent in each room.}
    \label{fig:pRTS_per_object_7}
    
\end{figure*}